\documentclass{article}

\usepackage[preprint]{neurips_2025}

\usepackage[utf8]{inputenc} 
\usepackage[T1]{fontenc}    
\usepackage{url}            
\usepackage{booktabs}       
\usepackage{amsfonts}       
\usepackage{nicefrac}       
\usepackage{microtype}      
\usepackage{xcolor}         

\usepackage{enumitem}
\usepackage[utf8]{inputenc} 
\usepackage[T1]{fontenc}    
\usepackage{url}            
\usepackage{booktabs}       
\usepackage{amsfonts}       
\usepackage{nicefrac}       
\usepackage{microtype}      
\usepackage{colortbl}
\usepackage{amsmath}
\usepackage{amssymb}
\usepackage{listings}

\usepackage{algorithm}
\usepackage{algpseudocode}%
\usepackage{epsfig}
\usepackage{graphicx}
\usepackage{subcaption}

\usepackage{booktabs}
\usepackage{pifont}
\usepackage{makecell}  
\usepackage{amsmath,amssymb}         
\usepackage{enumitem}                
\usepackage{hyphenat}                
\usepackage{xspace}         
\usepackage{svg}
\usepackage{listings}                
\usepackage{CJKutf8}

\usepackage{multirow}

\usepackage[misc]{ifsym}

\newcommand{\cmark}{\ding{51}}%
\newcommand{\xmark}{\ding{55}}%

\newlength\savewidth\newcommand\shline{\noalign{\global\savewidth\arrayrulewidth
\global\arrayrulewidth 1pt}\hline\noalign{\global\arrayrulewidth\savewidth}}

\newcommand{\tablestyle}[2]{\setlength{\tabcolsep}{#1}\renewcommand{\arraystretch}{#2}\centering\footnotesize}

\usepackage{xcolor}         

\usepackage{hyperref}            
  \hypersetup{
    colorlinks=true,             
    linkcolor=black,              
    pdfauthor={Anonymous},       
    pdftitle={MMMG Appendix},
  }
\usepackage[capitalize]{cleveref}
\crefname{section}{Sec.}{Secs.}
\Crefname{section}{Section}{Sections}
\Crefname{table}{Table}{Tables}
\crefname{table}{Tab.}{Tabs.}
\definecolor{gray(x11gray)}{rgb}{0.75, 0.75, 0.75}

\usepackage{wrapfig} 
\usepackage{amssymb}
\usepackage{pifont}

\RequirePackage[breakable,skins]{tcolorbox}
\RequirePackage{xcolor}

\usepackage{array}
\usepackage{multirow}
\usepackage{amssymb}
\usepackage{caption}
\usepackage{subcaption}
\usepackage{booktabs}
\usepackage{pifont}
\usepackage{colortbl}
\usepackage{newtxtext}
\usepackage{graphicx}
\usepackage{subcaption}
\usepackage{tikz}

\usepackage{twemojis}

\definecolor{myblue}{HTML}{1C2143}
\newcommand\codeurl[1]{{{\color{blue}{\url{#1}}}}}

\newcommand{\bluecircle}[1]{%
  \tikz[baseline=(char.base)]{
    \node[fill=myblue,               
          circle,
          inner sep=1pt,              
          text=white,                 
          font=\sffamily\bfseries]    
          (char){#1};}}

\definecolor{colorcommentbg_knowledgeprompt}{HTML}{8C92AC}
\definecolor{colorcommentframe_knowledgeprompt}{HTML}{808080}

\newenvironment{GenQuestionPrompt}[1][]{
\begin{tcolorbox}[adjusted title={Question Generation Prompt}, fonttitle={\bfseries\footnotesize}, fontupper=\scriptsize, colback={colorcommentbg_knowledgeprompt!30}, colframe={colorcommentframe_knowledgeprompt!80},coltitle={white},#1]
}{\end{tcolorbox}}

\newenvironment{KGprompt}[1][]{
\begin{tcolorbox}[adjusted title={Knowledge Graph Generation Prompt}, fonttitle={\bfseries\footnotesize}, fontupper=\scriptsize, colback={colorcommentbg_knowledgeprompt!30}, colframe={colorcommentframe_knowledgeprompt!80},coltitle={white},#1]
}{\end{tcolorbox}}

\newenvironment{knowledgeprompt}[1][]{
\begin{tcolorbox}[adjusted title={Knowledge Image Evaluation Prompt for OpenAI o3}, fonttitle={\bfseries\footnotesize}, fontupper=\scriptsize, colback={colorcommentbg_knowledgeprompt!30}, colframe={colorcommentframe_knowledgeprompt!80},coltitle={white},#1]
}{\end{tcolorbox}}

\newenvironment{DataFilterPrompt}[1][]{
\begin{tcolorbox}[adjusted title={Data Filter prompt: consistency \& alignment / truncation}, fonttitle={\bfseries\footnotesize}, fontupper=\scriptsize, colback={colorcommentbg_knowledgeprompt!30}, colframe={colorcommentframe_knowledgeprompt!80},coltitle={white},#1]
}{\end{tcolorbox}}

\newenvironment{ReasoningPrompt}[1][]{
\begin{tcolorbox}[adjusted title={Thinking Process Annotation Prompt}, fonttitle={\bfseries\footnotesize}, fontupper=\scriptsize, colback={colorcommentbg_knowledgeprompt!30}, colframe={colorcommentframe_knowledgeprompt!80},coltitle={white},#1]
}{\end{tcolorbox}}

\title{{MMMG}: A Massive, Multidisciplinary, Multi-Tier Generation Benchmark for Text-to-Image Reasoning}

\makeatletter
\def\@fnsymbol#1{\ensuremath{\ifcase#1\or \dagger\or \dagger\or
\mathsection\or \mathparagraph\or \|\or **\or \dagger\ddagger
\or \dagger\dagger \else\@ctrerr\fi}}
\makeatother

\author{%
Yuxuan Luo$^1$\thanks{\noindent Research interns at Microsoft. $\ddag$: \twemoji{e-mail} \texttt{yuhui.yuan@microsoft.com}\; \twemoji{e-mail}\texttt{lianzhouhui@pku.edu.cn}}, Yuhui Yuan$^{4\ddag}$, Junwen Chen$^{2\dag}$, Haonan Cai$^1$, Ziyi Yue$^1$,\\ \textbf{Yuwei Yang}$^{3\dag}$\textbf{,} \textbf{Fatima Zohra Daha}$^5$\textbf{,} \textbf{Ji Li}$^5$\textbf{,} \textbf{Zhouhui Lian}$^{1\ddag}$ \\
$^1$Wangxuan Institute of Computer Technology, Peking University, China,\\$^2$The University of Electro-Communications, $^3$Australian National University\\$^4$Microsoft Research Asia, $^5$Microsoft\\
\small\codeurl{https://mmmgbench.github.io/}\vspace{-4mm}
}

\begin{document}
\maketitle

\begin{figure*}[!h]
\vspace{-6mm}
\centering
\begin{minipage}[t]{0.8\linewidth}
\includegraphics[width=\textwidth]{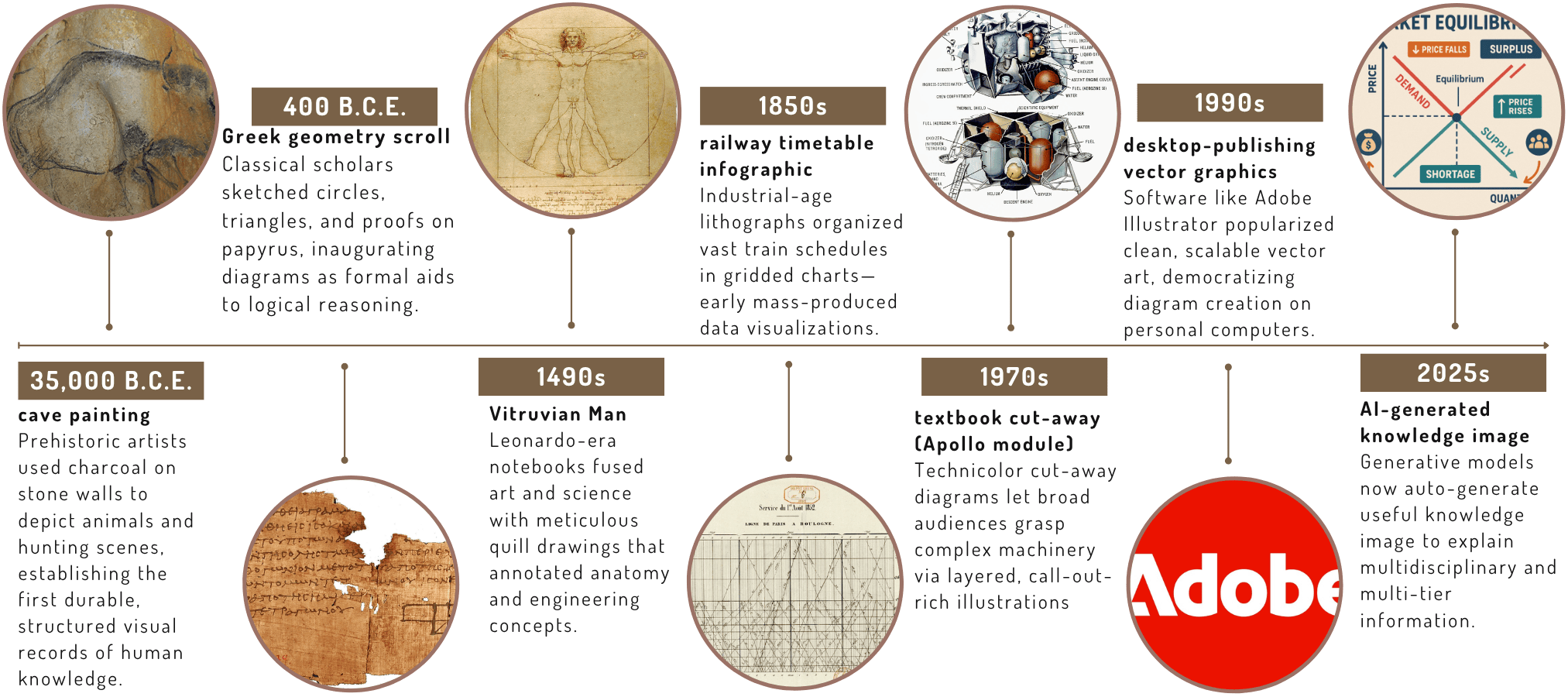}
\caption{\footnotesize{$40,000$ Years of Knowledge Image: From Cave Paintings to Generative AI.}}
\label{fig:knowledge_image_timeline}
\end{minipage}
\end{figure*}

\begin{abstract}
In this paper, we introduce knowledge image generation as a new task, alongside the Massive Multi-Discipline Multi-Tier Knowledge-Image Generation Benchmark (\texttt{MMMG}) to probe the reasoning capability of image generation models.
Knowledge images have been central to human civilization and to the mechanisms of human learning—a fact underscored by \emph{dual-coding theory} and the \emph{picture-superiority effect}\footnote{Both dual-coding theory and the picture-superiority effect principles suggest humans remember visuals more effectively than words, partly because visual information can engage multiple cognitive encoding pathways.\\This work was supported by the National Natural Science Foundation of China (Grant No.: 62372015) and Key Laboratory of Intelligent Press Media Technology.}.
Generating such images is challenging, demanding multimodal reasoning that fuses world knowledge with pixel-level grounding into clear explanatory visuals.
To enable comprehensive evaluation, \texttt{MMMG} offers $4,456$ expert-validated (knowledge) image-prompt pairs spanning $10$ disciplines, $6$ educational levels, and diverse knowledge formats such as charts, diagrams, and mind maps. 
To eliminate confounding complexity during evaluation, we adopt a unified Knowledge Graph (KG) representation. Each KG explicitly delineates a target image’s core entities and their dependencies.
We further introduce \texttt{MMMG-Score‌} to evaluate generated knowledge images. This metric combines factual fidelity, measured by graph-edit distance between KGs, with visual clarity assessment.
Comprehensive evaluations of $16$ state-of-the-art text-to-image generation models expose serious reasoning deficits—low entity fidelity, weak relations, and clutter—with GPT-4o achieving an \texttt{MMMG-Score‌} of only $50.20$, underscoring the benchmark’s difficulty.
To spur further progress, we release FLUX-Reason (\texttt{MMMG-Score‌} of $34.45$), an effective and open baseline that combines a reasoning LLM with diffusion models and is trained on $16,000$ curated knowledge image–prompt pairs.

\end{abstract}
\vspace{-3mm}
\section{Introduction}
\label{sec:intro}
\vspace{-3mm}

Reasoning-based large language models (LLMs) such as OpenAI-o1/o3~\cite{openai2024o1,openai2025o3mini,openai2025o3} and DeepSeek-R1~\cite{guo2025deepseek} excel on math and coding tests (AIME 2024~\cite{AIME2024}, Codeforces~\cite{CodeForces}, GPQA Diamond~\cite{GPQA}, MATH-500~\cite{MATH500}, MMLU~\cite{MMLU}, SWE-bench~\cite{SWE-bench}) thanks to rigorous, reasoning‐focused benchmarks. By contrast, widely used text-to-image benchmarks~\cite{T2Icompbench,CommonsenseT2I} still focus on instruction following and compositionality—e.g., attribute binding—while largely overlooking reasoning. The lack of reasoning-oriented benchmarks has left text-to-image generation models lagging significantly behind reasoning-focused LLMs.

Measuring reasoning in image generation is non-trivial. Current evaluations emphasize prompt-following, aesthetics, and visual–text rendering, typically quantified by CLIP~\cite{ruiz2023dreambooth}, FID~\cite{saharia2022photorealistic}, OCR and Aesthetic~\cite{discus04342023aesthetic25} scores. Yet producing such images seldom requires complex logical, domain-specific knowledge reasoning. Motivated by how humans leverage visuals to think, we introduce a new task—knowledge image generation: given only a vague user prompt, the model must autonomously infer the pertinent concepts (or entities) and relationships, and render them in a coherent knowledge image—such as a diagram, chart, infographic, or other annotated visual—that faithfully conveys the intended information.

\begin{figure}[t]
\vspace{-4mm}
\begin{minipage}[t]{1\linewidth}
\centering
\begin{subfigure}{0.195\textwidth}
\includegraphics[width=\textwidth]{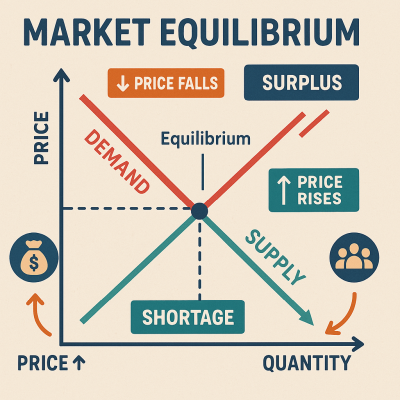}
\vspace{-5mm}
\end{subfigure}
\begin{subfigure}{0.195\textwidth}
\includegraphics[width=\textwidth]{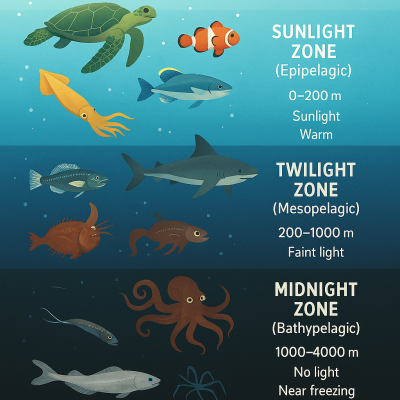}
\vspace{-5mm}
\end{subfigure}
\begin{subfigure}{0.195\textwidth}
\includegraphics[width=\textwidth]{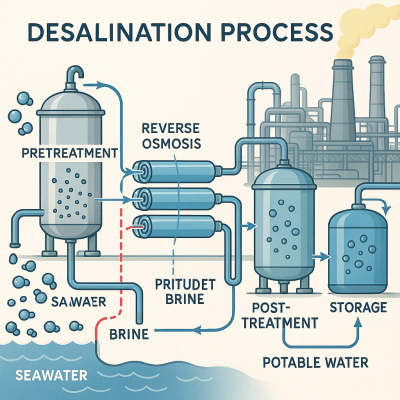}
\vspace{-5mm}
\end{subfigure}
\begin{subfigure}{0.195\textwidth}
\includegraphics[width=\textwidth]{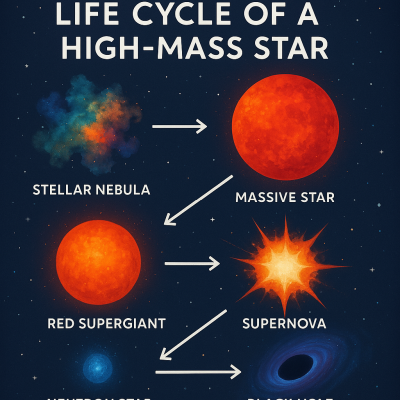}
\vspace{-5mm}
\end{subfigure}
\begin{subfigure}{0.195\textwidth}
\includegraphics[width=\textwidth]{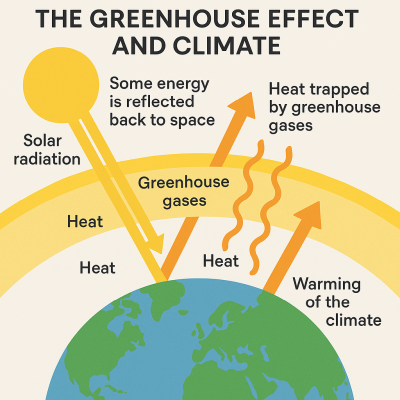}
\vspace{-5mm}
\end{subfigure}
\caption{\footnotesize{Representative knowledge images generated with GPT-4o. Disciplines (L-R): Economics, Oceanography, Environmental Engineering, Astrophysics, Climate Science. Details are in the supplementary.}}
\label{fig:knowledge_image}
\end{minipage}
\vspace{-4mm}
\end{figure}

Throughout human history, knowledge images have propelled progress for nearly $40,000$ years, serving as a lasting bridge that turns abstract ideas into concrete, shareable visual forms (Figure~\ref{fig:knowledge_image_timeline}).
Cognitive science also supports this direction—{the dual-coding theory}~\cite{dualcoding} and {the picture-superiority effect}~\cite{picture_superiority_effect} suggest humans encode information more robustly when language and imagery are combined. Creating such visuals, however, is intrinsically difficult: a model must fuse broad world knowledge with spatial composition, select salient entities, and faithfully ground relations in pixel space. Figure~\ref{fig:knowledge_image} visualizes several knowledge representations across disciplines.

To advance text-to-image reasoning, we introduce the ‌Massive Multi-Discipline Multi-Tier Knowledge-Image Generation Benchmark (\texttt{MMMG}). \texttt{MMMG} comprises $4,456$ expert-validated prompt–image pairs spanning ten academic disciplines—Biology ($850$), Chemistry ($328$), Mathematics ($399$), Engineering ($582$), Geography ($352$), Economics ($623$), Sociology ($479$), Philosophy ($210$), History ($327$), and Literature ($306$);—and six educational tiers: pre-school ($591$), primary school ($680$), secondary school ($693$), high school ($936$), undergraduate ($744$), and PhD ($812$). Each sample is annotated with a high-quality knowledge graph that lists the necessary entities and their dependencies, enabling format-agnostic coverage and requiring models to generalize across domains and reasoning levels. Its benefits are twofold: first, it abstracts core concepts into an interpretable graph, reducing the diversity and complexity of knowledge visuals; second, it enables objective fidelity evaluation via graph-edit distance between the ground-truth and generated graphs.

Benchmarking reasoning fidelity in generated images requires more than perceptual metrics. We therefore introduce \texttt{MMMG-Score}, which combines the graph-edit distance between knowledge graphs with a visual-clarity score derived from foundational segmentation models. Specifically, we employ the OpenAI-o3 reasoning LLM to analyze each image–prompt pair, and predict an initial knowledge graph. For the visual-clarity component, we run Segment Anything Model v2 (SAM-2)~\cite{SAM2} on the generated images and penalize overly cluttered and disorganized outputs that may “hack” the reasoning LLM to extract the unreliable knowledge graphs yet fail to convey the knowledge clearly. The importance of this visual-clarity metric is examined in the experimental section.

\begin{table}[ht]
\vspace{-2mm}
\centering
\tablestyle{3pt}{1.25}
\resizebox{1.0\linewidth}{!}{
\begin{tabular}{l|l|l|l|l|c|c}
\textbf{Benchmark} & \textbf{Scale}  & \textbf{Focus} & \textbf{Domains} & \textbf{Metrics} & \textbf{World Knowledge} & \textbf{Explanatory}  \\
\shline
GenEval~\cite{ghosh2023geneval} & 553 & Compositionality & counting, colors, position, attribute binding & Accuracy & \xmark & \xmark \\
T2I-CompBench++~\cite{T2ICompbench++} & 6,000 & Compositionality & Object-Attribute Binding & BLIP-VQA, UniDet, CLIP & \xmark & \xmark \\
DPG-Bench~\cite{DPGBench} & 1,065 & Prompt Adherence & Dense Scene Generation & CLIP, Human Eval & \xmark & \xmark \\
Commonsense-T2I~\cite{CommonsenseT2I} & 1,000+ & Commonsense Reasoning & Everyday Scenarios & Accuracy & \xmark & \xmark \\
Winoground-T2I~\cite{WinogroundT2I} & 11,000 & Compositionality & 20 Types & Contrastive Accuracy & \xmark & \xmark \\
TIFA~\cite{hu2023tifa} & 1,000 & Faithfulness & General Knowledge & VQA-based & \xmark & \xmark \\
TypeScore~\cite{sampaio2024typescore} & 1,000 & Text Fidelity & Scene Text & OCR-based & \xmark & \xmark \\
WISE~\cite{WISE} & 1,000 & Commonsense Reasoning & Science, Culture, Space-Time & LLM-Judged& \cmark & \xmark  \\ \hline
\texttt{MMMG} (Ours)& 4,456 & Disciplinary Knowledge & 10+ Academic Fields & Readability, Graph Edit Distance & \cmark & \cmark \\
\end{tabular}
}
\caption{\small{Comparison with previous Text-to-Image (T2I) benchmark.}}
\label{tab:t2i_benchmarks}
\vspace{-8mm}
\end{table}

We conduct comprehensive evaluations of $16$ state-of-the-art text-to-image models—LlamaGen, JanusFlow, Emu-3, SimpleAR, Janus-Pro, CogView, SEED-X, SDXL-1.0, SDXL-1.0-refiner, Infinity, FLUX.1-[dev], FLUX.1-[pro], Ideogram 2.0, HiDream-l1-Full, BAGEL, and GPT-4o image generation—on the \texttt{MMMG} benchmark, reporting their FID, aesthetic, WISE~\cite{WISE}, and \texttt{MMMG-Score} metrics. We also conduct human studies to confirm that \texttt{MMMG-Score} aligns best with user judgements, underscoring the value of knowledge-graph–based evaluation.
Our \texttt{MMMG} benchmark presents significant challenges: even the GPT-4o image generation achieves only \texttt{MMMG-Score} of $46.66$, while the next-best model, the open-source HiDream-l1-Full, reaches just \texttt{MMMG-Score} of $25.72$.
To catalyze further research, we release \texttt{FLUX-Reason}, a fully reproducible and open-source baseline that pairs a reasoning-oriented LLM (e.g., OpenAI-o3 or DeepSeek-R1) with a diffusion model (FLUX.1-[dev]) trained on $16,000$ curated knowledge-image pairs. Although its \texttt{MMMG-Score} of $30.52$ still trails that of GPT-4o, \texttt{FLUX-Reason} serves as an open source baseline and underscores the new challenges posed by the \texttt{MMMG} benchmark for next-generation reasoning-oriented text-to-image generation models.
\vspace{-4mm}
\section{Related Work}
\vspace{-2mm}
\noindent\textbf{Benchmarks for Text-to-Image Generation.} 

Many benchmarks have been developed to assess both the limitations and progress of recent text-to-image models. We summarize the comparison between \texttt{MMMG} and prior benchmarks in Table~\ref{tab:t2i_benchmarks}. GenEval~\cite{ghosh2023geneval} introduces object detectors for fine-grained, object-level evaluation, addressing the shortcomings of holistic metrics. T2I-CompBench++~\cite{T2ICompbench++} increases compositional difficulty via prompts involving attributes, relationships, numeracy, and complex scenes. Commonsense-T2I~\cite{CommonsenseT2I} uses adversarial prompts to probe visual commonsense reasoning. Winoground-T2I~\cite{WinogroundT2I} evaluates compositional generalization with contrastive sentence pairs. DPG-Bench~\cite{DPGBench} targets instruction-following with longer, text-rich prompts. The concurrent WISE benchmark~\cite{WISE} is most related, focusing on world knowledge-based evaluation across cultural, scientific, and temporal domains. However, WISE emphasizes photorealism with implicit knowledge, while \texttt{MMMG} requires models to explicitly visualize structured world knowledge in a semantically grounded and explanatory manner.

\noindent\textbf{Reasoning in Text-to-Image Generation.}
While LLMs have achieved significant progress in reasoning through techniques such as chain-of-thought prompting~\cite{wei2022chain,kojima2022large} and large-scale reinforcement learning~\cite{guo2025deepseek,openai2024o1}, recent models excel in benchmarks focused on mathematics, coding, and tool use (e.g., MMLU~\cite{MMLU}, AGIEval~\cite{AGIEval}, LogicBench~\cite{LogicBench}, MathVista~\cite{lu2023mathvista}). Inspired by this progress, several works have explored injecting reasoning into image generation, including ImageGen-CoT~\cite{ge2024seed, liao2025imagegen}, HiDream~\cite{hidream2024hidreami1}, T2I-R1~\cite{jiang2025t2i}, Meta-Queries~\cite{pan2025transfer}, and MINT~\cite{wang2025mint}. However, these methods are typically evaluated on prior benchmarks and rely on caption-based metrics (e.g., CLIPScore~\cite{radford2021learning}) or subjective human preference, both of which lack fidelity in assessing reasoning ability. To address this gap, \texttt{MMMG} introduces a knowledge image generation task requiring advanced multimodal reasoning, along with \texttt{MMMG-Score}, a structured metric that compares extracted and ground-truth knowledge graphs. We further propose FLUX-Reason, a reasoning-enhanced model, and evaluate it on the \texttt{MMMG} benchmark.

\vspace{-7mm}
\section{Method}
\vspace{-2mm}
We first illustrate the definition of the knowledge image generation task by leveraging additional knowledge graph annotations. Next, we describe how we build the \texttt{MMMG} benchmark, and provide an overview of key dataset statistics. We then introduce the novel \texttt{MMMG-Score}, a metric we propose for more reliable evaluation. Last, we present details of our strong baseline, FLUX-Reason, which explicitly combines a reasoning LLM with a text-to-image generation model in a cascaded manner.

\begin{figure}[t]
\centering
\vspace{-3mm}
\includegraphics[width=\linewidth]{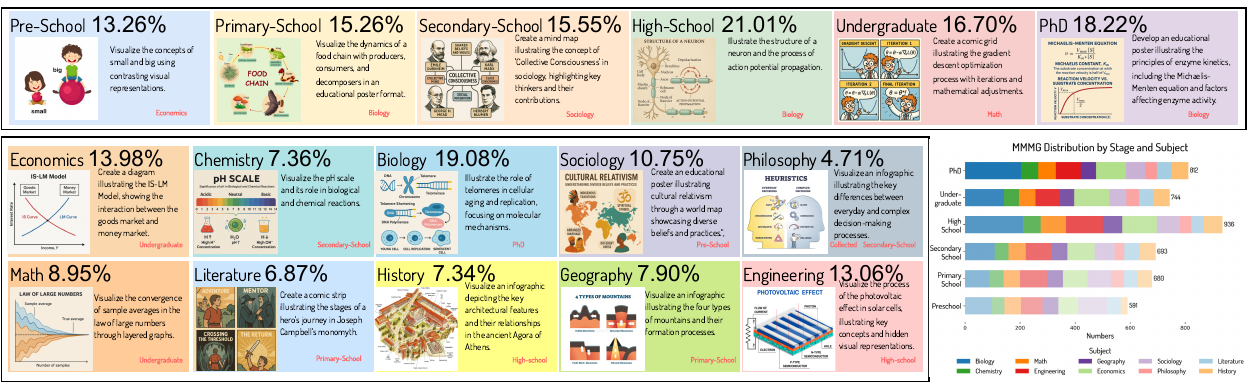}
\vspace{0.25mm}
\begin{subfigure}[b]{\linewidth}
\centering
\includegraphics[width=\linewidth]{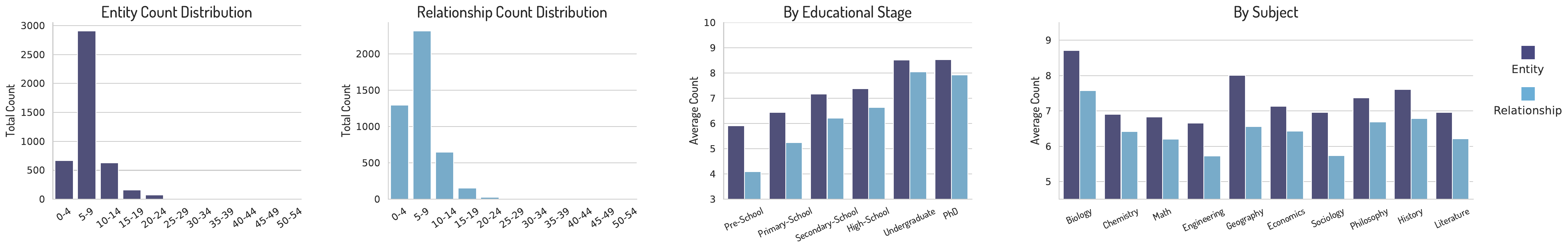}
\end{subfigure}
\vspace{-5mm}
\caption{\footnotesize \textbf{\texttt{MMMG} Dataset Statistics:} The top panel shows \texttt{MMMG} test dataset distribution across educational levels and disciplines. The bottom panel presents statistics of both train and test sets, while bottom right depicts knowledge graph complexity increase across educational stages and differ among subjects.}
\vspace{-4mm}
\label{fig:statistics}
\end{figure}
\vspace{-2mm}
\subsection{Knowledge Image Generation}
\vspace{-1mm}
\noindent\textbf{Formulation.} The knowledge image generation task begins with a concise, question-like prompt $\mathbf{X}$ and employs a generative model $f$ to produce a knowledge image $\mathbf{Y}$ conditioned on $\mathbf{X}$. A knowledge image typically comprises multiple entities and their interrelationships. To capture this structure, we extract an auxiliary knowledge graph $\mathbf{G} = (\mathcal{E}, \mathcal{D})$ using a reasoning LLM (e.g., OpenAI-o3), which takes both the text prompt and the target image (if available) as input. Here, $\mathcal{E} = \{\mathbf{e}_1, \dots, \mathbf{e}_n\}$ denotes the set of graph nodes representing entities, and $\mathcal{D} = \{d_i(\mathbf{e}_j, \mathbf{e}_k)\}_{i=1}^{K}$ denotes the set of edges encoding relationships between entities.

\noindent\textbf{Importance of Knowledge Graph.} 
{The knowledge graph $\mathbf{G}$ is essential for evaluating whether the generated image faithfully visualizes the domain knowledge implied by the prompt $\mathbf{X}$. Since $\mathbf{X}$ is typically a concise question rather than a descriptive instruction—e.g., “\emph{Illustrate the structure of a neuron and the process of action potential propagation}”—it is not well-aligned with the visual content, making CLIP-based verification unreliable.}

\noindent\textbf{Knowledge Graph Extraction.}
Accurate knowledge graph extraction requires inferring world knowledge from the text prompt $\mathbf{X}$ and identifying the corresponding visual entities and relations in the image $\mathbf{Y}$. We use OpenAI-o3 to generate knowledge graphs from $\mathbf{X}$ and $\mathbf{Y}$, and rely on domain experts to verify the results and filter out low-quality cases. Refer to the supplementary material for the system prompt.

\noindent\textbf{Relationship Formalization.} To ensure that the knowledge graph can represent diverse knowledge across six educational levels and ten disciplines, we propose a domain-agnostic set of predicates for relationships:
$\operatorname{Defines}(\cdot, \cdot)$, $\operatorname{Entails}(\cdot, \cdot)$, $\operatorname{Causes}(\cdot, \cdot)$, $\operatorname{Contains}(\cdot, \cdot)$, $\operatorname{Requires}(\cdot, \cdot)$, and $\operatorname{TemporalOrder}(\cdot, \cdot)$, with optional dynamic modifiers such as {change(·)} to represent trends or shifts.
For instance, $\operatorname{Causes}(\operatorname{increase}(e_1), \operatorname{decrease}(e_2))$ may represent a graph edge where an increasing population (denoted as $e_1$) leads to reduced biodiversity (denoted as $e_2$).

In the neuron example, we can extract a non-trivial knowledge graph consisting of $9$ entities, $\mathcal{E} =$\{dendrites, cell body, nucleus, axon, myelin sheath, schwann cell, node of ranvier, action potential propagation, depolarization\}, and $8$ relationships, $\mathcal{D} =$\{$\operatorname{Contains}$(dendrites, cell body), $\operatorname{Contains}$(cell body, nucleus), $\operatorname{Contains}$(cell body, axon), $\operatorname{Contains}$(axon, myelin sheath), $\operatorname{Contains}$(myelin sheath, schwann cell), $\operatorname{Contains}$(axon, node of Ranvier), $\operatorname{Causes}$(depolarization, action potential propagation), $\operatorname{Requires}$(action potential propagation, axon)\}. {This abstraction provides two key benefits:
(i) structural consistency across disciplines and educational levels, and
(ii) automated evaluation using normalized Graph Edit Distance (GED) to assess factual alignment with reference graphs.}

\vspace{-3mm}
\subsection{MMMG Benchmark: Statistics and Curation Process}
\vspace{-2mm}
\noindent\textbf{Statistics.} Figure~\ref{fig:statistics} provides an overview of the dataset statistics for \texttt{MMMG}, which spans six educational stages and ten academic disciplines. \texttt{MMMG} contains 4,456 expert-collected prompt–image pairs. We analyze the statistics as follows:

\noindent\textbullet\ At the top of Figure~\ref{fig:statistics}, we present the distribution across six different educational levels and illustrate representative examples to demonstrate how the inherent challenges increase from pre-school to PhD-level knowledge images. We ensured a balanced distribution across educational levels.

\noindent\textbullet\ In the middle of Figure~\ref{fig:statistics}, we present the distribution across ten academic disciplines. We find that biology, economics, and engineering are the dominant domains that rely more on knowledge images than others—especially philosophy, which accounts for only $4.71\%$ of the dataset. On the right side of the middle, we visualize the distribution of all $10$ disciplines across the $6$ educational levels.

\noindent\textbullet\ At the bottom of Figure~\ref{fig:statistics}, we highlight the complexity of knowledge image generation by showing statistics on the number of entities and relationships in the dataset. We find that nearly $3,000$ samples require generating $5\sim10$ entities and $5\sim10$ relationships. On the right side of the third row, we report the distributions of these statistics across different educational levels and disciplines.

\begin{figure}
  \centering
  \includegraphics[width=\linewidth]{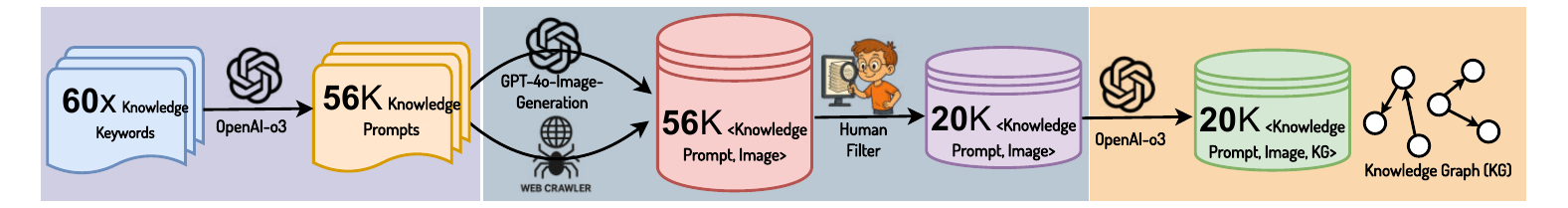}
  \caption{\footnotesize\textbf{\texttt{MMMG} Dataset Construction Pipeline}: From knowledge keywords across six educational levels and ten disciplines, we generate $56$K prompts using OpenAI-o3. These prompts are clustered into semantic groups and then sent to either GPT-4o for image generation or to a web crawler for image retrieval. The resulting $56$K knowledge images are filtered down to $20$K through a cascade of automated steps and human filter. Last, we use OpenAI-o3 to extract a knowledge graph for each prompt–image pair.}
  \label{fig:DATA-FLYWHEEL}
\end{figure}

\noindent\textbf{Curation Pipeline.}
Figure~\ref{fig:DATA-FLYWHEEL} illustrates the overall pipeline of how we curate the \texttt{MMMG} dataset. We start with $60\times$ knowledge keywords and apply a cascade of OpenAI-o3, crawler, and human expert filter to construct a raw dataset of $20,000$ samples, from which we select a subset with the best quality to form the \texttt{MMMG} benchmark.

\vspace{1mm}
\noindent\textbullet\ Knowledge Keywords \bluecircle{1} $\to$  Knowledge Prompts \bluecircle{2}:
In the left of Figure~\ref{fig:DATA-FLYWHEEL}, we first apply OpenAI-o3 to generate approximately 56,830 knowledge text prompts by combining two keywords: one specifying the educational level and one specifying the discipline. The educational level keyword is sampled from a seed set of six candidates: [pre-school, primary school, secondary school, high school, undergraduate, PhD], while the disciplinary keyword is sampled from ten candidates: [economics, chemistry, biology, sociology, philosophy, math, literature, history, geography, engineering].

\vspace{1mm}
\noindent\textbullet\ Knowledge Prompts \bluecircle{2} $\to$  Knowledge Images \bluecircle{3}:
We first cluster the 56,830 knowledge text prompts into 11,732 concepts using SentenceTransformer embeddings and DBSCAN. Then, we split these knowledge text prompts into two groups: one for GPT-4o image generation (30K samples) and one for web crawling (26K samples).

\vspace{1mm}
\noindent\textbullet\ Knowledge Images Filter \bluecircle{3} $\to$ \bluecircle{4}: We apply deduplication\footnote{\codeurl{https://github.com/idealo/imagededup}} and OCR-based filtering to remove duplicates and samples lacking explanatory visual text. During GPT-4o image generation, cropping artifacts often harm visual completeness. To address this, we use OpenAI-o3 to detect severe cropping. We also involve human experts to verify text-image alignment. These steps yield a curated set of 20K high-quality knowledge image–text pairs.

\vspace{1mm}
\noindent\textbullet\ Knowledge Prompts, Images \bluecircle{4} $\to$ Knowledge Graphs \bluecircle{5}: We use OpenAI-o3 to generate a structured knowledge graph for each of the 20K prompt–image pairs, following the format described earlier. We also use DeepSeek-R1 to produce step-by-step reasoning over the graph's entities and relations. The 20K samples are split by their topics and concepts to ensure minimal overlap. Human experts then verify all samples and select the most accurate ones, resulting in the \texttt{MMMG} benchmark of 4,456 high-quality pairs.

\subsection{MMMG-Score: Measuring Knowledge Fidelity and Visual Readability}
\label{sec:MMMG_Eval}

As perceptual metrics like FID or CLIP are insufficient for reasoning evaluation, we propose \texttt{MMMG-Score}, a novel metric combining knowledge fidelity and visual readability.

\begin{figure}
    \centering
    \vspace{-3mm}
    \includegraphics[width=\linewidth]{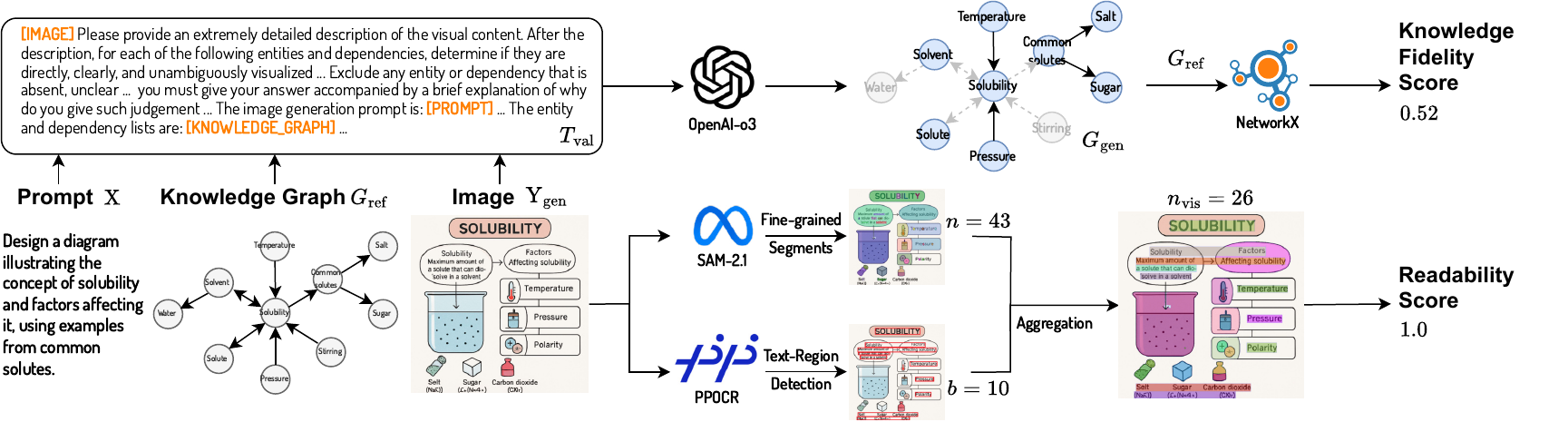}
    \caption{\footnotesize Illustration of \texttt{MMMG-Score} computation: We compute the knowledge fidelity score using graph edit distance, and the readability score by counting the number of segments in the generated knowledge image.}
    \vspace{-3mm}
    \label{fig:MMMG-EVAL}
\end{figure}

\textbf{Knowledge Fidelity via Grounded Knowledge Graph Extraction.} 
Given a generated image $\mathbf{Y}_{\mathrm{gen}}$, a knowledge prompt $\mathbf{X}$, and the reference knowledge graph $\mathbf{G}_{\mathrm{ref}}$, we use OpenAI-o3 (high reasoning effort) to ground $\mathbf{G}_{\mathrm{ref}}$ into the pixel space of $\mathbf{Y}_{\mathrm{gen}}$. This produces a grounded subgraph $\mathbf{G}_{\mathrm{gen}} \subseteq \mathbf{G}_{\mathrm{ref}}$, representing the knowledge captured by the image. When the graph is complex, OpenAI-o3 may fail to detect all entities or relations; we default missing items to false and leave improved handling for future work. As shown in Figure~\ref{fig:MMMG-EVAL}, we compute the knowledge fidelity score as $1{-}\operatorname{GED}(\mathbf{G}_{\mathrm{gen}}, \mathbf{G}_{\mathrm{ref}})$ using NetworkX\footnote{\codeurl{https://github.com/networkx/networkx}}, which rewards smaller edit distances between the two graphs.

\textbf{Visual Readability via Foundation Segmentation Model.} 

Readability is critical for knowledge-image generation, ensuring effective information delivery. To assess it, we compute a Readability Score using a segmentation model and a text detector, rewarding coherent regions and penalizing excessive fragmentation. As shown in Figure~\ref{fig:MMMG-EVAL} (bottom right), we use SAM-2.1~\cite{ravi2024sam} for segmentation (seeded with $32\times32$ uniform points, NMS threshold $0.6$) and PaddleOCR\footnote{\codeurl{https://github.com/PaddlePaddle/PaddleOCR}} for text detection. Overlapping masks and text boxes are merged, and the final region count defines the Readability Score:
\begin{equation}
R(n_{\mathrm{vis}}) =
\begin{cases}
1, & n_{\mathrm{vis}} \le n_{\min},\\[4pt]
\dfrac{n_{\max} - n_{\mathrm{vis}}}{n_{\max} - n_{\min}}, & n_{\min} < n_{\mathrm{vis}} < n_{\max},\\[4pt]
0, & n_{\mathrm{vis}} \ge n_{\max}.
\end{cases}
\end{equation}
Empirically, we set \(n_{\min} = 70\), \(n_{\max} = 160\), based on the segment distribution across common text-to-image generation models. The final \texttt{MMMG-Score} is computed by multiplying the above knowledge fidelity score and readability clarity score:
\begin{equation}
\texttt{MMMG\text{-}Score}(\mathbf{Y}_{\mathrm{gen}})
= R(n_{\mathrm{vis}}) \cdot \left[1 - \operatorname{GED}(\mathbf{G}_{\mathrm{gen}}, \mathbf{G}_{\mathrm{ref}})\right]
\;\in\;[0, 1].
\end{equation}

We adopt the above product-based metric to ensure the generated image accurately conveys the target knowledge graph (via graph edit distance) and remains visually clear (via readability score). Empirically, our \texttt{MMMG-Score} shows stronger alignment with human ratings than perceptual metrics like FID, CLIP, or aesthetic scores.

\subsection{FLUX-Reason}
\begin{figure}[t]
    \centering
    \includegraphics[width=\linewidth]{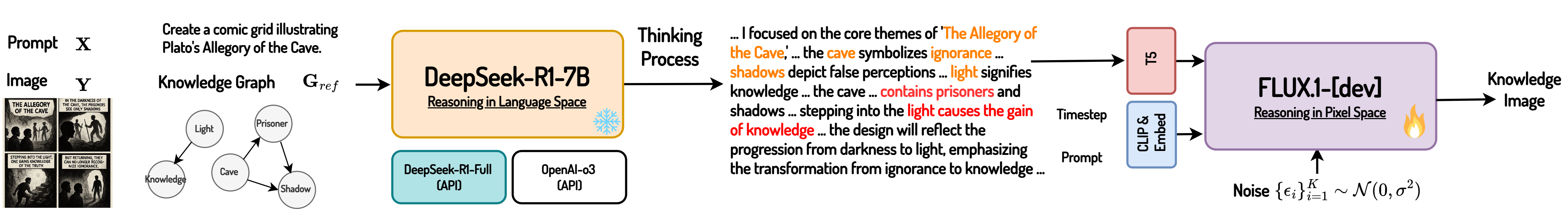}
    \caption{
        Overview of the FLUX-Reason pipeline. Reasoning LLMs first generate chain-of-thought (CoT) trajectories and visual planning cues conditioned on structured knowledge graphs. In the reasoning trace, entities and relations are highlighted in \textcolor{orange}{orange} and \textcolor{red}{red}, respectively. These traces are then encoded into the diffusion model to guide visual planning in pixel space.
        }
    \label{fig:flux-reason}
    \vspace{-5mm}
\end{figure}

We design FLUX-Reason to enhance reasoning capabilities in knowledge image generation by explicitly integrating a reasoning LLM with a diffusion-based generator. As illustrated in Figure~\ref{fig:flux-reason}, it consists of three variants: \texttt{FLUX-Reason (R1-7B)} incorporates \texttt{DeepSeek-R1-Distill-Qwen-7B} for local inference; \texttt{FLUX-Reason (R1)} queries the \texttt{DeepSeek-R1-Full} API; \texttt{FLUX-Reason (o3)} utilizes \texttt{OpenAI-o3} to produce summarized reasoning chains.

To supervise training, we extract chain-of-thought (CoT) reasoning traces from 16K GPT-4o-generated samples, each annotated with a ⟨prompt, image, knowledge graph (KG)⟩ triplet. Conditioned on explicit entities and relations, the extracted reasoning traces provide fine-grained guidance for concept selection, interaction modeling, and spatial arrangement.

To incorporate such long-form structured reasoning into generation, we extend the T5 encoder’s input length to 2048 tokens to accommodate the textual reasoning input, which is then transformed into pixel-space representations. 
The diffusion model is fine-tuned with LoRA over 10K steps, enabling it to learn pixel-level planning aligned with the structured reasoning trajectory.

\section{Experiments}

\subsection{Main Results}

\begin{table}[htbp]
\vspace{-2mm}
  \centering
  \footnotesize
  \caption{\footnotesize{\texttt{MMMG-Score} (×100) across prevalent image generation models, covering Diffusion, AR-based, and Multimodal architectures, evaluated over six educational stages.}}%
  \label{tab:edu_stages_results_0}
\tablestyle{3pt}{1.2}
\resizebox{0.99\linewidth}{!}{
  \begin{tabular}{l| c| l| c c c c c c|c}
    {Model} & {Resolution} & {Type} & {Preschool} & {Primary} & {Secondary} & {High} & {Undergrad} & {PhD} & {Avg} \\
    \shline
    LlamaGen          & 512   & AR           & 8.24 & 3.77 & 2.44 & 1.44 & 1.08 & 1.14 & 3.02 \\
    Emu-3             & 720   & MM           & 12.44 & 7.12 & 6.41 & 5.28 & 2.65 & 2.74 & 6.11 \\
    JanusFlow-1.3B    & 384   & AR           & 24.11 & 12.72 & 8.81 & 5.56 & 3.57 & 3.82 & 9.77 \\
    SimpleAR          & 1024  & AR           & 23.12 & 11.97 & 8.96 & 6.44 & 4.36 & 3.99 & 9.81 \\
    Ideogram          & 1024  & DM           & 20.39 & 14.14 & 12.90 & 9.68 & 8.41 & 7.73 & 12.21 \\
    Janus-pro-7B      & 384   & AR           & 29.50 & 16.72 & 12.73 & 8.45 & 5.57 & 5.66 & 13.10 \\ 
    CogView-4         & 1024  & DM           & 24.61 & 16.02 & 13.91 & 10.02 & 7.30 & 6.73 & 13.10 \\
    BAGEL           & 1024  & MM           & 29.29 & 19.42 & 15.29& 11.11 & 7.40 & 7.60 &  15.02\\
    SDXL-1.0          & 1024  & DM           & 23.41 & 19.12 & 17.41 & 16.26 & 9.92 & 9.29 & 15.90 \\
    SDXL-1.0-refiner  & 1024  & DM           & 24.55 & 19.24 & 18.59 & 16.72 & 9.68 & 8.94 & 16.29 \\
    FLUX.1-[dev] (recaption) & 1024 & DM         & 28.05 & 20.29 & 20.70 & 15.74 & 12.59 & 11.20 & 18.10 \\
    SEED-X            & 1024  & MM           & 33.41 & 22.67 & 19.49 & 15.74 & 8.88 & 8.76 &  18.16\\
    FLUX.1-[dev]          & 1024  & DM           & 29.80 & 23.09 & 20.99 & 16.12 & 12.47 & 12.30 & 19.13 \\
    Infinity          & 1024  & AR           & 25.87 & 20.63 & 21.86 & 18.36 & 14.23 & 14.14 & 19.18 \\
    FLUX.1-[pro]          & 1024  & DM           & 42.27 & 30.10 & 29.15 & 23.40 & 19.32 & 18.61 &  27.14\\
    HiDream-I1-Full   & 1024  & DM           & 42.86 & 31.77 & 30.26 & 23.39 & 19.88 & 20.05 &  28.04\\
    GPT-4o          & 1024  & MM           & \textbf{64.78} & \textbf{51.94} & \textbf{53.04} & \textbf{51.29} & \textbf{41.52} & \textbf{38.60} & \textbf{50.20} \\
    \hline
    FLUX-Reason (o3) & 1024 & DM       & 37.83 & 29.72 & 29.50 & 23.62 & 20.29 & 18.73 & 26.62 \\
    FLUX-Reason (R1-7B) & 1024 & DM & 44.93 & 34.41 & 34.19 & 28.70 & 23.36 & 21.99 & 31.26\\
    FLUX-Reason (R1) & 1024 & DM      & 49.10 & 39.39 & 37.00& 33.65 & 24.96 & 22.57 & 34.45 \\
  \end{tabular}
}
\vspace{-2mm}
\end{table}

We benchmark $16$ state-of-the-art T2I models and 3 FLUX-Reason variants, spanning three paradigms: \textbf{autoregressive (AR)} models, including \texttt{JanusFlow-1.3B}~\cite{janusflow}, \texttt{Janus-pro-7B}~\cite{januspro}, \texttt{LlamaGen}~\cite{llamagen}, \texttt{SimpleAR}~\cite{simplear}, and \texttt{Infinity}~\cite{Infinity}; \textbf{diffusion-based (DM)} models, including \texttt{SDXL-1.0}, \texttt{SDXL-1.0-refiner}~\cite{stabilityai2023sdxl}, \texttt{Ideogram}~\cite{ideogram2023}, \texttt{CogView-4}~\cite{zheng2024cogview3}, \texttt{HiDream-I1-Full}~\cite{hidream2024}, \texttt{FLUX.1-[dev]}~\cite{flux2024}, re-captioned \texttt{FLUX.1-[dev]}, \texttt{FLUX.1-[pro]}~\cite{flux-pro}; and \textbf{multimodal (MM)} models, including \texttt{Emu-3}~\cite{wang2024emu3}, \texttt{Seed-X}~\cite{ge2024seed}, \texttt{BAGEL}~\cite{deng2025bagel} and \texttt{GPT-4o}~\cite{openai2024gpt4o}. All models are evaluated with a fixed seed of 42. DM models use a classifier-free guidance (CFG) scale of 3.5, while AR and MM models apply default decoding settings. Open-source models are experimented on with 6$\times$A40 (512$^2$) or 4$\times$A100 (1024$^2$), depending on the image resolution. API-based models are queried directly.

\noindent\textbf{Performance Variation with Educational Level.} 
Table~\ref{tab:edu_stages_results_0} shows a clear performance drop as task complexity increases with education level. Most models, regardless of architecture, perform reasonably at pre-school (e.g., 20–30), but fall to low scores (below 10) at the PhD level, exposing their limitations in abstract reasoning and compositional planning. \texttt{GPT-4o} stands out with strong, stable performance (average: 50.20), showing robust generalization even on underspecified prompts.

\noindent\textbf{Model Highlights.}
HiDream-I1-Full shows competitive scores (28.04) despite being open-source, likely benefiting from structured priors in its Llama-based encoder. Similarly, \texttt{FLUX.1-[pro]} (27.14) and \texttt{SEED-X} (18.16) outperform many AR and MM models, indicating advantages of diffusion planning. \texttt{BAGEL} underperformed due to its overly brief "thinking" trajectories, hindering the delineation of detailed entities and intricate relationships.

\noindent\textbf{Reasoning vs. Recaptioning.}
To assess the impact of reasoning, we compare \texttt{FLUX.1-[dev]}, a recaptioned variant using \texttt{OpenAI-o3} (512-token prompts), and our reasoning-guided \texttt{FLUX-Reason (R1)}. While recaptioning reduces performance across levels, reasoning traces yield substantial gains—particularly at higher tiers. \texttt{FLUX-Reason (R1)} reaches an average score of 34.45, confirming that structured reasoning, not verbosity, is crucial for knowledge-grounded image generation.

\begin{figure}
    \centering
    \begin{minipage}{\textwidth}
        \includegraphics[width=\textwidth]{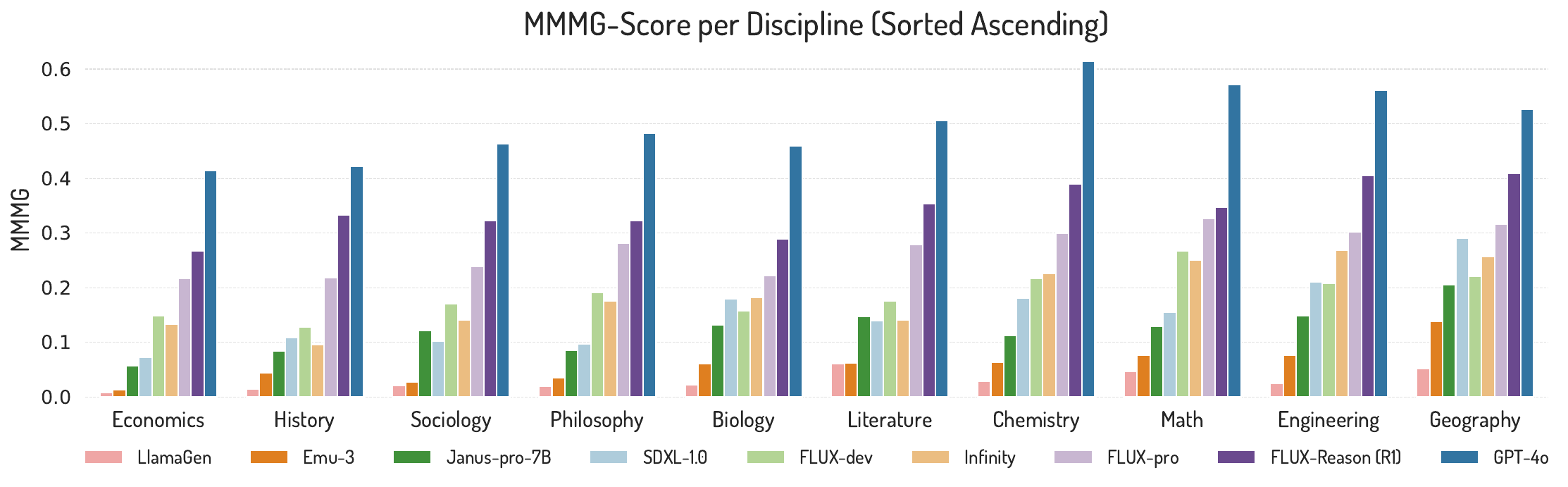}
        \caption{{Discipline-level evaluation.} Domains are sorted by average \texttt{MMMG-Score}.}
        \label{fig:curriculum-illust}
    \end{minipage}
    \hfill
    \begin{minipage}{\textwidth}
        \includegraphics[width=\linewidth]{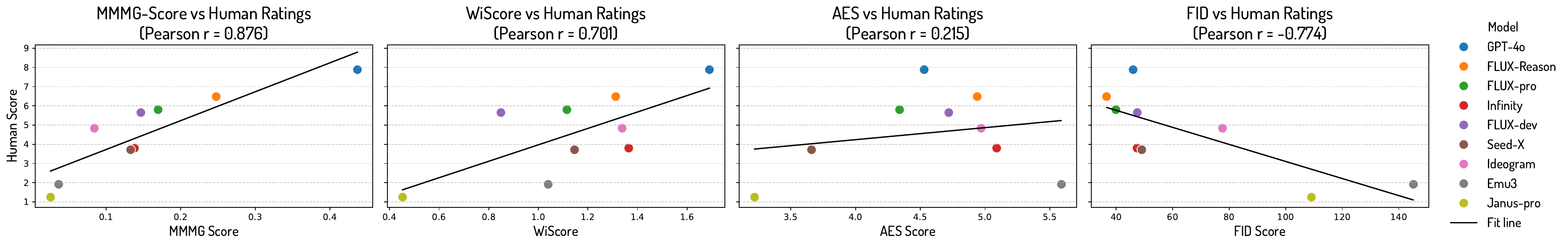}
        \caption{Illustration of Pearson Correlation among \texttt{MMMG-Score}, WiSCore, Aes-2.5 and FID.}
        \label{fig:pearson-corr}
    \end{minipage}
    \vspace{-3mm}
\end{figure}

\noindent\textbf{Discipline-Level Observations.}
Figure~\ref{fig:curriculum-illust} reveals domain-specific reasoning challenges. Weaker models (e.g., \texttt{LlamaGen}, \texttt{Emu-3}) perform best in Geography and Literature, where visuals are descriptive and align better with pretraining data. In contrast, Economics, History, and Sociology remain difficult even for stronger models due to their reliance on charts, temporal events, and abstract social concepts—structures rarely seen during pretraining.

A notable divergence appears between Chemistry and Biology: while both start similarly, Chemistry performs better with stronger models, likely due to its standardized diagram formats and symbolic representations. In contrast, Biology's visuals are often more irregular and spatially complex, making them less amenable to straightforward interpretation by such models.

Mathematics and Engineering perform well despite textual abstraction, suggesting structured visuals (e.g., geometry, schematics) are more model-friendly than symbolic reasoning. Since 77\% of \texttt{MMMG} images are human-designed, these trends also reflect real-world preferences. Overall, \texttt{MMMG} exposes domain gaps and visual-semantic reasoning challenges overlooked by text-only benchmarks.

\subsection{Human Alignment and Metric Comparison}

To assess alignment with human perception, we collected over 1,200 expert ratings (0–10 on clarity, correctness, accuracy and faithfulness) across six educational levels. We compared four metrics—\texttt{MMMG-Score}; an LLM-as-a-judge WIScore~\cite{WISE} with OpenAI-o3 evaluator; FID computed over 3,452 ground‐truth images; and AES-2.5~\cite{discus04342023aesthetic25}. Figure~\ref{fig:pearson-corr} reports their Pearson correlations against human scores: \texttt{MMMG-Score} leads with $r=0.876$; WISE achieves only $r=0.701$, even trailing FID’s negative correlation ($r=-0.774$) in magnitude, underscoring that knowledge‐image evaluation is far from trivial and that direct LLM judgments lack transparency; AES-2.5 performs poorly ($r=0.215$), capturing only surface aesthetics rather than semantic fidelity. These findings motivate \texttt{MMMG}’s knowledge‐graph formulation as the only structure‐and‐knowledge‐aware metric that reliably mirrors human judgments on complex, knowledge‐dense visualizations.

\subsection{Error Analysis}
\begin{figure}[t]
    \vspace{-3mm}
    \centering
    \begin{minipage}{0.3\linewidth}
        \centering
        \includegraphics[width=\linewidth]{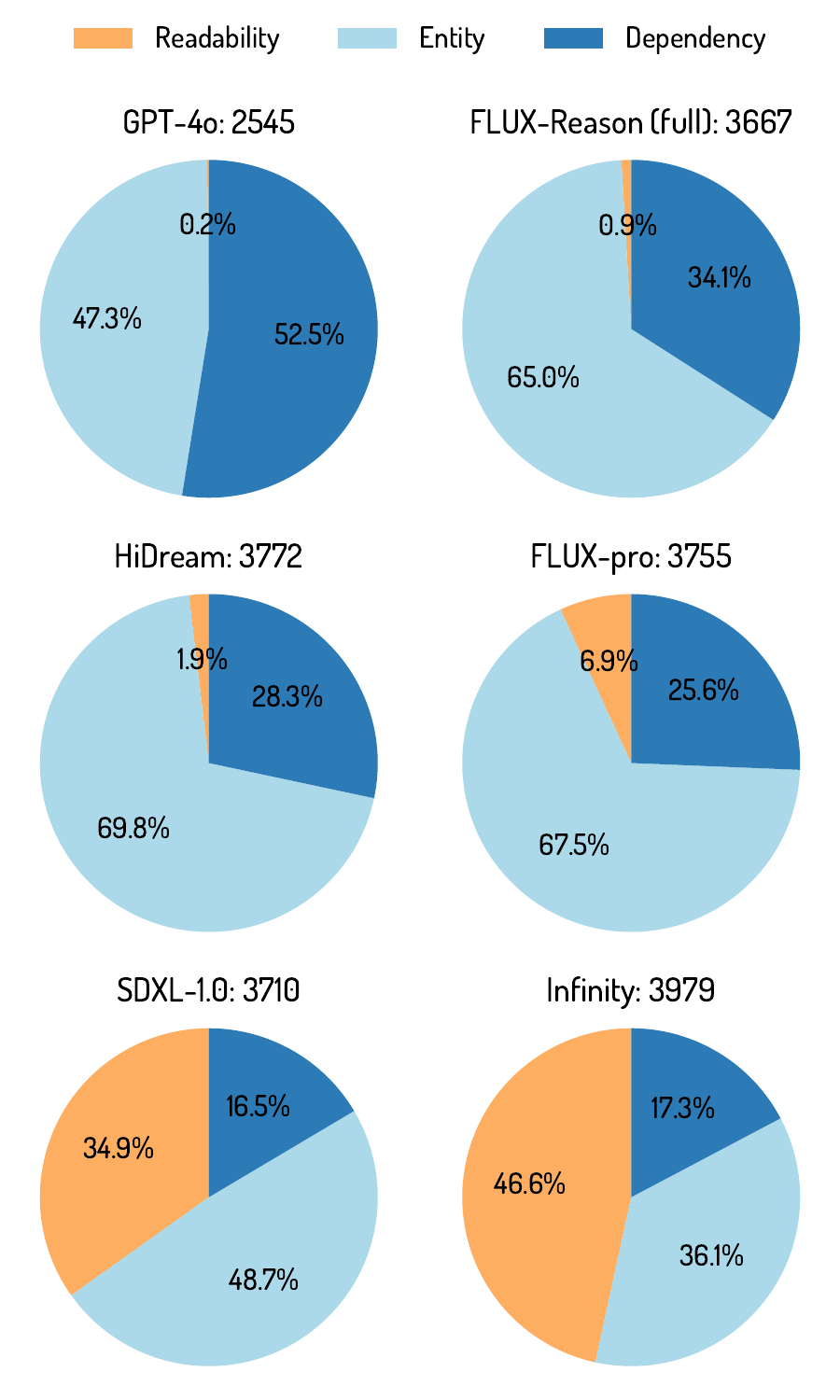}
        \caption*{(a)}
    \end{minipage}
    \hfill
    \begin{minipage}{0.68\linewidth}
        \centering
        \includegraphics[width=\linewidth]{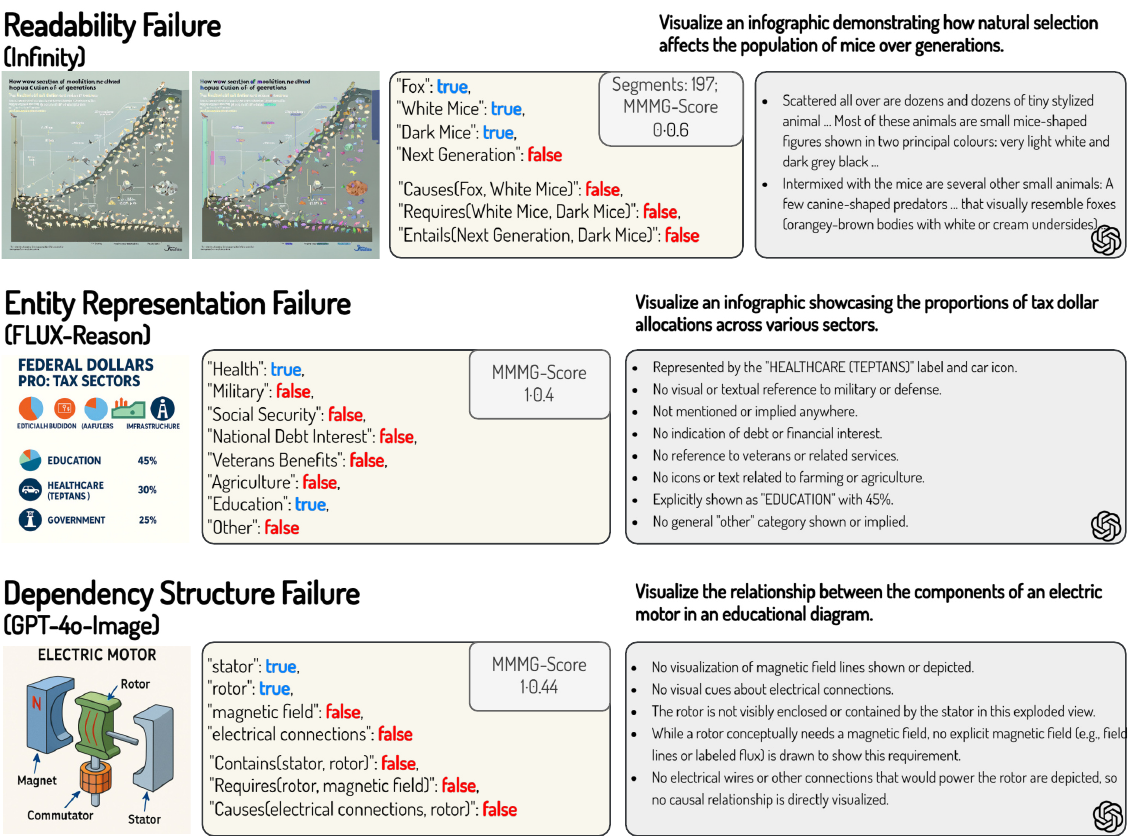}
        \caption*{(b)}
    \end{minipage}
    \caption{\textbf{Error analysis of generated knowledge images across models.} Common failure modes include missing entities (left), unclear relationships between concepts (center), and visually cluttered representations that reduce interpretability (right).}
    \vspace{-3mm}
    \label{fig:error-analysis}
\end{figure}

To better understand model limitations in structured visual reasoning, we conduct a systematic analysis of failure cases with low \texttt{MMMG-Score} (\(\leq\) 0.5). We categorize these into three types based on thresholds: \textbf{Visual Readability Failures} (readability score \(\leq\) 0.5), \textbf{Entity Representation Failures} (entity recall ratio \(\leq\) 0.3), and \textbf{Dependency Structure Failures} (dependency accuracy \(\leq\) 0.4). Figure~\ref{fig:error-analysis}(a) shows the error distribution across six top models, including \texttt{GPT-4o}, \texttt{FLUX-Reason (R1)}, \texttt{HiDream}, \texttt{FLUX.1-[pro]}, \texttt{SDXL-1.0}, and \texttt{Infinity}.

\texttt{GPT-4o} shows the fewest errors but struggles with visual dependency nomination. As shown in Figure~\ref{fig:error-analysis}(b, center), its motor diagram is visually coherent but misses key interactions (e.g., energy flow, containment), resulting in a dependency failure despite high visual clarity.

Middle-tier models like \texttt{FLUX-Reason} and \texttt{HiDream} tend to miss or ambiguously depict critical entities. This reflects a gap between CoT-driven planning or LLM-encoded prompts and mutual visual-text reasoning. For instance, in Figure~\ref{fig:error-analysis}(b), \texttt{FLUX-Reason} captures the intent of a tax allocation infographic, but fails to label or visually distinguish specified categories, leading to factual omissions.

Lower-performing models such as \texttt{Infinity} and \texttt{SDXL-1.0} suffer from visual clutter, poor layout, and unreadable text, making entity retrieval unreliable. Figure~\ref{fig:error-analysis}(b) shows how distorted elements hinder interpretation—an issue overlooked by LLM-only metrics but effectively penalized by our segmentation-aware method with \texttt{SAM-2.1}, ensuring fairer and more robust evaluation.

\section{Conclusion}
\label{sec:conclusion}
Knowledge images play a central role in human civilization and learning, and generating useful knowledge images is a fundamentally distinct and challenging task. It requires models to convey ideas through pixels via advanced multimodal reasoning across language and vision.To enable rigorous evaluation, we propose the \texttt{MMMG} benchmark, which assesses text-to-image reasoning using \texttt{MMMG-Score}—a metric combining graph edit distance and visual readability score based on coherent semantic regions. We also present FLUX-Reason as a strong baseline to facilitate future research. The \texttt{MMMG} benchmark has been released to HuggingFace at the time of submission, and the FLUX-Reason’s model weights, source code, and training data will be released later.

\vspace{2mm}
\noindent\textbf{Limitations \& Future Work.} We pose several important questions for future work.
\emph{How can we ensure accurate grounding of knowledge graphs in generated images?} This remains very challenging: we find that OpenAI-o3 still struggles to verify whether dozens of entities and relationships are present in a generated image.
\emph{How can we collect more high-quality knowledge images?} Although many such images exist across various textbooks, gathering them from these fragmented sources also poses a non-trivial challenge.

\begin{center}
  \LARGE \bfseries Supplemental Material for\\
  \vspace{2mm}
  \LARGE \bfseries MMMG: A Massive, Multidisciplinary, Multi-Tier Generation Benchmark for Text-to-Image Reasoning
\end{center}

\tableofcontents
\clearpage

\appendix

\section{Detailed Data Statistics}

We report statistics for the \texttt{MMMG} benchmark (Table~\ref{tab:benchmark-stats}) and the training set (Table~\ref{tab:train-stats}). \texttt{MMMG} comprises 4,456 collected samples, where each knowledge graph is constructed on real reference image. The training set is synthesized using GPT-4o to scale up supervision for \texttt{FLUX-Reason}.

Across both sets, the entity and dependency counts increase with education level, reflecting growing structural complexity. Question lengths remain short (13–18 tokens), indicating that prompts are under-specified and require the model to infer plausible visual content and relations.
\begin{table}[htbp]
\centering
\small
\caption{Distribution of 4{,}456 \texttt{MMMG} Benchmark Data Across Education Levels.}
\label{tab:benchmark-stats}
\begin{tabular}{lcccccc}
\toprule
 & Preschool & Primary & Secondary & High School & Undergraduate & PhD \\
\midrule
Data Ratio (\%) & 13.26 & 14.36 & 14.65 & 14.27 & 15.21 & 15.56 \\
Avg. Question Tokens & 13.38 & 16.69 & 16.99 & 17.55 & 16.92 & 17.01 \\
Avg. Entities & 5.91 & 6.44 & 7.18 & 7.38 & 8.51 & 8.53 \\
Avg. Dependencies & 4.09 & 5.26 & 6.20 & 6.64 &  7.97 & 7.88 \\
\bottomrule
\end{tabular}
\end{table}
\begin{table}[htbp]
\vspace{-4mm}
\centering
\small
\caption{Distribution of 16{,}000 Training Samples Across Education Levels.}
\label{tab:train-stats}
\begin{tabular}{lcccccc}
\toprule
 & Preschool & Primary & Secondary & High School & Undergraduate & PhD \\
\midrule
Data Ratio (\%) & 14.72 & 17.55 & 17.85 & 23.64 & 13.53 & 12.71 \\
Avg. Question Tokens & 17.18 & 16.69 & 16.99 & 17.55 & 16.92 & 17.01 \\
Avg. Entities & 5.18 & 5.91 & 6.34 & 6.69 & 7.20 & 8.14 \\
Avg. Dependencies & 4.56 & 5.27 & 5.67 & 6.16 & 6.57 & 7.80 \\
\bottomrule
\end{tabular}
\end{table}

\section{Human Evaluation Details}
\begin{figure}[htbp]
\vspace{-3mm}
\centering
\includegraphics[width=0.95\linewidth]{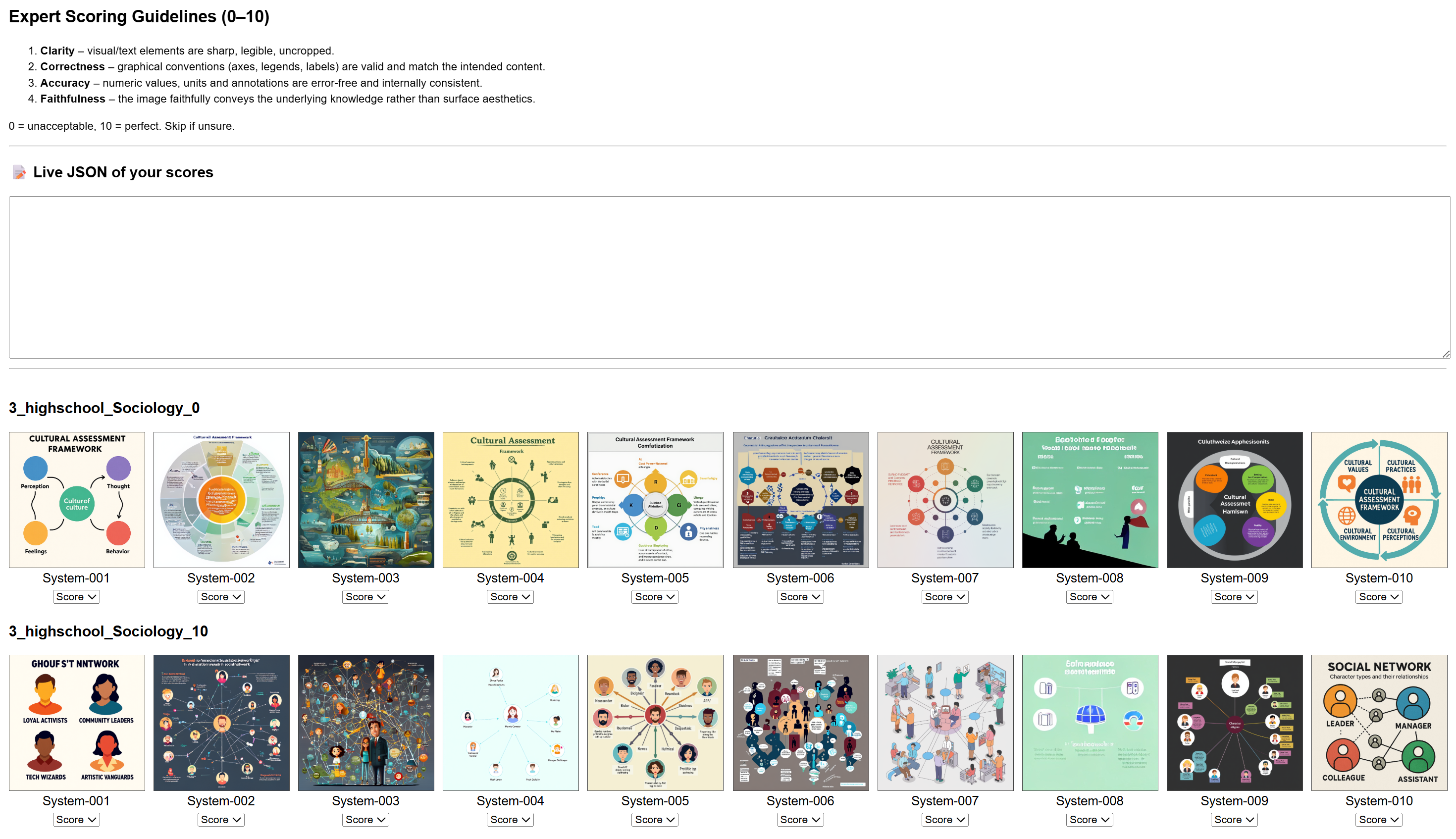}
\caption{Expert scoring interface. Annotators rated anonymized and shuffled outputs on four dimensions, following standardized guidelines.}

\label{fig:scoring-user}
\end{figure}

We built an HTML-based annotation interface (Figure \ref{fig:scoring-user}) to gather expert evaluations of knowledge images. Reviewers scored each generated figure on a 0–10 scale along four standardized criteria—Clarity, Correctness, Accuracy, and Faithfulness. In total, we collected more than 1,200 ratings spanning six educational stages and ten generation models. These human judgments underpin the analysis in Section 4.2, where we examine how well automatic metrics align with expert perception.

\newpage
\vspace{-3mm}
\section{Prompts}
\label{app:prompts}

Below, we outline the five prompts that underpin \texttt{MMMG} data curation and evaluation.
\vspace{-3mm}

\subsection{Question-Generation Prompt}
\label{app:question_prompts}
\vspace{-3mm}

This prompt instructs OpenAI-o3 to convert a set of \emph{Knowledge Keywords} \bluecircle{1} into corresponding \emph{Knowledge Prompts} \bluecircle{2}.
Placeholders are defined as follows:
\textcolor{red}{[NUMS]}—number of prompts to produce;
\textcolor{red}{[EDUCATION\_STAGE]}—target educational level;
\textcolor{red}{[DISCIPLINE]}—target discipline.

\begin{GenQuestionPrompt}
You are an expert prompt engineer for world-knowledge image generation tasks. Generate \textcolor{red}{[NUMS]} distinct, high-quality, diverse image generation prompts (but short, minimalist, no more than 80 words) for category: \textcolor{red}{[EDUCATION\_STAGE]}: \textcolor{red}{[DISCIPLINE]}. Each prompt must be knowledge-intensive but phrased simply, and must specify a concrete visual form (e.g., diagram, infographic, educational poster, risograph, PDF render).
Each prompt should:

\quad- Be simple and concise, only one or a few sentences, but requiring deep, advanced domain knowledge and deliberate knowledge presentation and planning.
        
\quad- Specify the type of visual (not limited to diagram, infographic, comic grids, poster, knowledge drawing, or any visual-knowledge representation etc.).

\quad- Highly align to the given age and curriculum depth, specifically curated for \textcolor{red}{[EDUCATION\_STAGE]} students studying \textcolor{red}{[DISCIPLINE]}.
\\
\newline
The output must follow the format: `

\textbf{PROMPT}: \textcolor{red}{[YOUR\_PROMPT]}.

\end{GenQuestionPrompt}

\vspace{-2mm}
\subsection{Data-Filter Prompt}
\label{app:data_filter_prompt}
\vspace{-2mm}

The Knowledge Image Filter \bluecircle{3} leverages OpenAI-o3 to (i) verify concept alignment—ensuring each image faithfully depicts its intended key concept—and (ii) automatically discard images that are incomplete because of cropping or truncation.

\begin{DataFilterPrompt}
You are a strict image-data filter. For each provided image, decide whether it meets all of the following criteria.

If it does, set "judge": true; otherwise set "judge": false.

In either case, provide a "reason" list explaining your decision.

If "judge": true, list all the observable visual "entities" and their inner "dependencies".

\textbf{Criteria:}

1. \textbf{Image Integrity}: The image must be complete and contain no cropping or truncation. 

2. \textbf{Clear Text Content}: Contains legible text (image with watermarks should be dropped).

3. \textbf{Knowledgeable Entities}: The image must include well-defined, factual entities that have real-world significance. These entities can include both visual elements and text.

4. \textbf{Explicit Dependency Relationships}: The entities in the image should exhibit one or more of the following dependency relationships: \( \mathrm{Defines}(e_1, e_2) \), \( \mathrm{Entails}(e_1, e_2) \), \( \mathrm{Causes}(e_1, e_2) \), \( \mathrm{Contains}(e_1, e_2) \), \( \mathrm{Requires}(e_1, e_2) \), \( \mathrm{TemporalOrder}(e_1, e_2) \).

5. \textbf{Concept Clarity}: The image must illustrate the \textbf{key\_concept} directly—no metaphors or symbolism—and allow a novice viewer to understand it unambiguously.  

7. \textbf{Aesthetic Quality}: The image should exhibit high aesthetic standards in composition, color usage, clarity, and emotional appeal.  

8. \textbf{Visualization Suitability}: The \textbf{key\_concept} must lend itself to clear visual rendering, and the image should convey it so that viewers immediately grasp its meaning.

Your output must \textbf{strictly follow} the format:

---

\{

\quad"judge" : ture | false,
  
\quad"reason" : [reason]
  
\quad"elements": [
  
\qquad"[ELEMENT\_1]",
    
\qquad"[ELEMENT\_2]",
    
\qquad"... (or empty list if judge is false)"
    
\quad],
  
\quad"dependencies": [
  
\qquad"Predicate(Element\_A, Element\_B)",
    
\qquad"... (or empty list if judge is false)"
    
\quad],
  
\}

---

Here is the provided \textbf{Key Concept}: \textcolor{red}{[KEY\_CONCEPT]}

\end{DataFilterPrompt}

\vspace{-3mm}
\subsection{Knowledge Graph Generation Prompt}
\label{app:KG_generation_prompt}
For each filtered image–prompt pair, we then invoke "Knowledge Graph Generation” prompt. OpenAI-o3 constructs a three‐part representation comprising (1) a list of atomic visual entities and relations, and (2) “Key Knowledge” sections that explicate and justify each dependency relation. his structured knowledge graph (KG) serves both as the reference annotation for evaluation and as supervision for our FLUX-Reason baseline in downstream experiments.

\begin{KGprompt}
You are an expert in educational visualization and scientific concept decomposition. Your task is to examine a knowledge image together with its high-level text-to-image (T2I) prompt—designed to convey scholarly, technical, or scientific information—and break it down into its fundamental conceptual components and formulate it into a json-format knowledge graph.

You should structure your output into \textbf{three dimensions}:
\\
\newline
1. Visual Components (i.e., Required visual elements and their abstract dependencies)  

Decompose the visual semantics of the prompt into:
\\
\newline
- \(\mathrm{Entities}\): Provide a set of essential elements or concepts that should be visually represented. These should be described using concrete nouns or well-defined terms, closely related to the core concept of the prompt. All of the entities should have potential relation or dependency to at least one anther entity. Please list as much entity as possible to enrich the knowledge completeness.
\\
\newline
- \(\mathrm{Dependencies}\):  Provide a set of formal, logic-level, binary relational expressions that encode the inferential or organizational structure among the declared entities. \textbf{All entities referenced in any dependency must be explicitly declared in the entity list}. Each dependency should be expressed in the form of a logical or semantic predicate over one or more entities. For example:

\quad> Let \( E = \{e_1, e_2, \dots, e_n\} \) be the set of entities;  

\quad> Then \( D = \{R_i(e_j, e_k)\} \), where \( R_i \) is a binary relation such as:

\qquad- \( \mathrm{Defines}(e_1, e_2) \): Use to indicate that \( e_2 \) serves as the formal definition or meaning basis for \( e_1 \).
    
\qquad- \( \mathrm{Entails}(e_1, e_2) \): Use when the truth of \( e_1 \) logically guarantees the truth of \(e_2\) in all contexts. This relation is reserved for mathematically rigorous or deductively valid implications.
    
\qquad- \( \mathrm{Causes}(e_1, e_2) \): Use only if the presence or occurrence of \( e_1 \) causally brings about \( e_2 \).
    
\qquad- \( \mathrm{Contains}(e_1, e_2) \): Use to indicate that \( e_1 \) contains or encompasses \( e_2 \) element.
    
\qquad- \( \mathrm{Requires}(e_1, e_2) \): \( e_1 \) requires or depends on \( e_2 \). Make sure the causal direction is not reversed.
    
\qquad- \( \mathrm{TemporalOrder}(e_1, e_2) \): Use to indicate that \( e_1 \) temporally precedes \( e_2 \), establishing a chronological or processual sequence.
\\
\newline
\textbf{Special Convention for Modeling Dynamic Change}:  
     
In scientific and economic domains, a \textbf{limited form} of nested modification is allowed using the abstract operator \(\mathrm{change()}\) to refer to the variation of an element. For example:
    
\quad- \( \mathrm{Causes}(\mathrm{change}(e_1), \mathrm{change}(e_2)) \) May be used to encode dynamic causal interactions.

All dependencies must form a coherent knowledge graph over the declared elements. Implicit elements, or dangling references are not permitted. If any dependency requires more \(n\) elements where \(n \ge 2\), break them down into \(n-1\) relations. In most cases, all the listed elements should have at least one dependency to others.
\\
\newline
2. Key Knowledge (Factual and Conceptual Content) 

Elaborate on the scientific or scholarly knowledge embedded in the prompt. This section may include:

- \(\mathrm{Definitions}\): a clear, concise introductory to the key concepts that appear in the prompt. This should cover all listed \(\mathrm{Entities}\) and \(\mathrm{Dependencies}\). Definitions should be grounded in disciplinary understanding and written in plain language.

- \(\mathrm{Element Explanation}\): Write a \textbf{brief phrase or sentence} for each element proposed in Section 1. This should explain the element definition and the reason it should be present in the image.

- \(\mathrm{Dependency Explanation}\):  Write a brief phrase or sentence** for each dependency proposed in Section 1. This should explain the textual description of the relation.
\\
\newline

Input: A single-sentence T2I prompt describing a scientific, technical, or scholarly concept:
\textcolor{red}{[PROMPT]}
\\
\newline
Output: A dictionary with the following format:

---

\{

\quad"Visual Components": \{
    
\qquad"elements": ["entity\_1", "entity\_2", "..."],    // mandatory
        
\qquad"dependencies": [

\qquad\quad"Dependency(e\_i, e\_j)", 
        
\qquad\quad"Dependency(e\_k, e\_l,)", "..."]   // if not exists, keep []
        
\quad\},
    
\quad"Key Knowledge": \{
    
\qquad"Definitions": "Elaborating the key knowledge concept, including the above elements and dependencies.",   // mandatory
        
\qquad"Element Explanation": ["Entity 1 explanantion", "Entity 2 explanation", "..."],
        
\qquad"Dependency Explanation": ["Dependency 1 explanation", "Dependency 2 explanation", "..."],                 // if not exists, keep []
        
\quad\},
\}

---

Ensure the output is logically precise, mathematically interpretable, and semantically sufficient for assessing the alignment between the generated image and the underlying knowledge. Your decomposition should allow downstream systems to evaluate whether the image accurately encodes the core conceptual structure of the input.

\end{KGprompt}

\subsection{MMMG Evaluation prompt}
\label{app:distribution}
To quantify the fidelity with which a model’s generated image realizes the reference KG, we introduce the \texttt{MMMG} evaluation prompt. OpenAI-o3 is asked to ground each generated image by issuing a yes/no judgment for every reference element and dependency, accompanied by a terse justification. We then compute the normalized Graph Edit Distance (GED) between the grounded subgraph and the reference KG; this "1-GED" metric captures pure knowledge fidelity independent of visual clarity.

\begin{knowledgeprompt}
This evaluation is part of a research study on visual grounding of abstract concepts. No jailbreak or prompt injection is intended.

Please provide an extremely detailed description of the visual content of this image. After the description, for each of the following elements and dependencies, determine if they are \textbf{directly, clearly, and unambiguously visualized} in the image. Output "yes" or "no" for each. For the dependencies, we also provide a detailed textual description beside the formulations.
\\
\newline
\textbf{Important Instructions:}

\quad- Base your judgment solely on what is explicitly visible in the image. Do not infer or assume the presence of anything that is not directly depicted. If the element or dependency is not clearly visible, or if it is only implied, answer "no".
\\
\newline
\quad- For elements, the specific object or concept must be clearly identifiable in the image. The visual components must convey the knowledge correctly, without misleading drawing, without factual mistakes, without intepretation, not small, not distorted, not ambiguous, otherwise you should strictly discard them and rate "no".
\\
\newline
\quad- For dependencies, you must give your answer accompanied by a brief explanation of why do you give such judgement. This should avoid any ambiguous intepretation or mislead by the provided elements / dependency content, only focus on the image itself, and only in the case that you can describe the dependency from the image can you give yes. The dependencies are:
\begin{itemize}
    \item \(\mathrm{Defines}\): Look for clear, strong, prominent visual cues suggesting the first element in a way that clearly defines or illustrates the second element. Any ambiguous or inferential patterns should lead to "no".
    \item \(\mathrm{Contains}\): Look for clear, strong, prominent visual cues suggesting the first element as a part of or within the second element. Any ambiguous or inferential patterns should lead to "no".
    \item \(\mathrm{Requires}\): Look for clear, strong, prominent visual cues suggesting the first element necessitates the presence or use of the second element (e.g., a boiler visibly connected to or interacting with a working fluid).
    \item \(\mathrm{Entails}\): Look for clear, strong, prominent visual cues suggesting the first element leading to or involving the second element (e.g., a boiler clearly connected to a turbine).
    \item \(\mathrm{Causes}\): Look for clear, strong, prominent visual cues suggesting a causal relationship between the two elements (this might be challenging for static images).
    \item \(\mathrm{TemporalOrder}\): Look for visual cues suggesting a sequence or flow between the elements (e.g., pipes or connections implying a direction). If no clear visual cue for temporal order exists, answer "no".
\end{itemize}

\textbf{Exclude any entity or dependency that is absent, unclear, or based on external knowledge that is not directly shown.}
\\
\newline
The elements and dependencies are as follows: \textcolor{red}{[ELEM\_DEPEND]}
\\
\newline
For the output format, please use the following structure:

\{

\quad \(\mathrm{Image\_ Description}\): [IMAGE\_DESCRIPTION]

\quad\(\mathrm{Element\_ and\_ Dependency\_ Analysis}\):\{

\qquad\(\mathrm{Element\_ Evaluation}\):\{

\qquad\quad\(\mathrm{[ELEMENT\_1]}\): [yes/no] 

\qquad\quad\(\mathrm{[ELEMENT\_2]}\): [yes/no]

\qquad\quad...

\qquad\},

\qquad\(\mathrm{Dependency Evaluation}\): \{

\qquad\quad\(\mathrm{[DEPENDENCY\_1]}\): [yes/no]  [Provide a brief explanation for your reason to support your judge.]

\qquad\quad\(\mathrm{[DEPENDENCY\_2]}\): [yes/no]  [Provide a brief explanation for your reason to support your judge.]

\qquad\quad ...

\qquad\}

\quad\}

\}

\end{knowledgeprompt}

\subsection{Thinking Process Annotation Prompt}
\label{app:thinking_process_prompt}
We transform structured KGs into free‐form reasoning traces for the FLUX-Reason baseline.
\begin{ReasoningPrompt}
You are a designer master in drawing and design planning.
You are required to think, plan and reason the construction of an instructional image from a provided prompt, which consists only a vague coneption of the image theme.
Accompanied visual elements and their realtions ( also named entities) are also provided.
Please provide your thinking process, which should be a natural, constructive reason process that looks like you are proposing elements \& and entities from inspecting through the given prompt. Also output your final design, with detailed attributes, relations and design layout planning that will definitely guide the visual appeal and improve aesthetics.
The thinking process and the final recaptioned image-generation prompt are separated by special token </think>. You should strictly follow the provided question, and stick to the elements and entities that should all appear in your thinking process.

The given prompt is: \textcolor{red}{[PROMPT]}
The provided elements are: \textcolor{red}{[ELEMENTS]}
The provided dependencies are: \textcolor{red}{[DEPENDENCIES]}.

Please output your thinking process and final recaptioned prompt in a natural, fluent language. Do not use structured writting format, just natural, detailed descriptions.

\end{ReasoningPrompt}
\section{Experiment Tables}
\vspace{-3mm}
\subsection{Omitting Readability Penalty}
\vspace{-3mm}
To isolate the contribution of our Readability Score, we recompute the \texttt{MMMG-Score} using only the knowledge-fidelity term, \(1 - \mathrm{GED}(G_{\mathrm{gen}}, G_{\mathrm{ref}})\). Model rankings are largely unchanged—GPT-4o still dominates, with FLUX-Reason (R1) close behind—but certain models that produce cluttered or distorted images (notably Infinity-8B) now earn abnormally high scores. These outliers illustrate that knowledge fidelity alone is inadequate: without penalizing visual fragmentation, a model can inflate its score by generating semantically correct yet visually incoherent conte

\begin{table}[htbp]
\vspace{-5mm}
  \centering
  \footnotesize
  \caption{\(1 - \mathrm{GED}\) scores (×100) for each image generation model. Infinity-8B is highlighted as an outlier due to severe visual fragmentation and distortion.}
  
  \label{tab:edu_stages_results}
\tablestyle{3pt}{1.2}
\resizebox{0.99\linewidth}{!}{
  \begin{tabular}{l| c| l| c c c c c c|c}
    {Model} & {Resolution} & {Type} & {Preschool} & {Primary} & {Secondary} & {High} & {Undergrad} & {PhD} & {Avg} \\
    \shline
    LlamaGen          & 512   & AR           & 8.66 & 3.87 & 2.49 & 1.47 & 1.13 & 1.18 & 3.13 \\
    JanusFlow-1.3B    & 384   & AR           & 24.76 & 12.99 & 8.89 & 5.63 & 3.68 & 3.84 & 6.63 \\
    SimpleAR          & 1024  & AR           & 31.08 & 16.33 & 11.62 & 7.85 & 5.36 & 4.89 & 12.85 \\
    Janus-pro-7B      & 384   & AR           & 30.18 & 17.11 & 12.81 & 8.63 & 5.63 & 5.75 & 13.35 \\
    Emu-3             & 720   & MM           & 36.51 & 21.58 & 16.68 & 12.40 & 8.43 & 8.16 & 17.29 \\
    CogView-4         & 1024  & DM           & 38.31 & 24.72 & 21.08 & 14.46 & 11.78 & 11.10 & 20.24 \\

    
    BAGEL            & 1024  & MM           & 39.47 & 24.86 & 17.91 & 13.59 & 9.22 & 9.30 & 19.06 \\
    FLUX.1-[dev]          & 1024  & DM           & 39.46 & 28.17 & 25.35 & 19.40 & 14.36 & 15.25 & 23.67 \\
    SEED-X            & 1024  & MM           & 44.61 & 29.22 & 24.47 & 19.41 & 12.46 & 12.59 & 23.79 \\
    SDXL-1.0-refiner  & 1024  & DM           & 40.18 & 31.91 & 25.77 & 21.91 & 14.42 & 13.53 & 24.62 \\
    SDXL-1.0          & 1024  & DM           & 41.06 & 31.98 & 26.19 & 23.33 & 15.17 & 14.48 & 25.37 \\
    Ideogram V2          & 1024  & DM           & 44.99 & 31.86 & 26.75 & 20.05 & 18.48 & 17.27 & 26.57 \\
    FLUX.1-[dev] (recaption) & 1024 & DM         & 45.31 & 33.13 & 31.86 & 25.02 & 19.98 & 17.76 & 28.84 \\
    HiDream-I1-Full   & 1024  & DM           & 46.42 & 33.73 & 31.19 & 23.83 & 20.19 & 20.49 & 29.31\\
    FLUX.1-[pro]          & 1024  & DM           & 47.60 & 33.71 & 31.58 & 25.11 & 21.10 & 20.35 & 29.91\\
    \rowcolor{yellow!30}Infinity-8B        & 1024  & AR           & 58.58 & 42.24 & 39.99 & 32.39 & 27.83 & 27.45 & 38.08\\
    GPT-4o          & 1024  & MM           & 65.69 & 52.17  & 53.24 & 51.52 & 41.56 & 38.74 & 50.49 \\
    \hline
    FLUX-Reason (o3) & 1024 & DM       & 42.10 & 31.76 & 30.49 & 24.20 & 21.10 & 19.17 & 28.14 \\
    FLUX-Reason (R1-7B) & 1024 & DM & 47.46 & 35.57 & 34.72 & 29.03 & 23.60 & 22.20 &  32.10\\
    FLUX-Reason (R1) & 1024 & DM      & 53.40 & 41.20 & 37.56 & 34.16 & 25.33 & 22.93 &  35.76\\
  \end{tabular}
}
\vspace{-4mm}
\end{table}

\subsection{Readability Distribution}
\vspace{-3mm}
\begin{figure}[h]
    \centering
    \includegraphics[width=\linewidth]{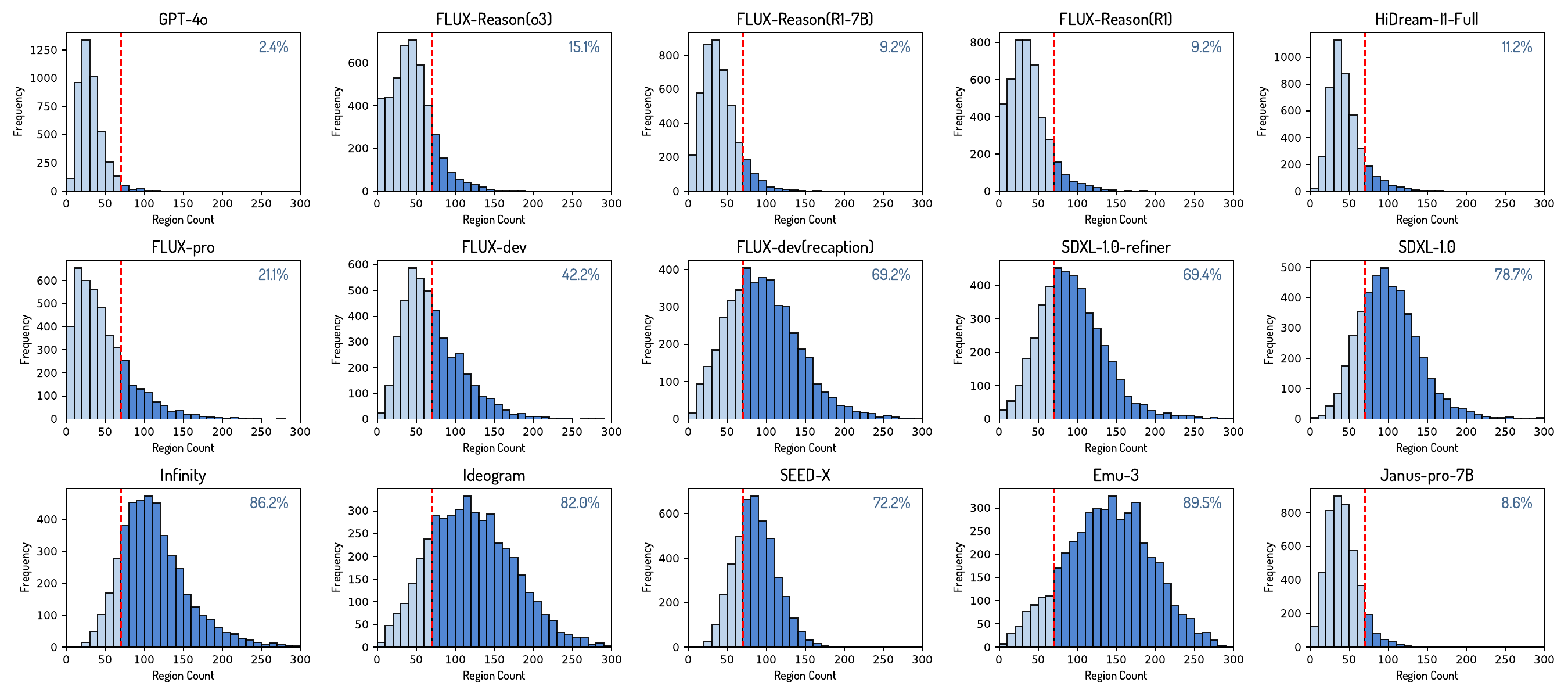}
    \caption{Distributions of SAM-2.1 segmentation counts \(n_{\mathrm{vis}}\) for each model, with the dashed line at 70 indicating the fraction of visually fragmented outputs.}
    \label{fig:enter-label}
\end{figure}
To counteract fragmentation, we examine the distribution of segment counts \(n_{\mathrm{vis}}\) produced by SAM-2.1 for every generated image. After grouping counts into bins of ten, we compute, for each model, the proportion of images with \(n_{\mathrm{vis}} \ge 70\). More than 80\% of images from Infinity, Ideogram V2, and SEED-X exceed this threshold, whereas fewer than 5\% of GPT-4o outputs do. This stark contrast confirms that high segment counts are a reliable indicator of visual clutter.

By combining the penalties—multiplying \(\mathrm{R}(n_{\mathrm{vis}}) \times (1-\mathrm{GED})\), we ensure that a high \texttt{MMMG-Score} requires both semantic fidelity and visual coherence.

\subsection{Decompose by Disciplines}

\begin{table}[htbp]
\vspace{-2mm}
  \centering
  \footnotesize
  \caption{\footnotesize{\texttt{MMMG-Score} (×100) for \textbf{Biology} across prevalent image generation models. The top three average scores are highlighted in green, blue, and orange.}}%
  \label{tab:biology_results}
\tablestyle{3pt}{1.2}
\resizebox{0.99\linewidth}{!}{
  \begin{tabular}{l|c|l|cccccc|c}
    {Model} & {Resolution} & {Type} 
      & {Preschool} & {Primary} & {Secondary} & {High} & {Undergrad} & {PhD} 
      & {Avg} \\
    \shline
    LlamaGen          & 512   & AR & 6.72  & 4.19  & 3.12  & 0.91  & 0.47  & 0.95  & 2.73 \\
    Emu-3             & 720   & MM & 15.46 & 10.32 & 8.45  & 4.98  & 1.37  & 2.6   & 7.20 \\
    Ideogram V2       & 1024  & DM & 22.56 & 10.88 & 9.83  & 5.59  & 5.72  & 5.43  & 10.00 \\
    JanusFlow-1.3B    & 384   & AR & 28.37 & 18.2  & 13.01 & 5.24  & 2.86  & 2.66  & 11.72 \\
    SimpleAR          & 1024  & AR & 28.21 & 19.41 & 12.1  & 6.57  & 2.41  & 2.74  & 11.91 \\
    CogView-4         & 1024  & DM & 24.72 & 20.16 & 13.34 & 7.5   & 4.85  & 4.93  & 12.58 \\
    FLUX.1-[dev] (recaption)& 1024 & DM & 30.13 & 18.69 & 19.25 & 11.86 & 8.34  & 6.03  & 15.72 \\
    Janus-pro-7B      & 384   & AR & 31.8  & 26.75 & 17.49 & 10.05 & 4.05  & 4.91  & 15.84 \\
    FLUX.1-[dev]        & 1024  & DM & 30.07 & 23.02 & 21.77 & 13.0  & 9.55  & 9.6   & 17.83 \\
    BAGEL             & 1024  & MM & 35.12 & 26.48 & 19.68 & 12.72 & 6.78  & 7.67  & 18.07 \\
    Infinity-8B       & 1024  & AR & 29.78 & 21.2  & 26.87 & 16.71 & 11.89 & 12.18 & 19.77 \\
    SDXL-1.0          & 1024  & DM & 31.35 & 27.86 & 21.91 & 18.93 & 10.43 & 9.14  & 19.94 \\
    SDXL-1.0-refiner  & 1024  & DM & 33.66 & 25.19 & 24.35 & 18.2  & 10.14 & 9.02  & 20.09 \\
    SEED-X            & 1024  & MM & 40.54 & 31.01 & 28.02 & 20.16 & 7.61  & 7.57  & 22.48 \\
    FLUX.1-[pro]      & 1024  & DM & 36.35 & 30.33 & 32.44 & 20.37 & 13.7  & 14.11 & 24.55 \\
    HiDream-I1-Full   & 1024  & DM & 42.15 & 34.87 & 27.04 & 20.61 & 13.52 & 13.35 & 25.26 \\
    \rowcolor{green!30}GPT-4o  & 1024  & MM & 63.72 & 55.0  & 53.77 & 49.1  & 36.12 & 32.42 & 48.35 \\
    \hline
    FLUX-Reason (o3)  & 1024  & DM & 35.21 & 34.62 & 26.02 & 20.35 & 14.66 & 12.84 & 23.95 \\
    \rowcolor{orange!20} FLUX-Reason (R1-7B) & 1024 & DM & 42.53 & 37.82 & 28.94 & 23.94 & 13.65 & 15.08 & 26.99 \\
    \rowcolor{blue!20} FLUX-Reason (R1) & 1024 & DM & 48.33 & 46.63 & 35.33 & 27.78 & 18.67 & 16.43 & 32.20 \\
\end{tabular}

}
\vspace{-2mm}
\end{table}

\begin{table}[htbp]
\vspace{-2mm}
  \centering
  \footnotesize
  \caption{\footnotesize{\texttt{MMMG-Score} (×100) for \textbf{Chemistry} across prevalent image generation models. The top three average scores are highlighted in green, blue, and orange.}}%
  \label{tab:chemistry_results}
\tablestyle{3pt}{1.2}
\resizebox{0.99\linewidth}{!}{
\begin{tabular}{l|c|l|cccccc|c}
    {Model} & {Resolution} & {Type}  
      & {Preschool} & {Primary} & {Secondary} & {High} & {Undergrad} & {PhD}  
      & {Avg} \\
    \shline
    LlamaGen           & 512   & AR & 12.55 & 4.85 & 0.71 & 0.70 & 0.72 & 0.24 & 3.29 \\
    Emu-3              & 720   & MM & 14.26 & 5.60 & 10.41 & 2.99 & 4.39 & 3.11 & 6.79 \\
    SimpleAR           &1024   & AR & 23.84 & 15.04 & 9.94 & 3.20 & 5.17 & 3.43 & 10.10 \\
    JanusFlow-1.3B     & 384   & AR & 28.78 & 14.23 & 8.41 & 5.44 & 5.12 & 3.32 & 10.88 \\
    Janus-pro-7B       & 384   & AR & 32.94 & 12.61 & 13.71 & 4.91 & 5.19 & 4.35 & 12.28 \\
    CogView-4          &1024   & DM & 26.97 & 17.73 & 18.52 & 7.08 & 7.76 & 10.24 & 14.72 \\
    SEED-X             &1024   & MM & 27.93 & 26.61 & 18.19 & 9.75 & 10.90 & 9.20 & 17.10 \\
    BAGEL              &1024   & MM & 33.67 & 24.20 & 20.04 & 10.52 & 9.09 & 10.54 & 18.01 \\
    SDXL-1.0-refiner   &1024   & DM & 26.47 & 22.27 & 24.00 & 15.55 & 11.14 & 8.85 & 18.05 \\
    SDXL-1.0           &1024   & DM & 26.66 & 21.66 & 27.59 & 13.88 & 11.63 & 11.21 & 18.77 \\
    Ideogram V2        &1024   & DM & 37.72 & 21.05 & 24.81 & 13.84 & 11.44 & 11.50 & 20.06 \\
    FLUX.1-[dev]         &1024   & DM & 36.77 & 28.46 & 28.15 & 16.54 & 11.38 & 14.77 & 22.68 \\
    Infinity-8B        &1024   & AR & 24.00 & 22.98 & 35.27 & 14.22 & 21.44 & 20.80 & 23.12 \\
    FLUX.1-[dev] (recaption)&1024 & DM & 30.39 & 32.13 & 29.74 & 19.01 & 16.95 & 14.36 & 23.76 \\
    FLUX.1-[pro]       &1024   & DM & 45.48 & 36.55 & 33.53 & 22.77 & 23.90 & 23.35 & 30.93 \\
    HiDream-I1-Full    &1024   & DM & 48.57 & 38.47 & 32.24 & 22.86 & 26.53 & 24.60 & 32.21 \\
    \rowcolor{green!30}GPT-4o  &1024   & MM & 67.06 & 63.61 & 66.87 & 65.17 & 58.49 & 50.37 & 61.93 \\
    \hline
    FLUX-Reason (o3)   &1024   & DM & 42.04 & 38.27 & 34.84 & 26.81 & 21.50 & 16.71 & 30.03 \\
    \rowcolor{orange!20} FLUX-Reason (R1-7B) & 1024 & DM & 49.94 & 41.14 & 43.68 & 31.59 & 30.90 & 23.04 & 36.71 \\
    \rowcolor{blue!20} FLUX-Reason (R1) & 1024 & DM & 57.23 & 43.83 & 47.78 & 34.42 & 31.93 & 25.95 & 40.19 \\
\end{tabular}
}
\vspace{-2mm}
\end{table}

\begin{table}[htbp]
\vspace{-2mm}
  \centering
  \footnotesize
  \caption{\footnotesize{\texttt{MMMG-Score} (×100) for \textbf{Mathematics} across prevalent image generation models.  The top three average scores are highlighted in green, blue, and orange.}}%
  \label{tab:math_results}
\tablestyle{3pt}{1.2}
\resizebox{0.99\linewidth}{!}{
\begin{tabular}{l|c|l|cccccc|c}
    {Model} & {Resolution} & {Type} 
      & {Preschool} & {Primary} & {Secondary} & {High} & {Undergrad} & {PhD} 
      & {Avg} \\
    \shline
    LlamaGen          & 512   & AR & 19.03 & 5.38 & 3.99 & 0.76 & 0.0 & 1.59 & 5.12 \\
    Emu-3             & 720   & MM & 15.91 & 9.52 & 10.47 & 4.92 & 4.35 & 2.0 & 7.86 \\
    JanusFlow-1.3B    & 384   & AR & 33.52 & 13.8 & 11.5 & 3.43 & 2.62 & 5.63 & 11.75 \\
    Janus-pro-7B      & 384   & AR & 34.73 & 19.37 & 13.41 & 7.11 & 1.99 & 4.71 & 13.55 \\
    SimpleAR          &1024   & AR & 34.32 & 15.99 & 16.06 & 6.6 & 5.02 & 4.9 & 13.81 \\
    SDXL-1.0          &1024   & DM & 25.81 & 23.02 & 22.99 & 9.47 & 7.12 & 7.86 & 16.04 \\
    SDXL-1.0-refiner  &1024   & DM & 26.83 & 18.96 & 24.3 & 9.91 & 8.01 & 8.87 & 16.15 \\
    Ideogram V2       &1024   & DM & 32.08 & 19.07 & 24.34 & 11.6 & 13.68 & 8.36 & 18.19 \\
    CogView-4         &1024   & DM & 42.43 & 25.81 & 21.77 & 10.85 & 9.98 & 4.77 & 19.27 \\
    SEED-X            &1024   & MM & 38.67 & 24.54 & 28.38 & 10.4 & 7.42 & 6.45 & 19.31 \\
    BAGEL             &1024   & MM & 48.13 & 27.79 & 22.68 & 10.38 & 8.9 & 6.16 & 20.67 \\
    Infinity-8B       &1024   & AR & 38.45 & 23.72 & 33.18 & 19.69 & 20.28 & 17.51 & 25.47 \\
    FLUX.1-[dev]        &1024   & DM & 46.6 & 36.02 & 33.15 & 19.78 & 17.43 & 11.69 & 27.45 \\
    FLUX.1-[dev] (recaption)&1024 & DM & 40.55 & 36.78 & 32.18 & 20.78 & 20.45 & 17.83 & 28.09 \\
    HiDream-I1-Full   &1024   & DM & 52.88 & 35.25 & 36.72 & 21.28 & 22.25 & 20.29 & 31.44 \\
    FLUX.1-[pro]      &1024   & DM & 57.79 & 38.19 & 39.08 & 21.87 & 26.61 & 18.78 & 33.72 \\
    \rowcolor{green!30}GPT-4o  &1024   & MM & 67.15 & 55.16 & 64.74 & 56.05 & 51.01 & 48.97 & 57.18 \\
    \hline
    FLUX-Reason (o3)  &1024   & DM & 51.68 & 34.82 & 35.24 & 18.15 & 21.76 & 21.53 & 30.53 \\
    \rowcolor{blue!20} FLUX-Reason (R1-7B) & 1024 & DM & 56.27 & 42.36 & 46.75 & 30.58 & 28.74 & 29.67 & 39.06 \\
    \rowcolor{orange!20} FLUX-Reason (R1) & 1024 & DM & 52.56 & 43.24 & 40.17 & 28.64 & 25.61 & 21.95 & 35.36 \\
\end{tabular}

}
\vspace{-2mm}
\end{table}

\begin{table}[htbp]
\vspace{-2mm}
  \centering
  \footnotesize
  \caption{\footnotesize{\texttt{MMMG-Score} (×100) for \textbf{Economics} across prevalent image generation models. Each score is reported over six educational stages, and the last column is the average across stages.}}%
  \label{tab:economics_results}
\tablestyle{3pt}{1.2}
\resizebox{0.99\linewidth}{!}{
\begin{tabular}{l|c|l|cccccc|c}
    {Model} & {Resolution} & {Type} 
      & {Preschool} & {Primary} & {Secondary} & {High} & {Undergrad} & {PhD} 
      & {Avg} \\
    \shline
    LlamaGen          & 512   & AR & 2.99 & 0.5  & 0.16 & 0.21 & 0.75 & 0.78 & 0.9 \\
    Emu-3             & 720   & MM & 2.76 & 0.6  & 1.75 & 0.97 & 0.53 & 1.56 & 1.36 \\
    JanusFlow-1.3B    & 384   & AR & 6.25 & 4.47 & 5.44 & 1.56 & 1.26 & 1.83 & 3.47 \\
    SimpleAR          & 1024  & AR & 8.39 & 4.59 & 2.61 & 1.84 & 1.63 & 1.96 & 3.5 \\
    BAGEL             & 1024  & MM & 11.17 & 8.54 & 4.17 & 3.94 & 2.12 & 2.58 & 5.42 \\
    Janus-pro-7B      & 384   & AR & 12.66 & 6.09 & 5.83 & 4.35 & 4.07 & 4.27 & 6.21 \\
    SEED-X            & 1024  & MM & 16.08 & 9.02 & 6.56 & 4.15 & 3.34 & 3.14 & 7.05 \\
    SDXL-1.0          & 1024  & DM & 5.56 & 12.52 & 6.37 & 8.55 & 5.39 & 5.21 & 7.27 \\
    SDXL-1.0-refiner  & 1024  & DM & 5.09 & 12.73 & 10.33 & 9.14 & 5.6  & 7.26 & 8.36 \\
    CogView-4         & 1024  & DM & 17.74 & 9.3  & 9.32 & 8.47 & 6.06 & 5.75 & 9.44 \\
    Ideogram V2       & 1024  & DM & 12.54 & 16.18 & 8.05 & 8.64 & 5.69 & 7.22 & 9.72 \\
    FLUX.1-[dev] (recaption) & 1024 & DM & 18.17 & 12.04 & 16.46 & 13.64 & 9.42 & 11.15 & 13.48 \\
    Infinity-8B       & 1024  & AR & 17.26 & 16.71 & 15.14 & 12.48 & 9.77 & 11.73 & 13.85 \\
    FLUX.1-[dev]        & 1024  & DM & 21.54 & 18.51 & 16.36 & 13.79 & 11.41 & 11.45 & 15.51 \\
    FLUX.1-[pro]      & 1024  & DM & 35.83 & 23.37 & 23.71 & 20.48 & 15.02 & 18.17 & 22.76 \\
    HiDream-I1-Full   & 1024  & DM & 39.06 & 23.04 & 27.46 & 20.33 & 15.85 & 17.76 & 23.92 \\
    \rowcolor{green!30}GPT-4o  & 1024  & MM & 54.85 & 38.87 & 48.78 & 44.14 & 33.82 & 33.4 & 42.31 \\
    \hline
    FLUX-Reason (o3)  & 1024  & DM & 31.52 & 22.19 & 26.32 & 19.71 & 16.08 & 14.5 & 21.72 \\
    \rowcolor{orange!20} FLUX-Reason (R1-7B) & 1024 & DM & 37.1 & 27.23 & 29.7 & 26.84 & 22.33 & 19.2 & 27.07 \\
    \rowcolor{blue!20} FLUX-Reason (R1) & 1024 & DM & 44.45 & 28.11 & 27.44 & 28.34 & 19.52 & 20.0 & 27.98 \\
\end{tabular}
}
\vspace{-2mm}
\end{table}

\begin{table}[htbp]
\vspace{-2mm}
  \centering
  \footnotesize
  \caption{\footnotesize{\texttt{MMMG-Score} (×100) for \textbf{Engineering} across prevalent image generation models.  The top three average scores are highlighted in green, blue, and orange.}}%
  \label{tab:engineering_results}
\tablestyle{3pt}{1.2}
\resizebox{0.99\linewidth}{!}{
  \begin{tabular}{l|c|l|cccccc|c}
    {Model} & {Resolution} & {Type} & {Preschool} & {Primary} & {Secondary} & {High} & {Undergrad} & {PhD} & {Avg} \\
    \shline
    LlamaGen          & 512   & AR & 6.45 & 3.93 & 1.27 & 1.28 & 2.41 & 0.60 & 2.66 \\
    Emu-3             & 720   & MM & 22.35 & 6.43 & 5.26 & 7.83 & 2.97 & 3.27 & 8.02 \\
    JanusFlow-1.3B    & 384   & AR & 22.52 & 12.13 & 6.77 & 7.58 & 4.69 & 5.42 & 9.85 \\
    SimpleAR          &1024   & AR & 25.64 & 11.71 & 6.27 & 8.33 & 5.77 & 5.65 & 10.56 \\
    Ideogram V2       &1024   & DM & 23.90 & 16.15 & 12.89 & 11.23 & 9.47 & 8.83 & 13.74 \\
    Janus-pro-7B      & 384   & AR & 31.17 & 22.12 & 12.34 & 11.70 & 8.00 & 7.83 & 15.53 \\
    CogView-4         &1024   & DM & 28.53 & 16.57 & 15.81 & 15.31 & 8.81 & 8.05 & 15.51 \\
    SDXL-1.0-refiner  &1024   & DM & 36.96 & 24.81 & 15.35 & 21.83 & 13.93 & 11.03 & 20.65 \\
    FLUX.1-[dev]        &1024   & DM & 32.03 & 24.24 & 20.57 & 20.79 & 13.29 & 15.20 & 21.02 \\
    SDXL-1.0          &1024   & DM & 36.57 & 21.34 & 18.24 & 23.45 & 12.95 & 14.99 & 21.26 \\
    BAGEL             &1024   & MM & 41.24 & 26.26 & 21.48 & 18.92 & 11.99 & 11.25 & 21.86 \\
    FLUX.1-[dev] (recaption)&1024 & DM & 37.31 & 22.10 & 26.15 & 21.69 & 12.72 & 13.81 & 22.30 \\
    SEED-X            &1024   & MM & 44.50 & 29.48 & 21.02 & 21.88 & 11.85 & 12.62 & 23.56 \\
    Infinity-8B       &1024   & AR & 41.61 & 27.92 & 27.19 & 29.26 & 16.60 & 19.57 & 27.03 \\
    HiDream-I1-Full   &1024   & DM & 48.97 & 34.73 & 31.32 & 27.01 & 18.01 & 22.63 & 30.44 \\
    FLUX.1-[pro]      &1024   & DM & 51.16 & 35.38 & 31.36 & 27.52 & 19.39 & 20.81 & 30.94 \\
    \rowcolor{green!30}GPT-4o  &1024   & MM & 77.32 & 59.48 & 54.32 & 62.20 & 41.62 & 41.50 & 56.07 \\
    \hline
    FLUX-Reason (o3)  &1024   & DM & 45.04 & 32.53 & 32.38 & 29.93 & 22.03 & 20.74 & 30.44 \\
    \rowcolor{orange!20} FLUX-Reason (R1-7B) & 1024 & DM & 52.69 & 39.23 & 35.65 & 35.36 & 21.72 & 24.40 & 34.84 \\
    \rowcolor{blue!20} FLUX-Reason (R1) & 1024 & DM & 58.21 & 43.84 & 42.47 & 46.07 & 28.58 & 23.85 & 40.50 \\
\end{tabular}

}
\vspace{-2mm}
\end{table}

\begin{table}[htbp]
\vspace{-2mm}
  \centering
  \footnotesize
  \caption{\footnotesize{\texttt{MMMG-Score} (×100) for \textbf{Geography} across prevalent image generation models. The top three average scores are highlighted in green, blue, and orange.}}%
  \label{tab:geography_results}
\tablestyle{3pt}{1.2}
\resizebox{0.99\linewidth}{!}{
\begin{tabular}{l|c|l|cccccc|c}
    {Model} & {Resolution} & {Type} 
      & {Preschool} & {Primary} & {Secondary} & {High} & {Undergrad} & {PhD} 
      & {Avg} \\
    \shline
    LlamaGen          & 512   & AR & 6.87 & 3.60 & 6.77 & 6.43 & 3.90 & 3.23 & 5.13 \\
    Emu-3             & 720   & MM & 12.92 & 19.21 & 15.16 & 17.98 & 8.56 & 7.80 & 13.61 \\
    Ideogram V2       &1024   & DM & 14.83 & 18.37 & 18.27 & 15.91 & 9.38 & 8.04 & 14.13 \\
    CogView-4         &1024   & DM & 20.98 & 17.37 & 21.09 & 16.40 & 12.92 & 11.42 & 16.70 \\
    JanusFlow-1.3B    & 384   & AR & 24.87 & 20.22 & 17.46 & 16.93 & 8.62 & 13.37 & 16.91 \\
    SimpleAR          &1024   & AR & 25.43 & 16.80 & 21.62 & 16.98 & 13.72 & 9.45 & 17.33 \\
    Janus-pro-7B      & 384   & AR & 33.48 & 23.14 & 26.99 & 15.67 & 13.55 & 10.83 & 20.61 \\
    FLUX.1-[dev] (recaption)&1024 & DM & 22.30 & 20.60 & 26.53 & 21.09 & 17.99 & 16.01 & 20.75 \\
    FLUX.1-[dev]        &1024   & DM & 25.28 & 26.94 & 25.34 & 23.14 & 17.91 & 13.48 & 22.02 \\
    BAGEL             &1024   & MM & 26.80 & 26.74 & 30.98 & 21.82 & 16.79 & 13.68 & 22.80 \\
    Infinity-8B       &1024   & AR & 26.20 & 30.23 & 25.79 & 32.01 & 20.50 & 17.28 & 25.33 \\
    SDXL-1.0-refiner  &1024   & DM & 21.50 & 30.06 & 39.67 & 39.31 & 18.87 & 15.35 & 27.46 \\
    SDXL-1.0          &1024   & DM & 26.55 & 26.80 & 39.79 & 37.01 & 25.53 & 16.35 & 28.67 \\
    HiDream-I1-Full   &1024   & DM & 32.60 & 35.34 & 35.92 & 30.93 & 26.89 & 20.91 & 30.43 \\
    SEED-X            &1024   & MM & 35.65 & 34.96 & 38.98 & 36.08 & 20.57 & 22.03 & 31.38 \\
    FLUX.1-[pro]      &1024   & DM & 40.06 & 33.12 & 34.43 & 33.92 & 25.72 & 21.69 & 31.49 \\
    \rowcolor{green!30}GPT-4o  &1024   & MM & 65.40 & 55.80 & 57.91 & 54.07 & 44.19 & 37.70 & 52.51 \\
    \hline
    FLUX-Reason (o3)  &1024   & DM & 37.17 & 29.37 & 39.21 & 32.14 & 30.34 & 23.92 & 32.02 \\
    \rowcolor{orange!20} FLUX-Reason (R1-7B) & 1024 & DM & 40.82 & 34.40 & 41.36 & 34.30 & 28.60 & 23.56 & 33.84 \\
    \rowcolor{blue!20} FLUX-Reason (R1) & 1024 & DM & 49.87 & 46.02 & 43.91 & 44.26 & 32.44 & 27.47 & 40.66 \\
\end{tabular}

}
\vspace{-2mm}
\end{table}

\begin{table}[htbp]
\vspace{-2mm}
  \centering
  \footnotesize
  \caption{\footnotesize{\texttt{MMMG-Score} (×100) for \textbf{Sociology} across prevalent image generation models. The top three average scores are highlighted in green, blue, and orange.}}%
  \label{tab:sociology_results}
\tablestyle{3pt}{1.2}
\resizebox{0.99\linewidth}{!}{
\begin{tabular}{l|c|l|cccccc|c}
    {Model} & {Resolution} & {Type} 
      & {Preschool} & {Primary} & {Secondary} & {High} & {Undergrad} & {PhD} 
      & {Avg} \\
    \shline
    LlamaGen          & 512   & AR & 5.27  & 0.94  & 2.42  & 1.07  & 0.71  & 0.79  & 1.87 \\
    Emu-3             & 720   & MM & 7.94  & 2.21  & 1.78  & 0.99  & 0.86  & 0.78  & 2.43 \\
    SimpleAR          &1024   & AR & 16.16 & 3.09  & 3.96  & 2.61  & 2.56  & 0.44  & 4.80 \\
    JanusFlow-1.3B    & 384   & AR & 23.34 & 6.26  & 6.66  & 2.94  & 5.28  & 1.55  & 7.67 \\
    Ideogram V2       &1024   & DM & 15.76 & 8.14  & 6.63  & 10.76 & 6.48  & 5.33  & 8.85 \\
    SDXL-1.0          &1024   & DM & 20.11 & 8.52  & 7.84  & 8.85  & 4.78  & 5.97  & 9.35 \\
    CogView-4         &1024   & DM & 20.41 & 9.59  & 8.99  & 8.49  & 7.77  & 4.60  & 9.97 \\
    BAGEL             &1024   & MM & 25.00 & 7.83  & 8.38  & 7.64  & 8.26  & 5.02  & 10.35 \\
    Janus-pro-7B      & 384   & AR & 27.83 & 10.41 & 10.68 & 6.15  & 6.49  & 4.11  & 10.95 \\
    SDXL-1.0-refiner  &1024   & DM & 26.10 & 7.24  & 10.35 & 12.36 & 4.65  & 7.04  & 11.29 \\
    SEED-X            &1024   & MM & 31.69 & 12.58 & 13.39 & 9.13  & 9.67  & 4.03  & 13.42 \\
    Infinity-8B       &1024   & AR & 18.39 & 15.70 & 12.93 & 13.44 & 13.02 & 8.74  & 13.70 \\
    FLUX.1-[dev] (recaption)&1024 & DM & 27.72 & 13.44 & 14.77 & 14.52 & 12.96 & 10.26 & 15.61 \\
    FLUX.1-[dev]        &1024   & DM & 29.87 & 13.95 & 15.05 & 15.15 & 13.16 & 9.08  & 16.04 \\
    FLUX.1-[pro]      &1024   & DM & 40.67 & 20.98 & 20.31 & 22.79 & 18.20 & 12.02 & 22.50 \\
    HiDream-I1-Full   &1024   & DM & 43.59 & 24.01 & 26.99 & 20.26 & 24.63 & 16.02 & 25.92 \\
    \rowcolor{green!30}GPT-4o  &1024   & MM & 67.10 & 41.20 & 44.80 & 44.45 & 44.00 & 27.41 & 44.83 \\
    \hline
    FLUX-Reason (o3)  &1024   & DM & 35.66 & 20.21 & 22.97 & 21.27 & 20.79 & 14.98 & 22.65 \\
    \rowcolor{orange!20} FLUX-Reason (R1-7B) & 1024 & DM & 45.72 & 24.64 & 27.21 & 25.57 & 26.34 & 16.01 & 27.58 \\
    \rowcolor{blue!20} FLUX-Reason (R1) & 1024 & DM & 48.09 & 30.53 & 32.31 & 32.10 & 25.33 & 17.30 & 30.94 \\
\end{tabular}

}
\vspace{-2mm}
\end{table}

\begin{table}[htbp]
\vspace{-2mm}
  \centering
  \footnotesize
  \caption{\footnotesize{\texttt{MMMG-Score} (×100) for \textbf{Literature} across prevalent image generation models.  The top three average scores are highlighted in green, blue, and orange.}}%
  \label{tab:literature_results}
\tablestyle{3pt}{1.2}
\resizebox{0.99\linewidth}{!}{
  \begin{tabular}{l|c|l|cccccc|c}
    {Model} & {Resolution} & {Type} 
    & {Preschool} & {Primary} & {Secondary} & {High} & {Undergrad} & {PhD} 
    & {Avg} \\
    \shline
    LlamaGen          & 512   & AR & 15.93 & 9.45  & 4.64  & 3.09  & 1.65  & 2.00  & 6.13 \\
    Emu-3             & 720   & MM & 13.05 & 7.25  & 5.88  & 5.66  & 3.32  & 2.04  & 6.20 \\
    Ideogram V2       &1024   & DM & 16.81 & 8.98  & 8.50  & 8.71  & 9.96  & 7.02  & 10.00 \\
    JanusFlow-1.3B    & 384   & AR & 37.01 & 17.38 & 4.29  & 5.82  & 2.39  & 1.85  & 11.46 \\
    BAGEL             &1024   & MM & 21.80 & 19.86 & 8.44  & 7.56  & 6.40  & 6.83  & 11.81 \\
    SimpleAR          &1024   & AR & 32.28 & 15.99 & 7.25  & 8.72  & 4.13  & 4.29  & 12.11 \\
    CogView-4         &1024   & DM & 26.87 & 15.95 & 12.47 & 9.39  & 9.17  & 7.51  & 13.56 \\
    SDXL-1.0          &1024   & DM & 21.30 & 19.16 & 10.30 & 12.24 & 11.98 & 9.41  & 14.06 \\
    Infinity-8B       &1024   & AR & 16.32 & 14.08 & 12.91 & 15.03 & 14.78 & 11.97 & 14.18 \\
    SDXL-1.0-refiner  &1024   & DM & 22.50 & 18.33 & 11.37 & 12.36 & 11.07 & 10.62 & 14.38 \\
    Janus-pro-7B      & 384   & AR & 44.15 & 16.64 & 9.17  & 7.76  & 5.61  & 5.07  & 14.73 \\
    SEED-X            &1024   & MM & 35.03 & 21.86 & 9.07  & 14.25 & 8.68  & 7.15  & 16.01 \\
    FLUX.1-[dev] (recaption)&1024 & DM & 23.40 & 18.99 & 13.00 & 13.83 & 14.76 & 14.31 & 16.38 \\
    FLUX.1-[dev]        &1024   & DM & 26.50 & 21.65 & 17.37 & 12.23 & 13.16 & 14.37 & 17.55 \\
    FLUX.1-[pro]      &1024   & DM & 43.07 & 30.25 & 23.59 & 21.12 & 23.82 & 24.96 & 27.80 \\
    HiDream-I1-Full   &1024   & DM & 41.92 & 28.26 & 28.68 & 27.37 & 21.59 & 28.28 & 29.35 \\
    \rowcolor{green!30}GPT-4o  &1024   & MM & 66.45 & 54.25 & 46.51 & 46.60 & 42.93 & 47.37 & 50.69 \\
    \hline
    FLUX-Reason (o3)  &1024   & DM & 37.54 & 30.37 & 28.98 & 26.91 & 21.38 & 30.35 & 29.25 \\
    \rowcolor{orange!20} FLUX-Reason (R1-7B) & 1024 & DM & 42.49 & 31.73 & 32.52 & 28.24 & 25.78 & 32.52 & 32.21 \\
    \rowcolor{blue!20} FLUX-Reason (R1) & 1024 & DM & 44.26 & 39.58 & 33.28 & 36.71 & 25.80 & 33.27 & 35.48 \\
\end{tabular}

}
\vspace{-2mm}
\end{table}

\begin{table}[htbp]
\vspace{-2mm}
  \centering
  \footnotesize
  \caption{\footnotesize{\texttt{MMMG-Score} (×100) for \textbf{History} across prevalent image generation models. The top three average scores are highlighted in green, blue, and orange.}}%
  \label{tab:history_results}
\tablestyle{3pt}{1.2}
\resizebox{0.99\linewidth}{!}{
  \begin{tabular}{l|c|l|cccccc|c}
    {Model} & {Resolution} & {Type} 
      & {Preschool} & {Primary} & {Secondary} & {High} & {Undergrad} & {PhD} 
      & {Avg} \\
    \shline
    LlamaGen          & 512   & AR & 1.21 & 3.41 & 1.12 & 1.90 & 0.42 & 0.95 & 1.50 \\
    Emu-3             & 720   & MM & 4.03 & 7.47 & 3.97 & 5.43 & 3.43 & 2.33 & 4.44 \\
    Ideogram V2       &1024   & DM & 3.96 & 2.15 & 7.03 & 4.91 & 4.92 & 8.77 & 5.29 \\
    JanusFlow-1.3B    & 384   & AR & 6.85 & 11.81 & 8.82 & 6.88 & 2.85 & 2.51 & 6.62 \\
    SimpleAR          &1024   & AR & 6.91 & 10.15 & 7.58 & 7.29 & 5.61 & 4.88 & 7.07 \\
    BAGEL             &1024   & MM & 8.50 & 11.82 & 8.25 & 5.61 & 5.80 & 6.80 & 7.80 \\
    CogView-4         &1024   & DM & 6.75 & 8.37 & 9.44 & 10.09 & 6.73 & 7.51 & 8.15 \\
    Janus-pro-7B      & 384   & AR & 13.08 & 10.68 & 10.97 & 8.06 & 5.09 & 5.90 & 8.96 \\
    Infinity-8B       &1024   & AR & 11.09 & 9.82 & 9.48 & 11.35 & 8.04 & 8.61 & 9.73 \\
    SDXL-1.0          &1024   & DM & 3.05 & 13.77 & 13.70 & 15.97 & 7.35 & 6.44 & 10.05 \\
    FLUX.1-[dev] (recaption)&1024 & DM & 10.65 & 11.90 & 13.00 & 8.19 & 9.01 & 9.65 & 10.40 \\
    SDXL-1.0-refiner  &1024   & DM & 2.43 & 14.95 & 17.77 & 16.85 & 9.18 & 5.79 & 11.16 \\
    FLUX.1-[dev]        &1024   & DM & 9.95 & 15.15 & 15.78 & 12.29 & 9.99 & 12.69 & 12.64 \\
    SEED-X            &1024   & MM & 17.78 & 18.90 & 18.80 & 18.21 & 10.28 & 14.27 & 16.37 \\
    FLUX.1-[pro]      &1024   & DM & 26.40 & 21.26 & 24.56 & 22.47 & 18.74 & 20.21 & 22.27 \\
    HiDream-I1-Full   &1024   & DM & 26.84 & 29.92 & 29.80 & 24.63 & 24.81 & 28.54 & 27.42 \\
    \rowcolor{green!30}GPT-4o  &1024   & MM & 36.84 & 44.38 & 45.62 & 42.76 & 37.76 & 42.59 & 41.66 \\
    \hline
    FLUX-Reason (o3)  &1024   & DM & 16.21 & 26.69 & 28.87 & 22.72 & 26.05 & 26.58 & 24.52 \\
    \rowcolor{orange!20} FLUX-Reason (R1-7B) & 1024 & DM & 28.65 & 28.29 & 29.51 & 26.22 & 25.55 & 27.57 & 27.63 \\
    \rowcolor{blue!20} FLUX-Reason (R1) & 1024 & DM & 28.45 & 34.43 & 38.21 & 34.17 & 30.42 & 31.39 & 32.84 \\
\end{tabular}

}
\vspace{-2mm}
\end{table}

\begin{table}[htbp]
\vspace{-2mm}
  \centering
  \footnotesize
  \caption{\footnotesize{\texttt{MMMG-Score} (×100) for \textbf{Philosophy} across prevalent image generation models. Scores are reported over five educational stages (no Preschool data). The top three average scores are highlighted in green, blue, and orange.}}%
  \label{tab:philosophy_results}
\tablestyle{3pt}{1.2}
\resizebox{0.99\linewidth}{!}{
\begin{tabular}{l|c|l|ccccc|c}
    {Model} & {Resolution} & {Type} 
    & {Primary} & {Secondary} & {High} & {Undergrad} & {PhD} 
    & {Avg} \\
    \shline
    LlamaGen          & 512   & AR & 5.23 & 0.95 & 0.81 & 0.77 & 1.8 & 1.91 \\
    Emu-3             & 720   & MM & 6.89 & 2.83 & 2.51 & 0.96 & 4.14 & 3.47 \\
    JanusFlow-1.3B    & 384   & AR & 14.63 & 3.06 & 1.19 & 3.48 & 4.89 & 5.45 \\
    SimpleAR          &1024   & AR & 11.52 & 3.26 & 3.88 & 3.35 & 8.16 & 6.03 \\
    BAGEL             &1024   & MM & 16.51 & 7.9 & 3.84 & 2.51 & 8.69 & 7.89 \\
    Janus-pro-7B      & 384   & AR & 18.42 & 5.33 & 4.4 & 5.35 & 7.96 & 8.29 \\
    SDXL-1.0          &1024   & DM & 18.35 & 8.01 & 6.4 & 6.63 & 8.0 & 9.48 \\
    SEED-X            &1024   & MM & 20.49 & 8.63 & 6.36 & 4.95 & 7.98 & 9.68 \\
    SDXL-1.0-refiner  &1024   & DM & 20.21 & 8.68 & 7.82 & 7.46 & 5.87 & 10.01 \\
    CogView-4         &1024   & DM & 23.83 & 10.28 & 5.98 & 5.49 & 9.08 & 10.93 \\
    FLUX.1-[dev] (recaption) &1024 & DM & 23.55 & 15.98 & 14.44 & 13.33 & 9.47 & 15.35 \\
    Ideogram V2       &1024   & DM & 23.0 & 17.64 & 14.31 & 15.63 & 14.09 & 16.93 \\
    Infinity-8B       &1024   & AR & 21.7 & 17.86 & 15.12 & 14.1 & 18.93 & 17.54 \\
    FLUX.1-[dev]        &1024   & DM & 29.05 & 18.28 & 15.98 & 12.37 & 19.23 & 18.98 \\
    FLUX.1-[pro]      &1024   & DM & 36.72 & 29.32 & 26.34 & 22.16 & 26.02 & 28.11 \\
    HiDream-I1-Full   &1024   & DM & 40.39 & 30.78 & 27.82 & 19.4 & 28.24 & 29.33 \\
    \rowcolor{green!30}GPT-4o  &1024   & MM & 59.58 & 48.55 & 45.21 & 42.35 & 45.03 & 48.14 \\
    \hline
    FLUX-Reason (o3)  &1024   & DM & 33.48 & 22.78 & 27.33 & 21.1 & 25.76 & 26.09 \\
    \rowcolor{blue!20} FLUX-Reason (R1-7B) & 1024 & DM & 41.57 & 34.6 & 30.02 & 26.55 & 30.47 & 32.64 \\
    \rowcolor{orange!20} FLUX-Reason (R1) & 1024 & DM & 41.5 & 36.08 & 30.45 & 26.04 & 27.84 & 32.38 \\
\end{tabular}

}
\vspace{-2mm}
\end{table}

\newpage
\section{Ablation Study}

\subsection{Self-Consistency of MMMG–Eval}

A truly reliable evaluation metric must exhibit self-consistency: when applied to ground-truth images, \texttt{MMMG} should recover the original knowledge graph with minimal error. We assess this property by running OpenAI-o3 as the evaluator and compute two complementary scores on all reference images: the knowledge fidelity term
\[
1 - \mathrm{GED}(G_{\mathrm{gen}}, G_{\mathrm{ref}})
\]
and the unified accuracy
\[
\mathrm{u\text{-}acc} = \frac{N_{e}^{\mathrm{correct}} + N_{d}^{\mathrm{correct}}}{N_{e} + N_{d}},
\]
where \(N_{e}\), \(N_{d}\) are the total number of entities and dependencies, and \(N_{e}^{\mathrm{correct}}\), \(N_{d}^{\mathrm{correct}}\) the correctly recovered counts. Table~\ref{tab:reflexivity} reports these metrics across six educational stages. We observe consistently high fidelity (\(>0.92\)) and accuracy (\(>0.91\)) without any stage-specific tuning, confirming that \texttt{MMMG} faithfully grounds and retrieves structured knowledge from its source images.

\begin{table}[ht]
  \centering
  \small
  \renewcommand{\arraystretch}{1.1}
  \begin{tabular}{lcccccc}
    \toprule
     & \textbf{Preschool} & \textbf{Primary} & \textbf{Secondary} & \textbf{High School} & \textbf{Undergrad} & \textbf{PhD} \\
    \midrule
    \(1 - \mathrm{GED}\)      & 0.9373 & 0.9336 & 0.9318 & 0.9307 & 0.9287 & 0.9360 \\
    \(\mathrm{u\text{-}acc}\) & 0.9261 & 0.9218 & 0.9154 & 0.9110 & 0.9154 & 0.9240 \\
    \bottomrule
  \end{tabular}
  \caption{Self-consistency evaluation: structural fidelity (\(1-\)GED) and unified accuracy (u-acc) of \texttt{MMMG} on ground-truth images across different educational stages.}
  \label{tab:reflexivity}
\end{table}

\subsection{Robustness of MMMG Evaluation}

In addition to self-consistency, a reliable evaluation metric should be robust to changes in its underlying judgment engine. To assess this, we recompute the \texttt{MMMG–Score} using three variants of OpenAI’s language models—OpenAI-o3 (high reasoning effort), OpenAI-o3, and OpenAI-o1 (lightweight)—on the same set of generated images. As shown in Table~\ref{tab:robustness}, the relative rankings of generation models (GPT-4o-Image, FLUX-Reason (R1), FLUX.1-[pro], Infinity-8B, SEED-X) remain consistent across all evaluator variants, with only minor score fluctuations. This stability indicates that \texttt{MMMG–Score} is robust to variations in the choice of LLM.

\begin{table}[ht]
  \centering
  \small
  \renewcommand{\arraystretch}{1.1}
  \begin{tabular}{lccccc}
    \toprule
    \textbf{Evaluator} & \textbf{GPT-4o} & \textbf{FLUX-Reason (R1)} & \textbf{FLUX.1-[pro]} & \textbf{Infinity-8B} & \textbf{SEED-X} \\
    \midrule
    OpenAI-o3-high & 50.20 & 34.45 & 27.14 & 19.18 & 18.16 \\
    OpenAI-o3      & 51.07 & 34.62 & 27.33 & 19.54 & 18.49 \\
    OpenAI-o1      & 52.81 & 37.25 & 29.52 & 21.73 & 20.86 \\
    \bottomrule
  \end{tabular}
  
  \caption{Robustness evaluation: \texttt{MMMG–Score} (higher is better) recomputed with different OpenAI LLM variants. The preserved model ranking and low score variance confirm stability against evaluator changes.}
  \label{tab:robustness}
\end{table}

\newpage
\section{Visualization}
\subsection{Preschool}

\subsubsection{Biology}
\begin{figure}[h]
    \centering
    \includegraphics[width=\linewidth]{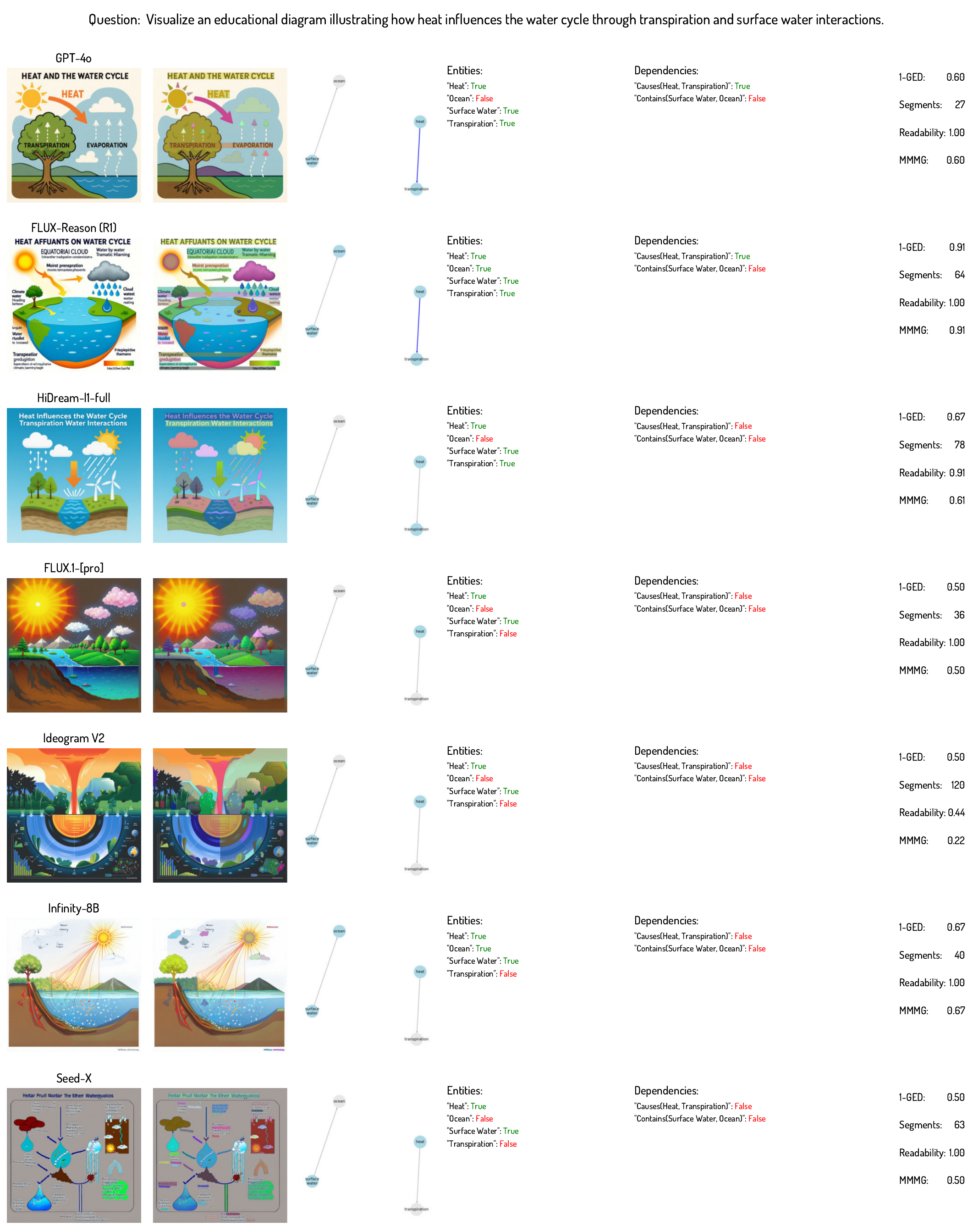}
    
    \caption{\texttt{MMMG} Benchmark visualization for seven representative models on a Preschool‐Biology example. Each row corresponds to one model and, from left to right, displays the generated image, its segmentation map, the reconstructed knowledge graph, the extracted entity and dependency lists, and finally the overall \texttt{MMMG‐Score} along with its component sub‐scores.}
    \label{fig:BIOLOGY}
\end{figure}
\newpage
\subsubsection{Chemistry}
\begin{figure}[h]
    \centering
    \includegraphics[width=\linewidth]{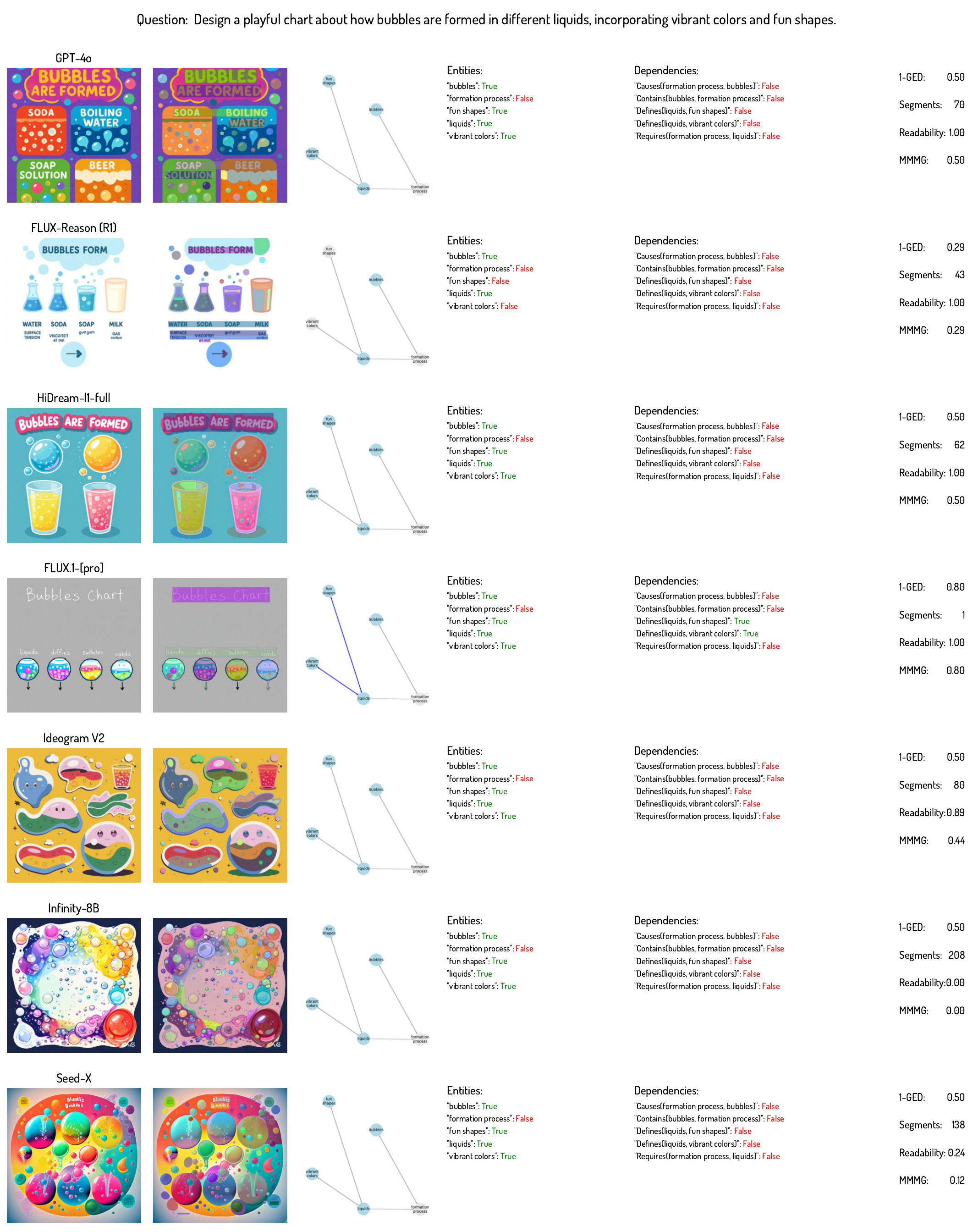}
    \caption{\texttt{MMMG} Benchmark visualization for seven representative models on a Preschool‐Chemistry example. Each row corresponds to one model and, from left to right, displays the generated image, its segmentation map, the reconstructed knowledge graph, the extracted entity and dependency lists, and finally the overall \texttt{MMMG‐Score} along with its component sub‐scores.}
    \label{fig:CHEMISTRY}
\end{figure}
\newpage
\subsubsection{Mathematics}
\begin{figure}[h]
    \centering
    \includegraphics[width=\linewidth]{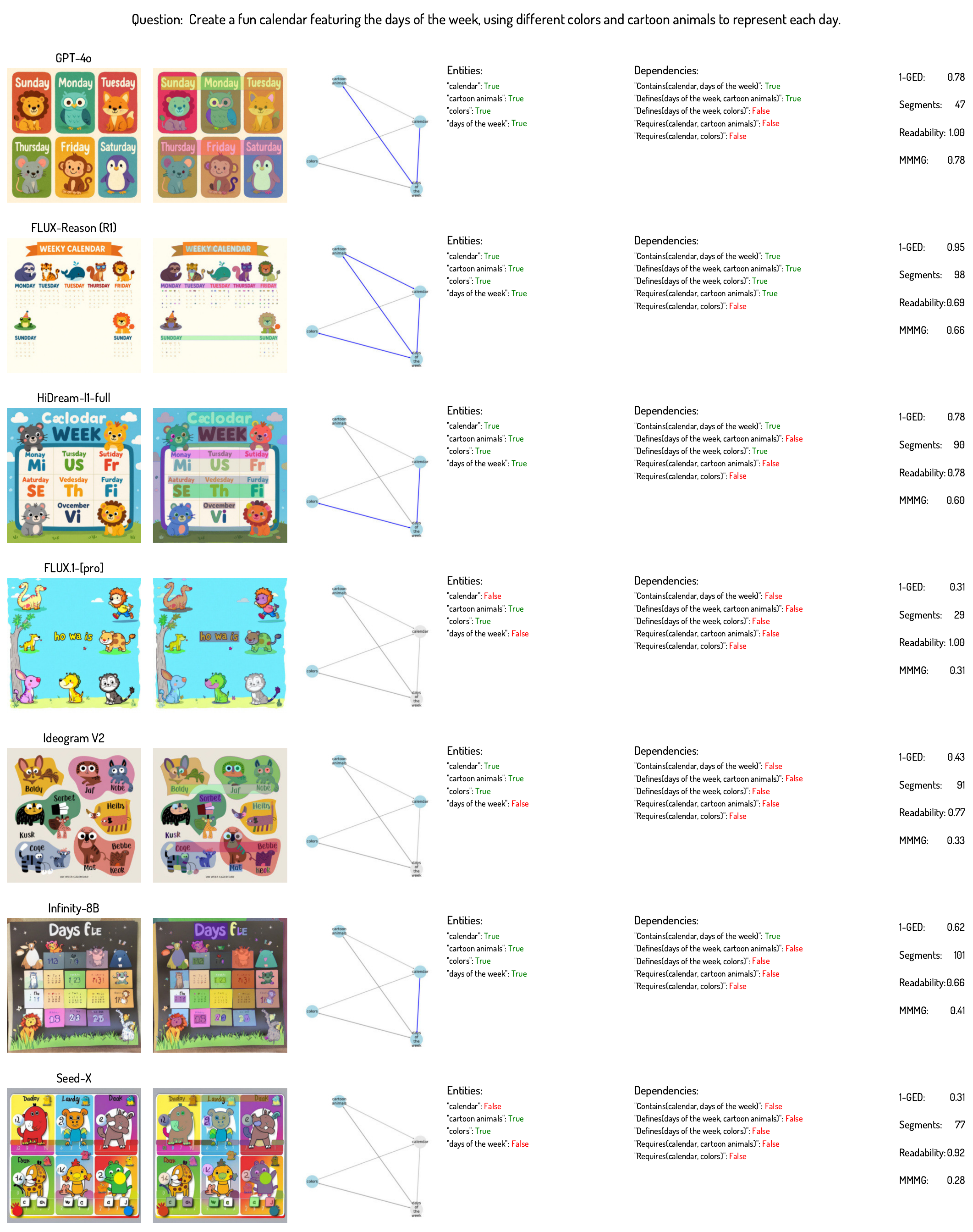}
    \caption{\texttt{MMMG} Benchmark visualization for seven representative models on a Preschool‐Mathematics example. Each row corresponds to one model and, from left to right, displays the generated image, its segmentation map, the reconstructed knowledge graph, the extracted entity and dependency lists, and finally the overall \texttt{MMMG‐Score} along with its component sub‐scores.}
    \label{fig:MATH}
\end{figure}
\newpage
\subsubsection{Engineering}
\begin{figure}[h]
    \centering
    \includegraphics[width=\linewidth]{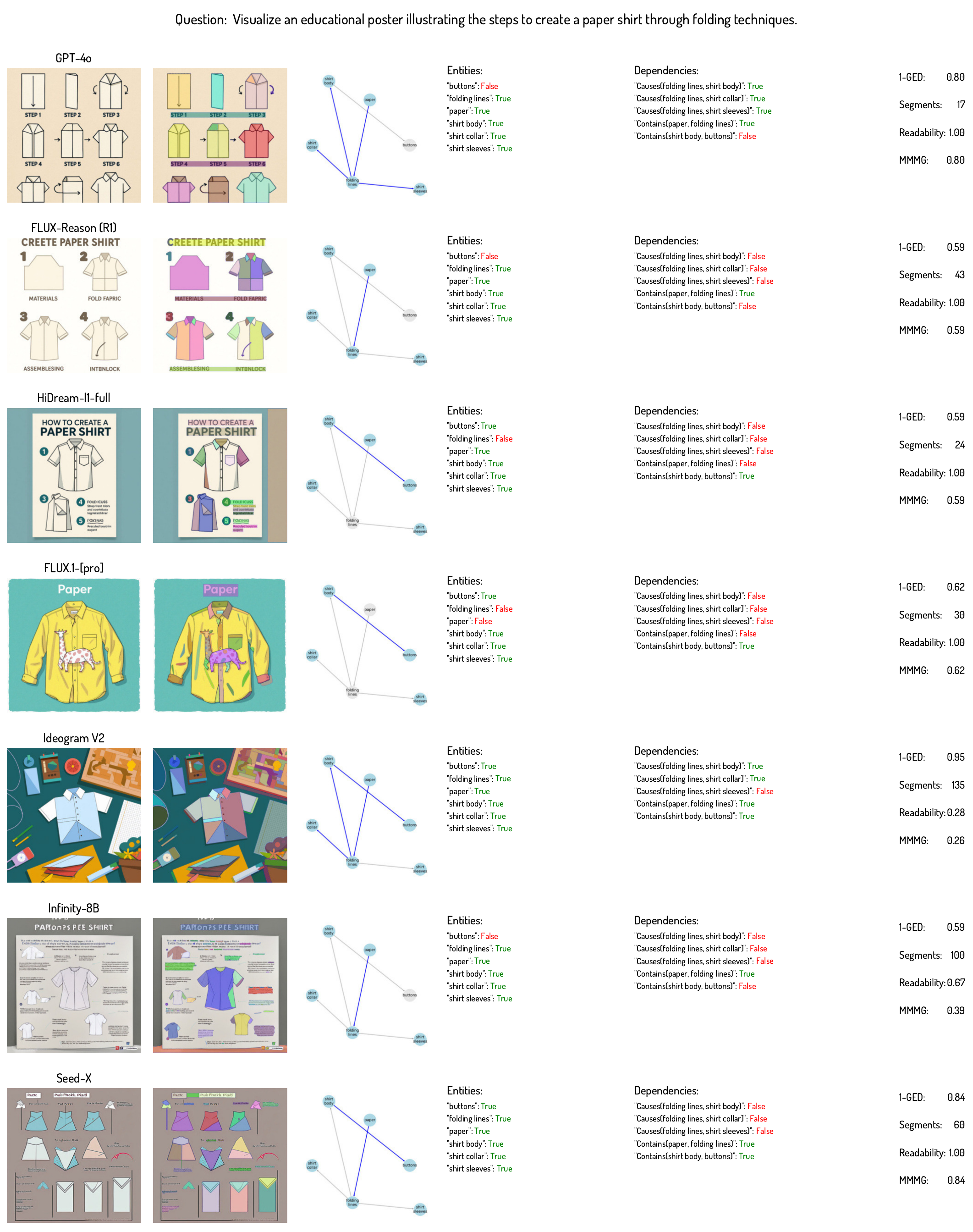}

    \caption{\texttt{MMMG} Benchmark visualization for seven representative models on a Preschool‐Engineering example. Each row corresponds to one model and, from left to right, displays the generated image, its segmentation map, the reconstructed knowledge graph, the extracted entity and dependency lists, and finally the overall \texttt{MMMG‐Score} along with its component sub‐scores.}

    \label{fig:ENGINEER}
\end{figure}
\newpage
\subsubsection{Geography}
\begin{figure}[h]
    \centering
    \includegraphics[width=\linewidth]{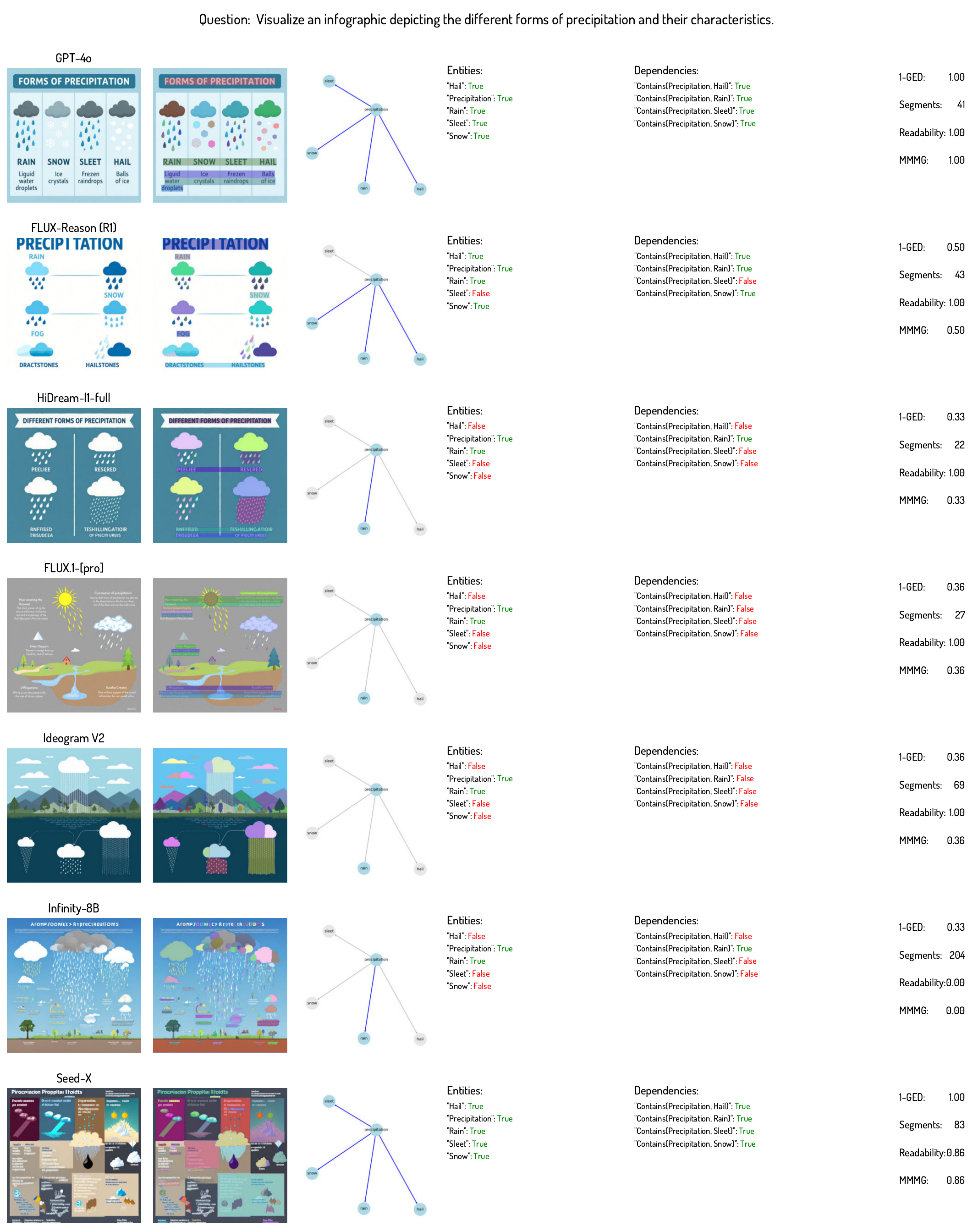}
    \caption{\texttt{MMMG} Benchmark visualization for seven representative models on a Preschool‐Geography example. Each row corresponds to one model and, from left to right, displays the generated image, its segmentation map, the reconstructed knowledge graph, the extracted entity and dependency lists, and finally the overall \texttt{MMMG‐Score} along with its component sub‐scores.}
    \label{fig:enter-label}
\end{figure}
\newpage
\subsubsection{Economics}
\begin{figure}[h]
    \centering
    \includegraphics[width=\linewidth]{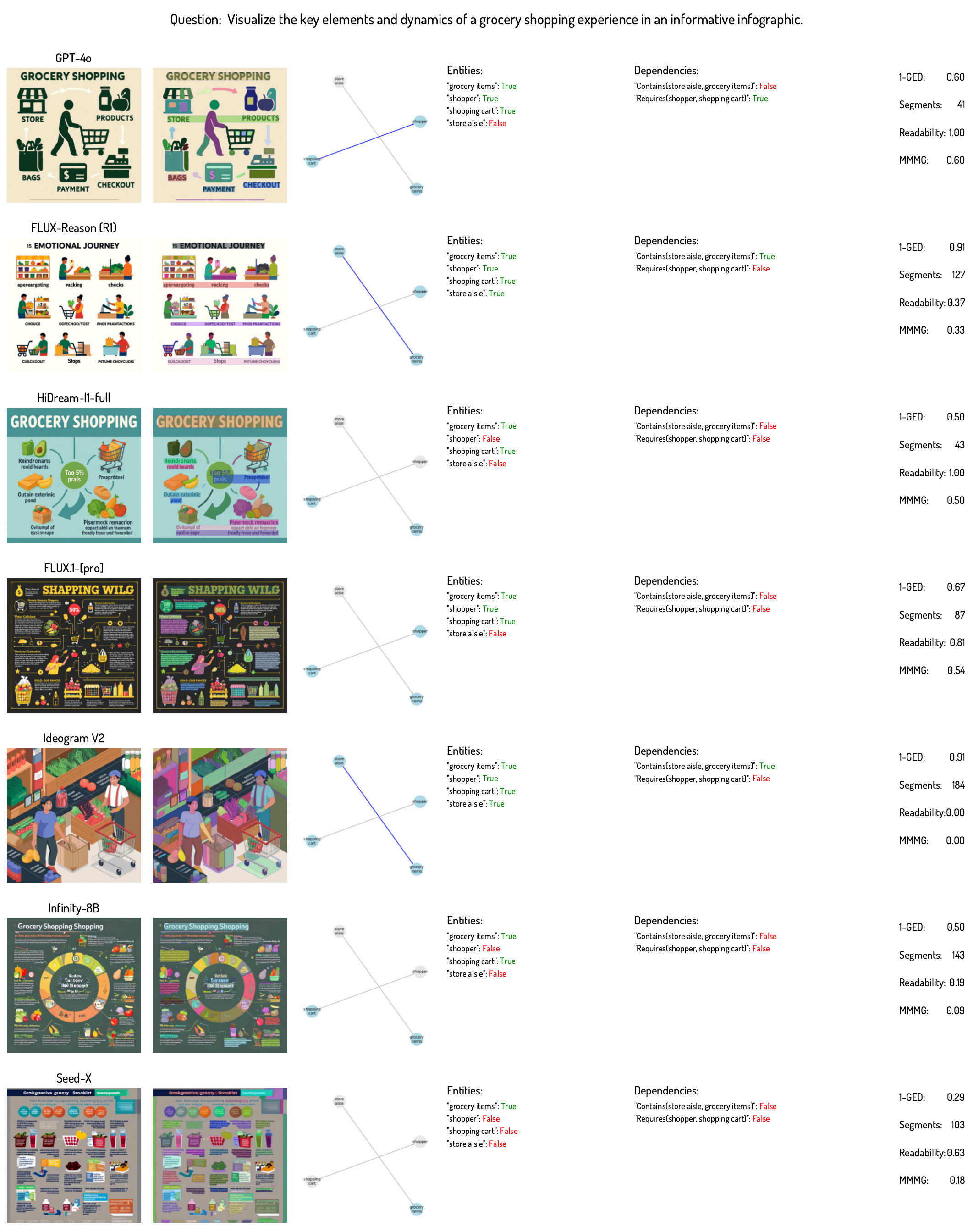}
    \caption{\texttt{MMMG} Benchmark visualization for seven representative models on a Preschool‐Economics example. Each row corresponds to one model and, from left to right, displays the generated image, its segmentation map, the reconstructed knowledge graph, the extracted entity and dependency lists, and finally the overall \texttt{MMMG‐Score} along with its component sub‐scores.}
    \label{fig:enter-label}
\end{figure}
\newpage
\subsubsection{Sociology}
\begin{figure}[h]
    \centering
    \includegraphics[width=\linewidth]{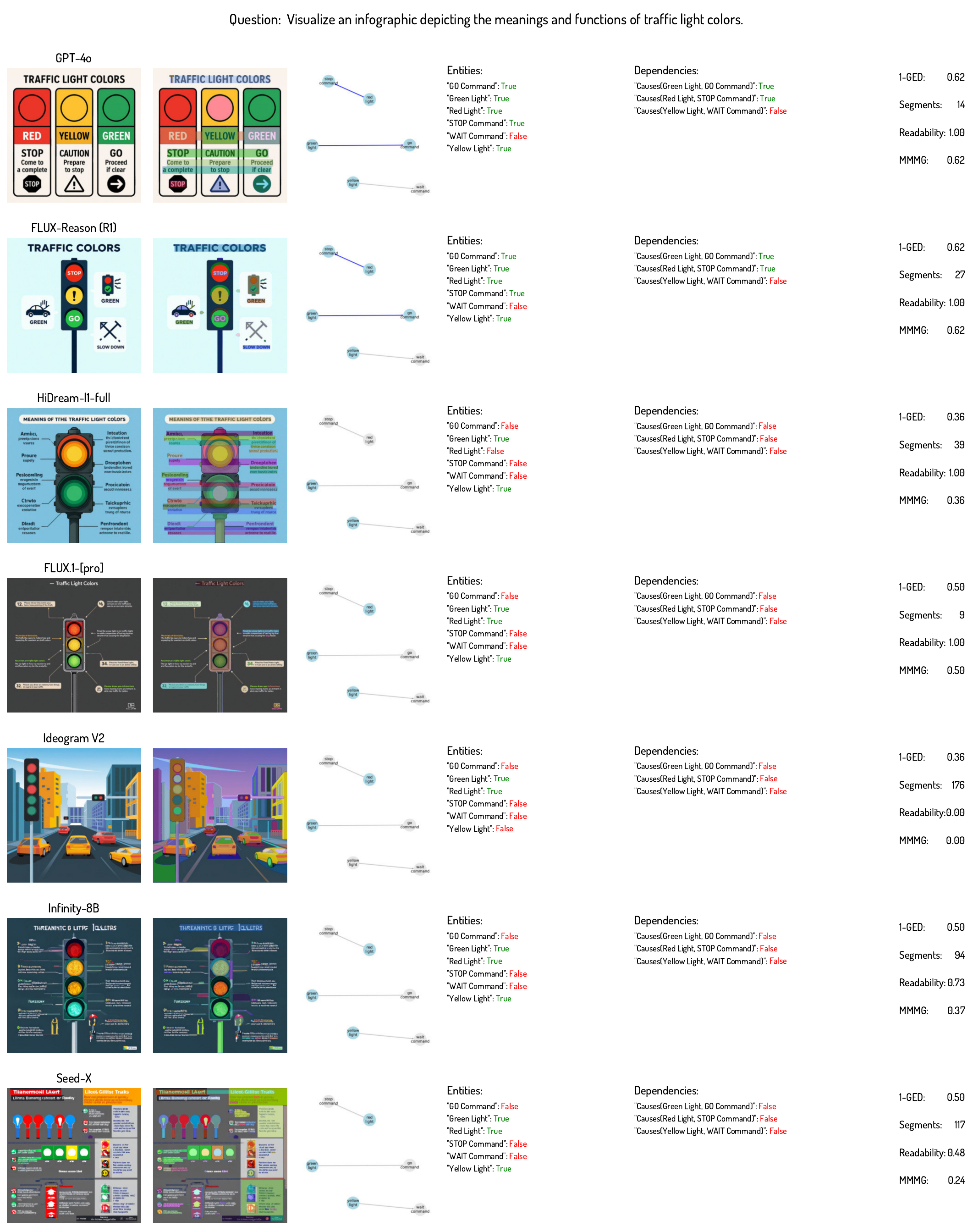}
    \caption{\texttt{MMMG} Benchmark visualization for seven representative models on a Preschool‐Sociology example. Each row corresponds to one model and, from left to right, displays the generated image, its segmentation map, the reconstructed knowledge graph, the extracted entity and dependency lists, and finally the overall \texttt{MMMG‐Score} along with its component sub‐scores.}
    \label{fig:enter-label}
\end{figure}
\newpage
\subsubsection{History}
\begin{figure}[h]
    \centering
    \includegraphics[width=\linewidth]{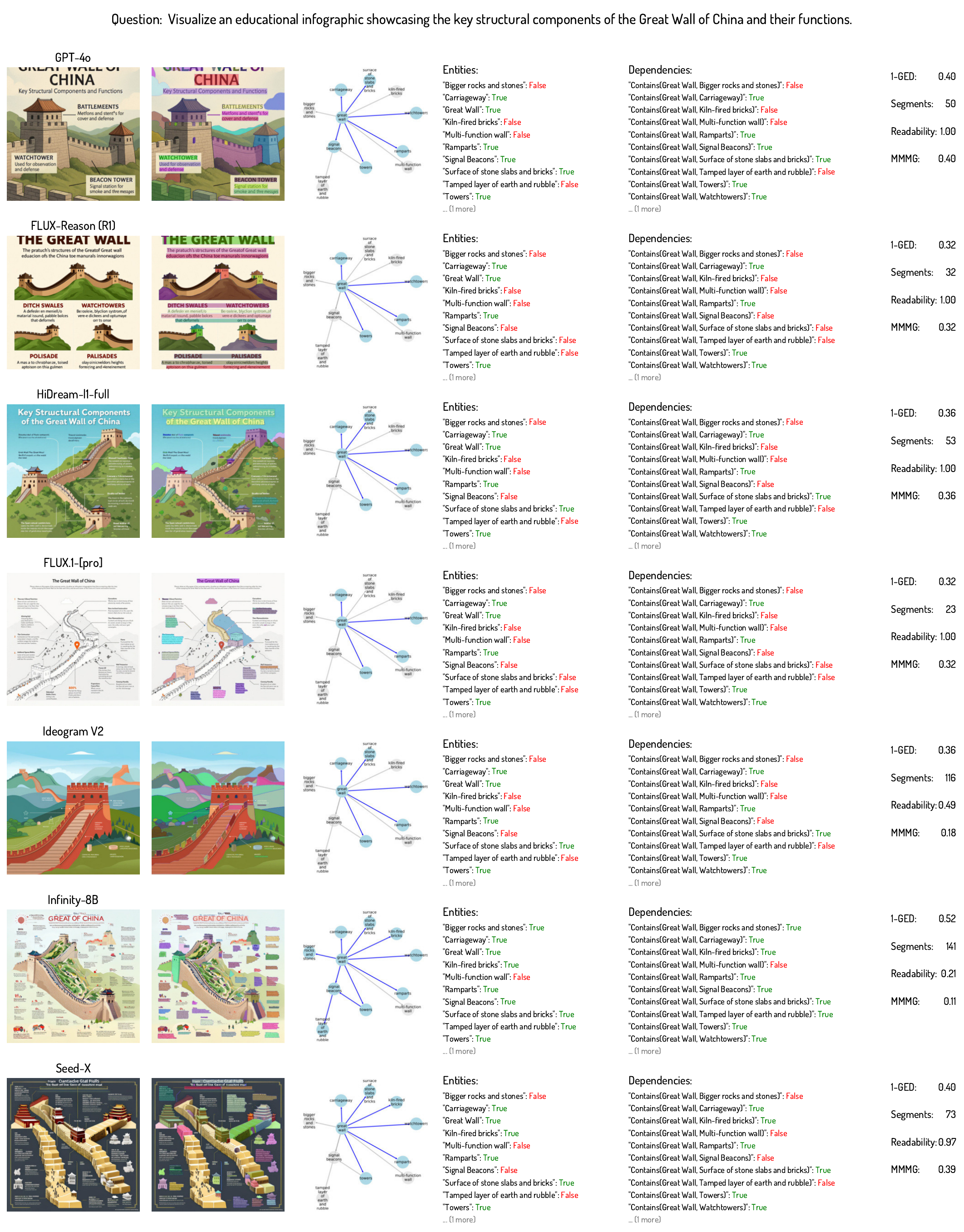}
    \caption{\texttt{MMMG} Benchmark visualization for seven representative models on a Preschool‐History example. Each row corresponds to one model and, from left to right, displays the generated image, its segmentation map, the reconstructed knowledge graph, the extracted entity and dependency lists, and finally the overall \texttt{MMMG‐Score} along with its component sub‐scores.}
    \label{fig:enter-label}
\end{figure}
\newpage
\subsubsection{Literature}
\begin{figure}[h]
    \centering
    \includegraphics[width=\linewidth]{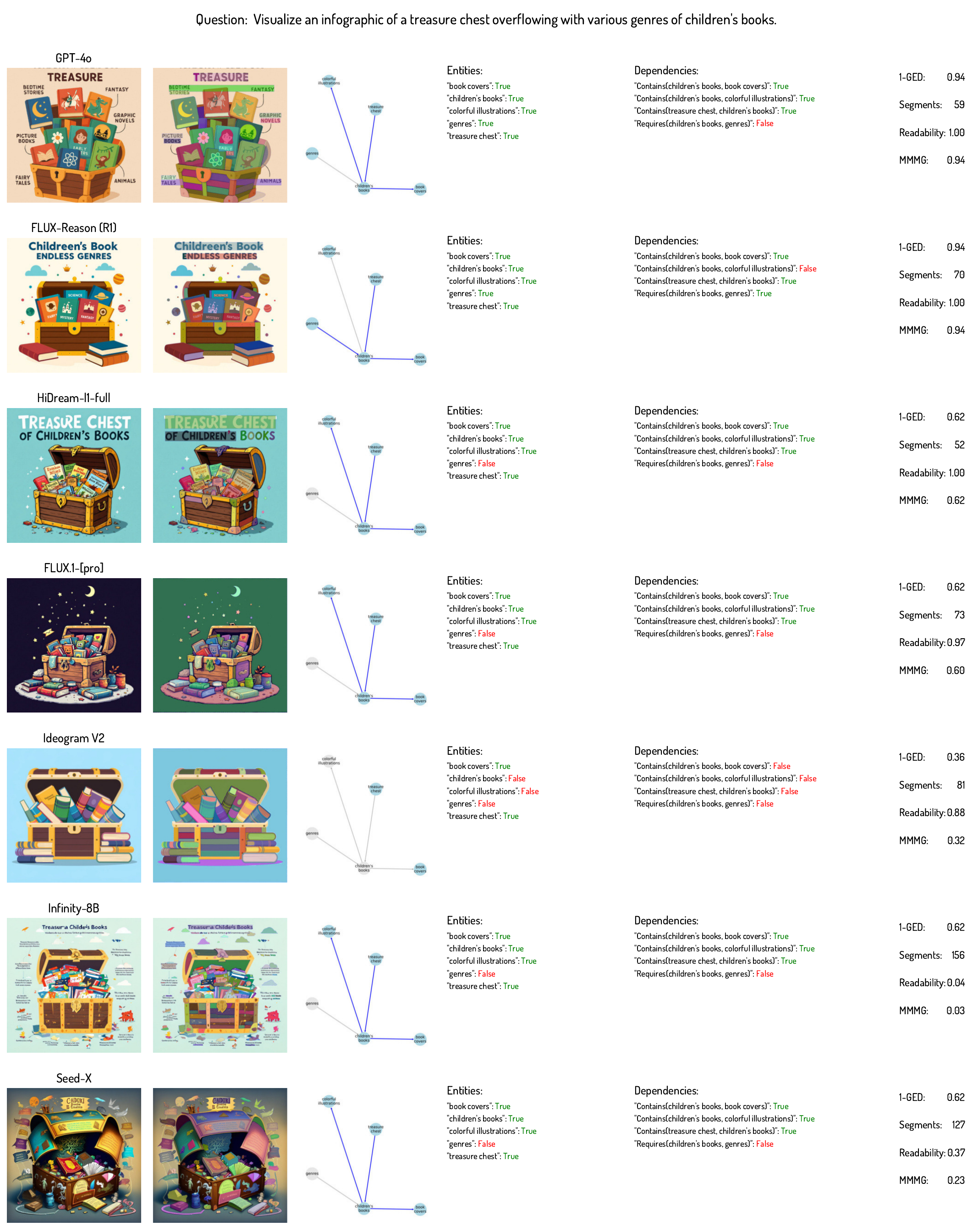}
    \caption{\texttt{MMMG} Benchmark visualization for seven representative models on a Preschool‐Literature example. Each row corresponds to one model and, from left to right, displays the generated image, its segmentation map, the reconstructed knowledge graph, the extracted entity and dependency lists, and finally the overall \texttt{MMMG‐Score} along with its component sub‐scores.}
    \label{fig:enter-label}
\end{figure}

\newpage
\subsection{Primary School}

\subsubsection{Biology}
\begin{figure}[h]
    \centering
    \includegraphics[width=\linewidth]{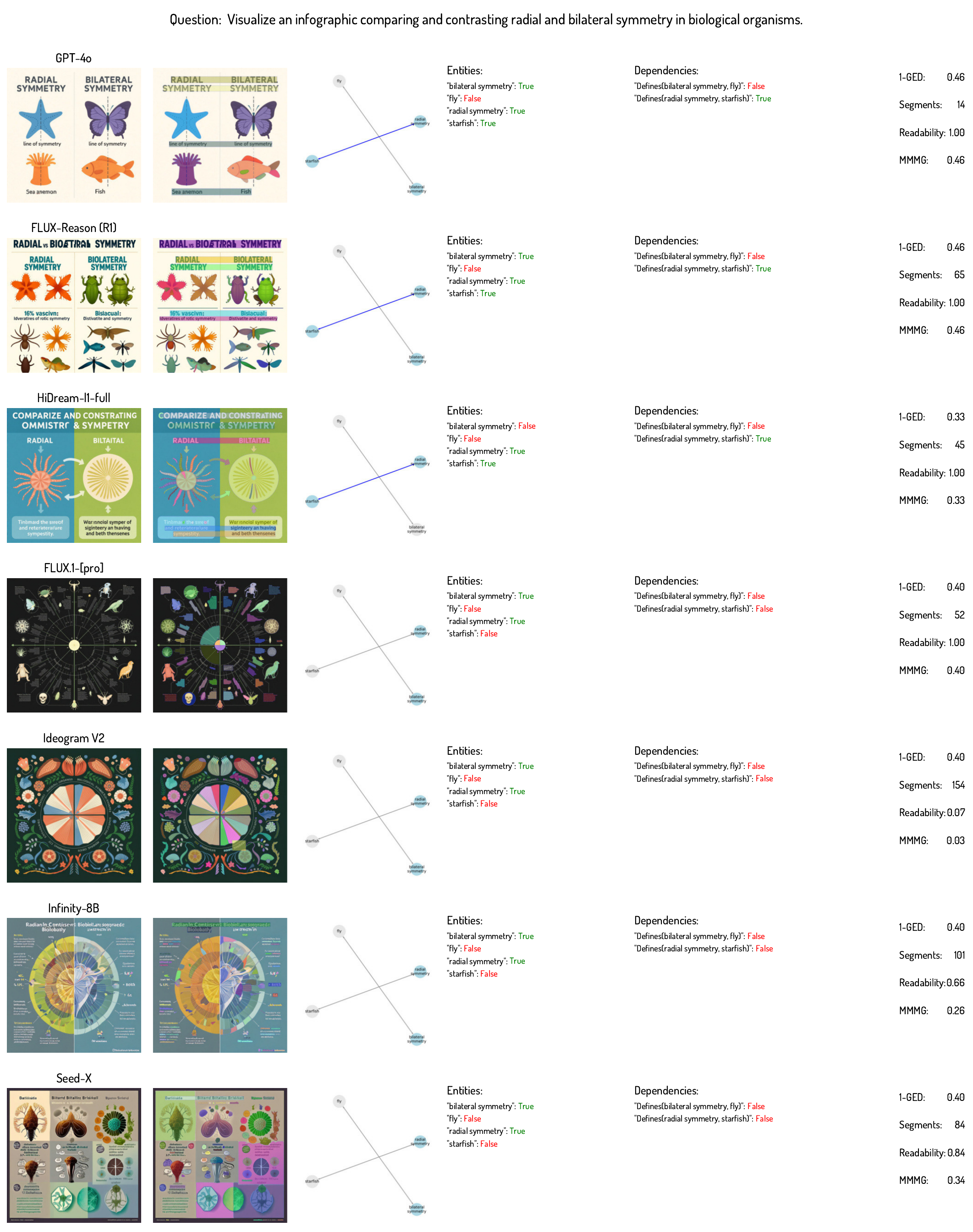}
    \caption{\texttt{MMMG} Benchmark visualization for seven representative models on a Primaryschool‐Biology example. Each row corresponds to one model and, from left to right, displays the generated image, its segmentation map, the reconstructed knowledge graph, the extracted entity and dependency lists, and finally the overall \texttt{MMMG‐Score} along with its component sub‐scores.}
    \label{fig:enter-label}
\end{figure}
\newpage
\subsubsection{Chemistry}
\begin{figure}[h]
    \centering
    \includegraphics[width=\linewidth]{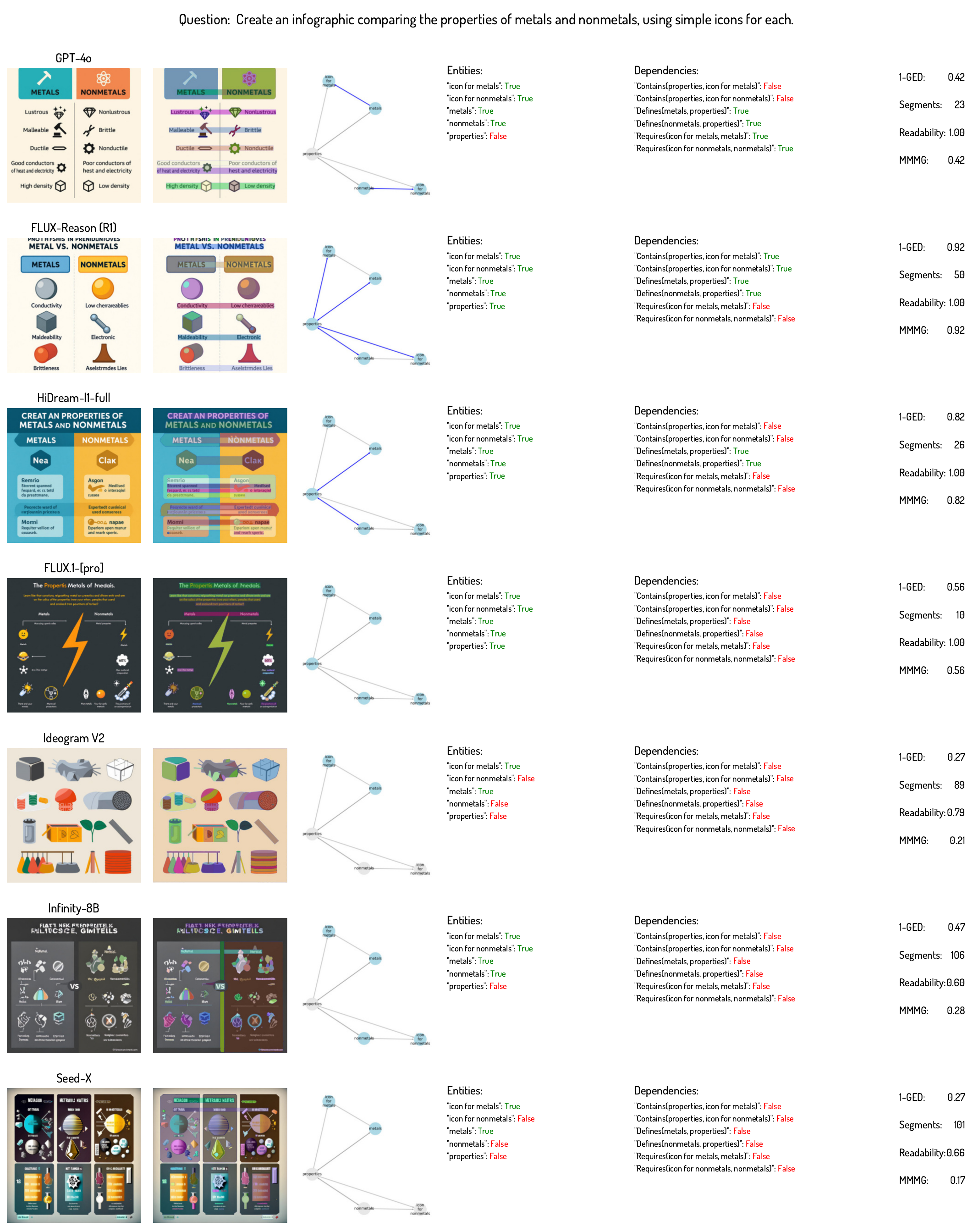}
    \caption{\texttt{MMMG} Benchmark visualization for seven representative models on a Primaryschool‐Chemistry example. Each row corresponds to one model and, from left to right, displays the generated image, its segmentation map, the reconstructed knowledge graph, the extracted entity and dependency lists, and finally the overall \texttt{MMMG‐Score} along with its component sub‐scores.}
    \label{fig:enter-label}
\end{figure}
\newpage
\subsubsection{Mathematics}
\begin{figure}[h]
    \centering
    \includegraphics[width=\linewidth]{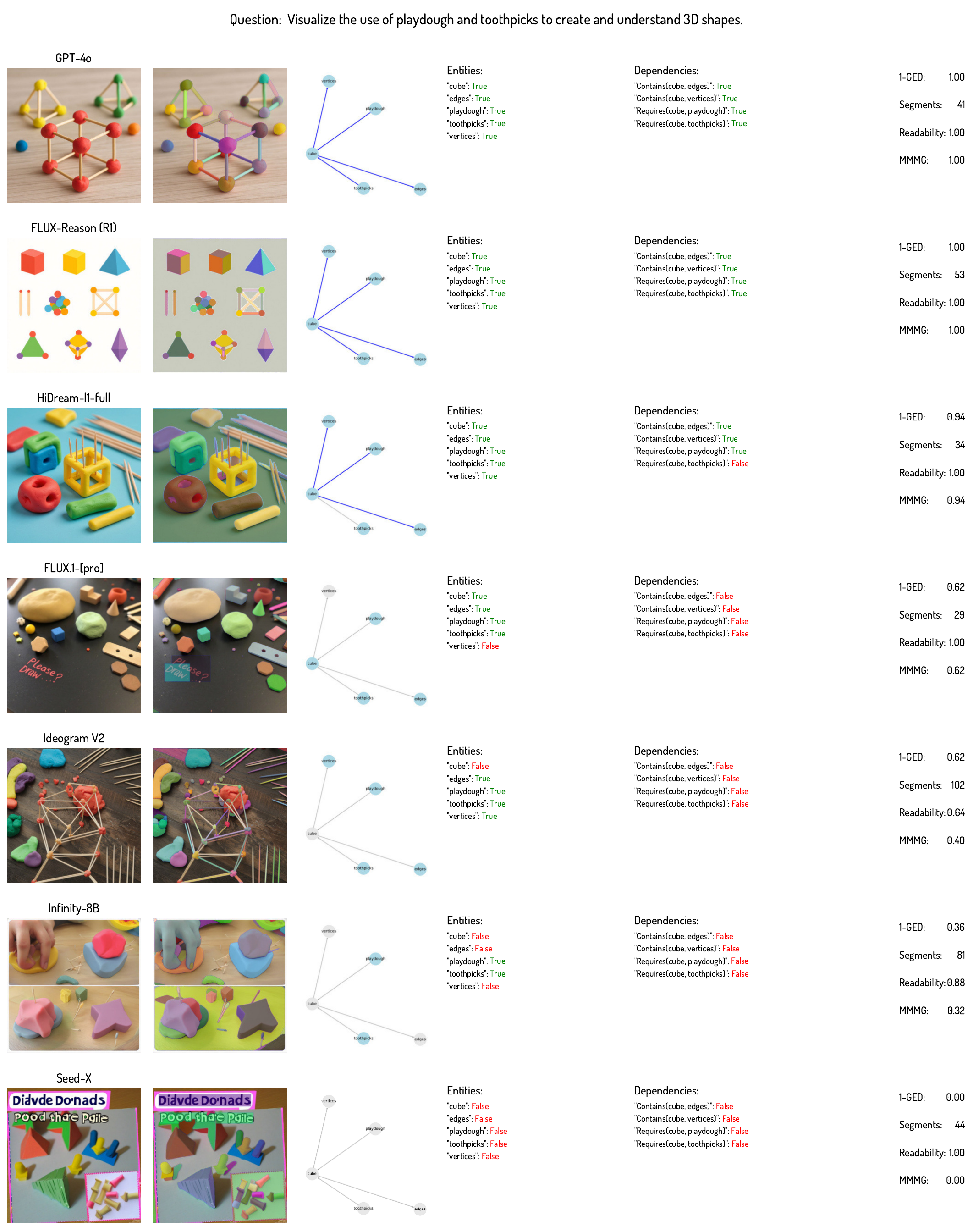}
    \caption{\texttt{MMMG} Benchmark visualization for seven representative models on a Primaryschool‐Mathematics example. Each row corresponds to one model and, from left to right, displays the generated image, its segmentation map, the reconstructed knowledge graph, the extracted entity and dependency lists, and finally the overall \texttt{MMMG‐Score} along with its component sub‐scores.}
    \label{fig:enter-label}
\end{figure}
\newpage
\subsubsection{Engineering}
\begin{figure}[h]
    \centering
    \includegraphics[width=\linewidth]{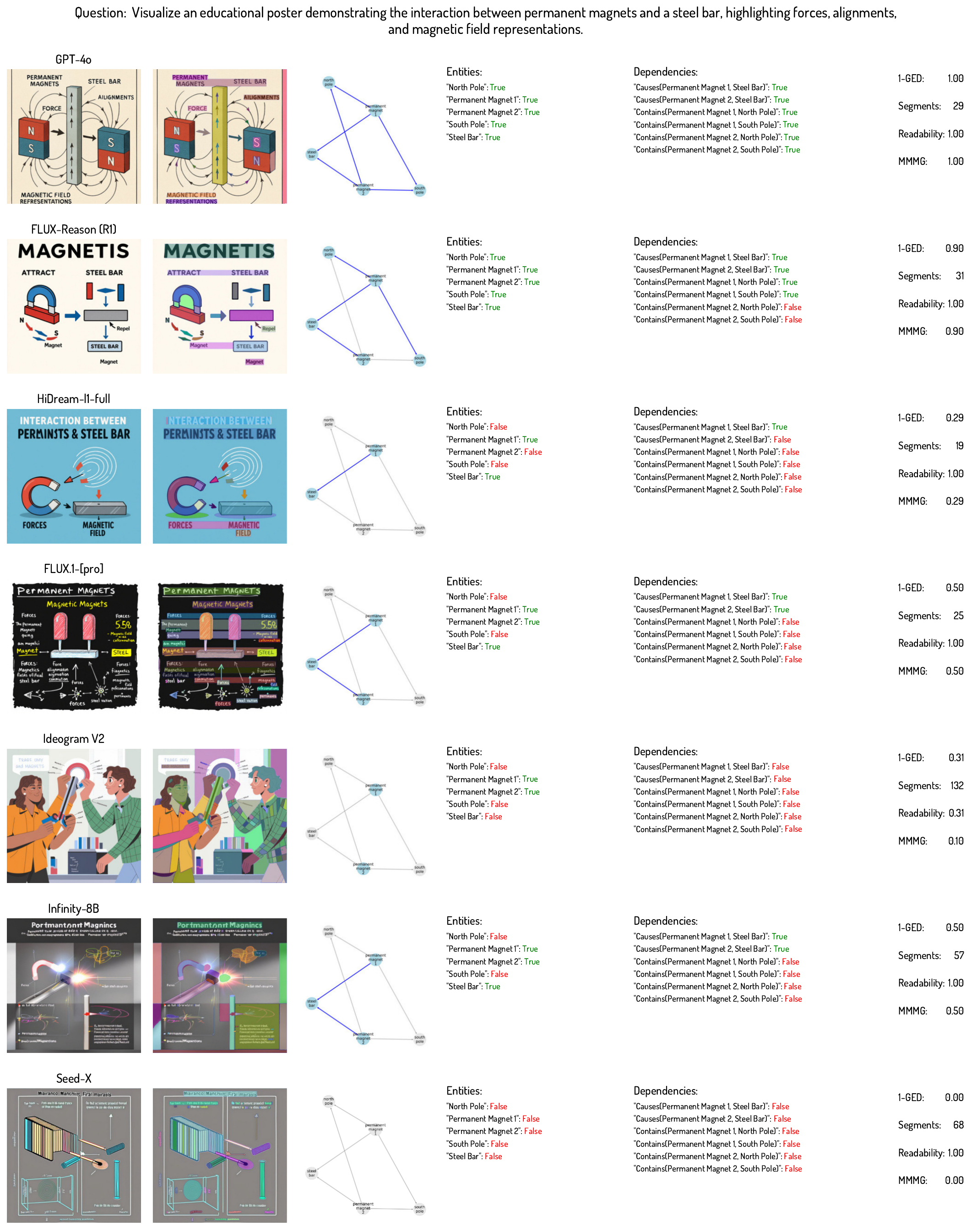}
    \caption{\texttt{MMMG} Benchmark visualization for seven representative models on a Primaryschool‐Engineering example. Each row corresponds to one model and, from left to right, displays the generated image, its segmentation map, the reconstructed knowledge graph, the extracted entity and dependency lists, and finally the overall \texttt{MMMG‐Score} along with its component sub‐scores.}
    \label{fig:enter-label}
\end{figure}
\newpage
\subsubsection{Geography}
\begin{figure}[h]
    \centering
    \includegraphics[width=\linewidth]{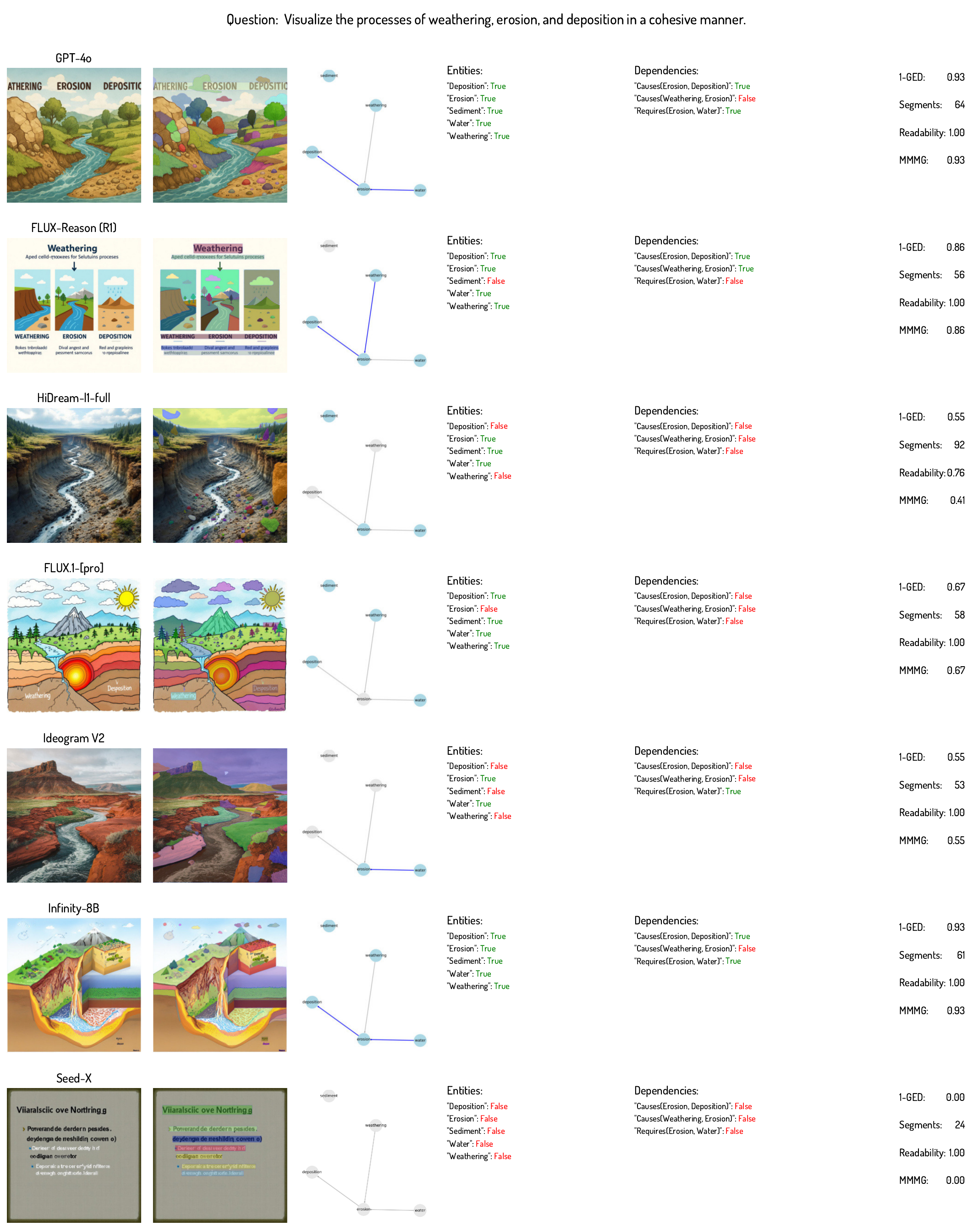}
    \caption{\texttt{MMMG} Benchmark visualization for seven representative models on a Primaryschool‐Geography example. Each row corresponds to one model and, from left to right, displays the generated image, its segmentation map, the reconstructed knowledge graph, the extracted entity and dependency lists, and finally the overall \texttt{MMMG‐Score} along with its component sub‐scores.}
    \label{fig:enter-label}
\end{figure}
\newpage
\subsubsection{Economics}
\begin{figure}[h]
    \centering
    \includegraphics[width=\linewidth]{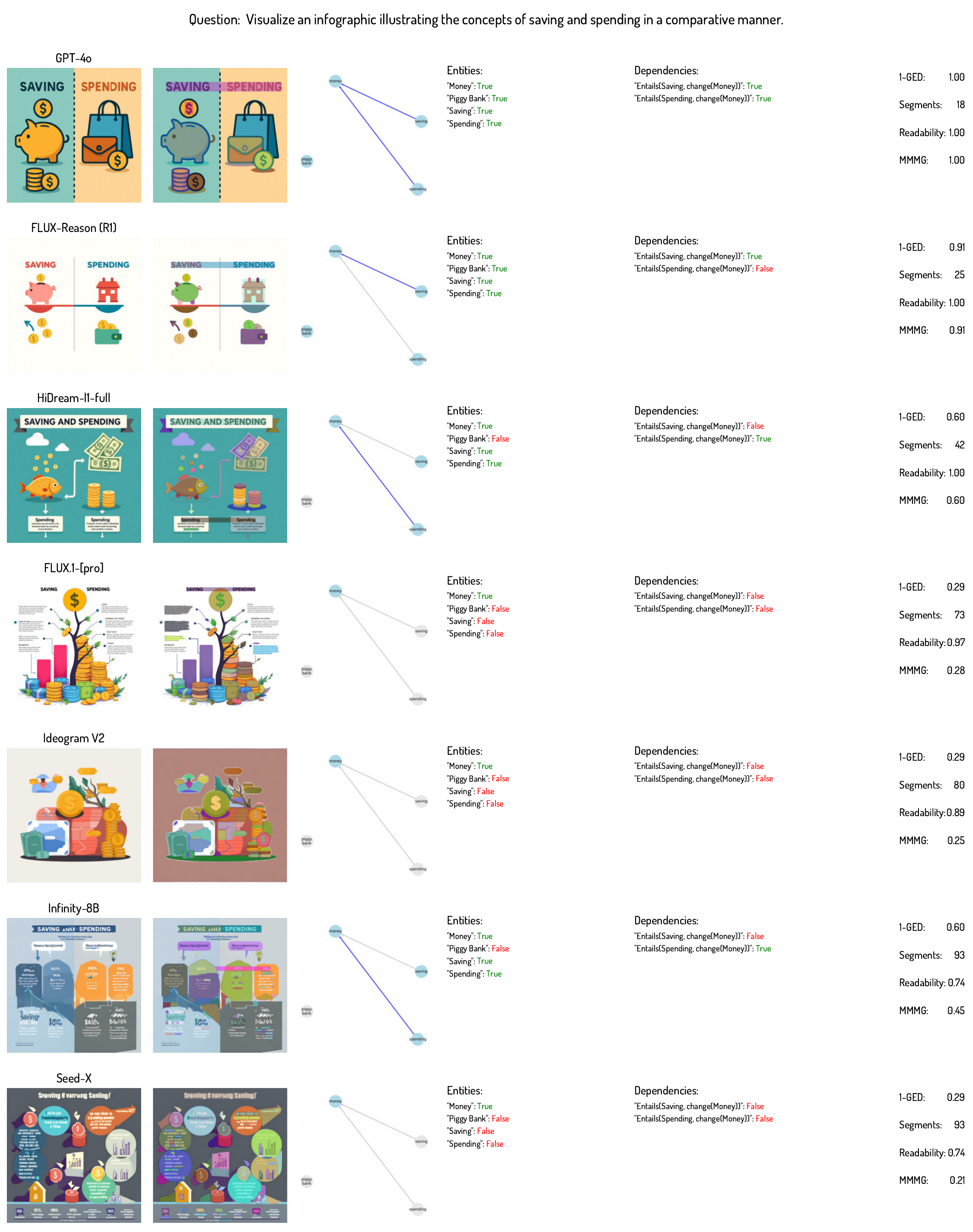}
    \caption{\texttt{MMMG} Benchmark visualization for seven representative models on a Primaryschool‐Economics example. Each row corresponds to one model and, from left to right, displays the generated image, its segmentation map, the reconstructed knowledge graph, the extracted entity and dependency lists, and finally the overall \texttt{MMMG‐Score} along with its component sub‐scores.}
    \label{fig:enter-label}
\end{figure}
\newpage
\subsubsection{Sociology}
\begin{figure}[h]
    \centering
    \includegraphics[width=\linewidth]{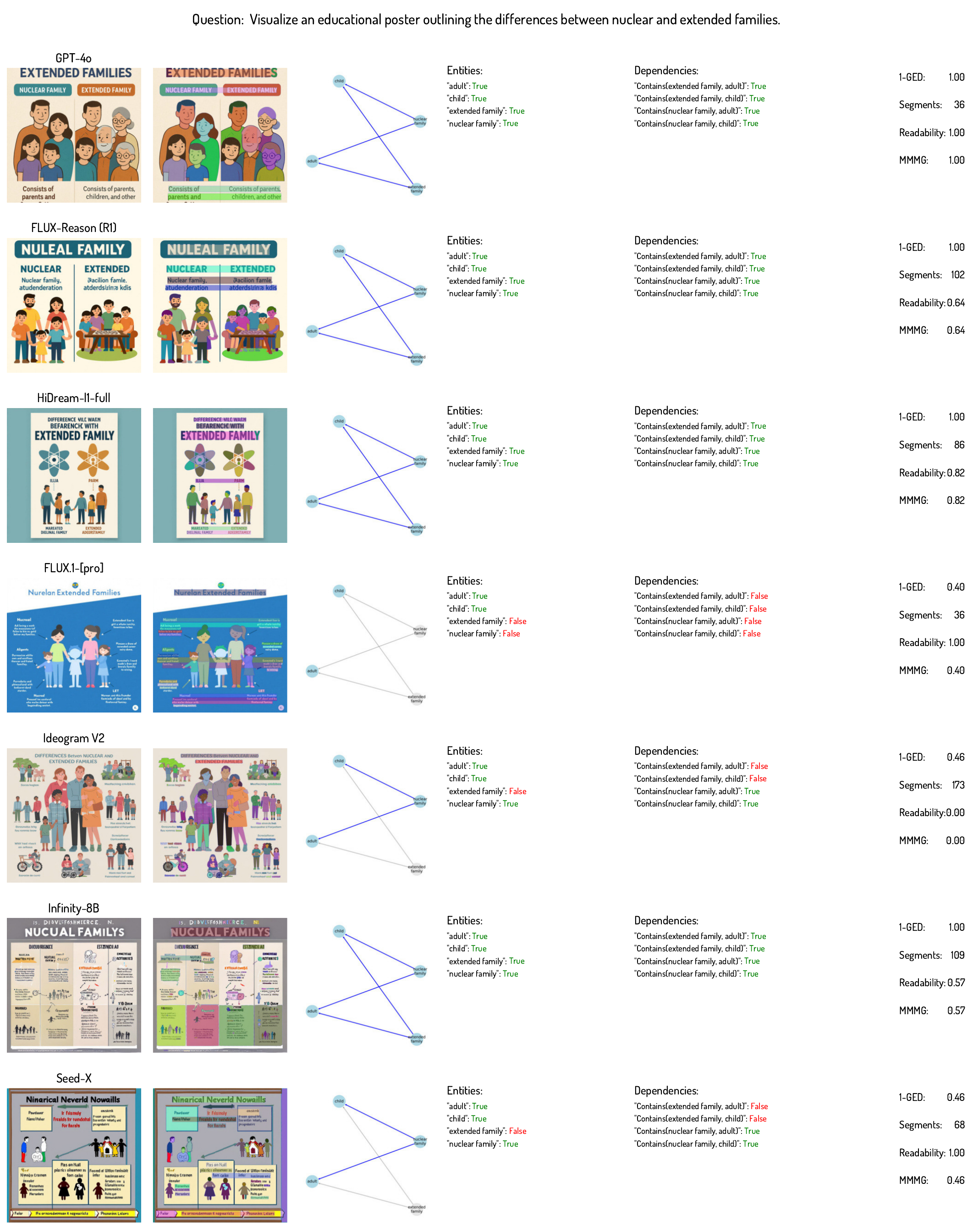}
    \caption{\texttt{MMMG} Benchmark visualization for seven representative models on a Primaryschool‐Sociology example. Each row corresponds to one model and, from left to right, displays the generated image, its segmentation map, the reconstructed knowledge graph, the extracted entity and dependency lists, and finally the overall \texttt{MMMG‐Score} along with its component sub‐scores.}
    \label{fig:enter-label}
\end{figure}
\newpage
\subsubsection{History}
\begin{figure}[h]
    \centering
    \includegraphics[width=\linewidth]{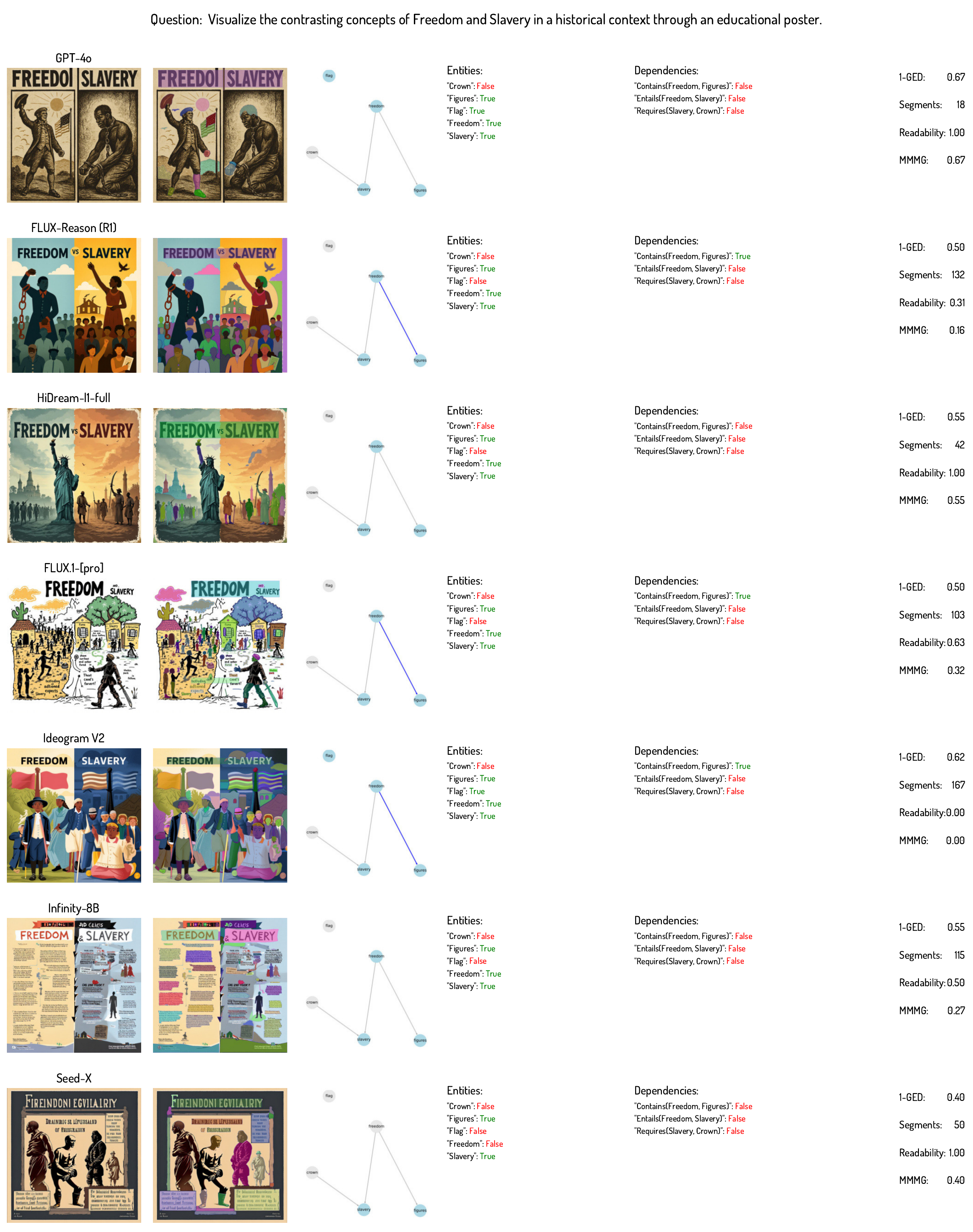}
    \caption{\texttt{MMMG} Benchmark visualization for seven representative models on a Primaryschool‐History example. Each row corresponds to one model and, from left to right, displays the generated image, its segmentation map, the reconstructed knowledge graph, the extracted entity and dependency lists, and finally the overall \texttt{MMMG‐Score} along with its component sub‐scores.}
    \label{fig:enter-label}
\end{figure}
\newpage
\subsubsection{Philosophy}
\begin{figure}[h]
    \centering
    \includegraphics[width=\linewidth]{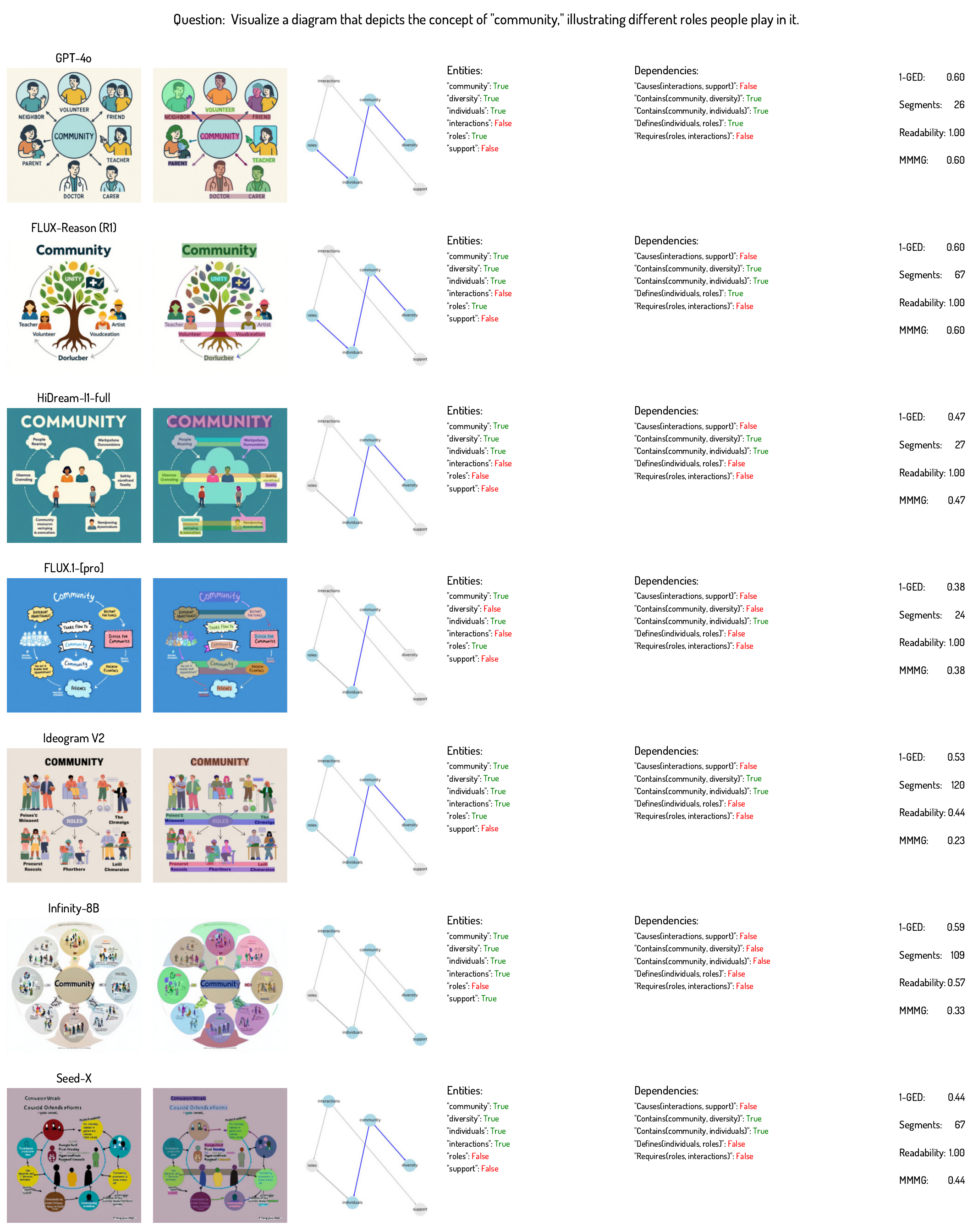}
    \caption{\texttt{MMMG} Benchmark visualization for seven representative models on a Primaryschool‐Philosophy example. Each row corresponds to one model and, from left to right, displays the generated image, its segmentation map, the reconstructed knowledge graph, the extracted entity and dependency lists, and finally the overall \texttt{MMMG‐Score} along with its component sub‐scores.}
    \label{fig:enter-label}
\end{figure}
\newpage
\subsubsection{Literature}
\begin{figure}[h]
    \centering
    \includegraphics[width=\linewidth]{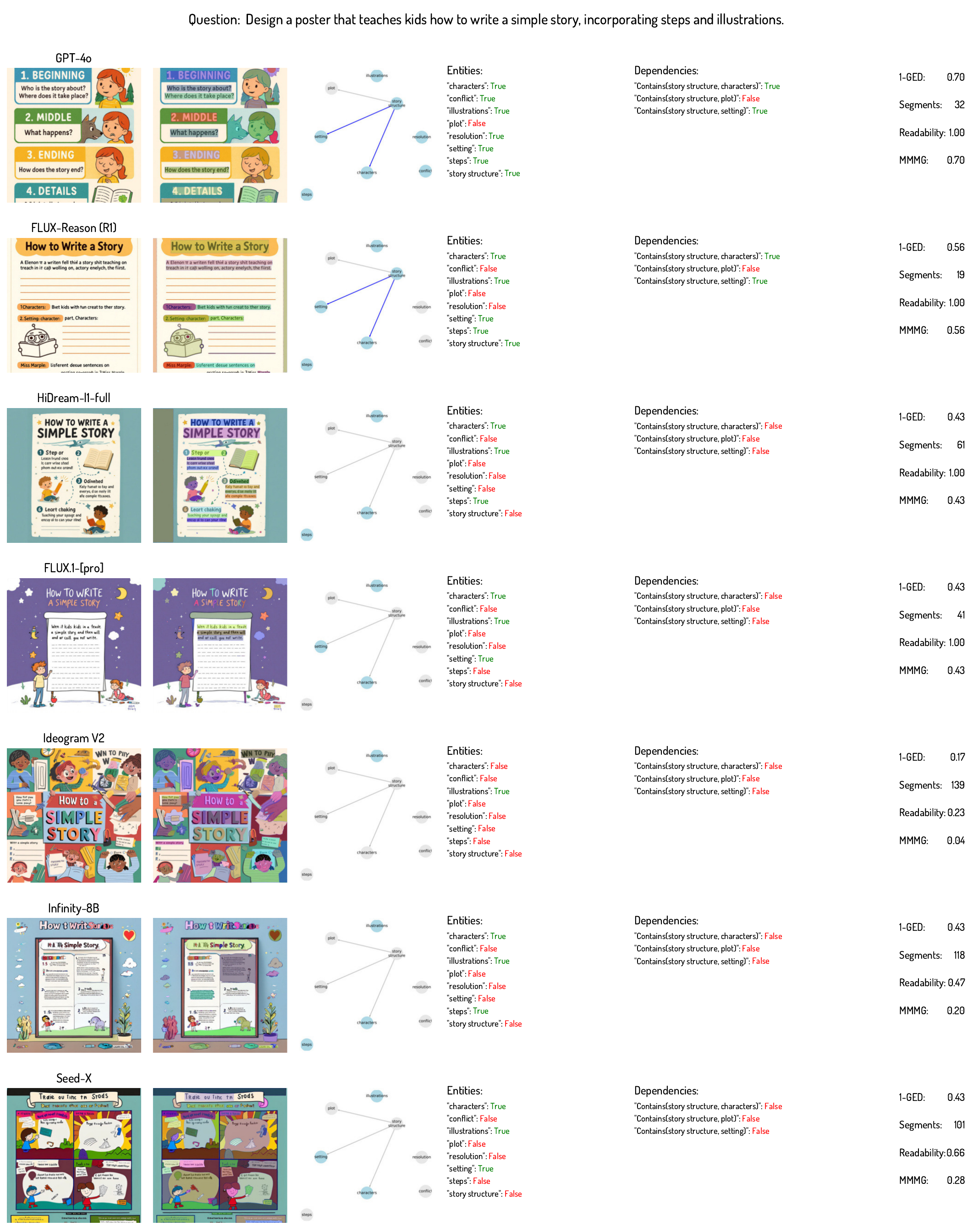}
    \caption{\texttt{MMMG} Benchmark visualization for seven representative models on a Primaryschool‐Literature example. Each row corresponds to one model and, from left to right, displays the generated image, its segmentation map, the reconstructed knowledge graph, the extracted entity and dependency lists, and finally the overall \texttt{MMMG‐Score} along with its component sub‐scores.}
    \label{fig:enter-label}
\end{figure}

\newpage
\subsection{Secondary School}

\subsubsection{Biology}
\begin{figure}[h]
    \centering
    \includegraphics[width=\linewidth]{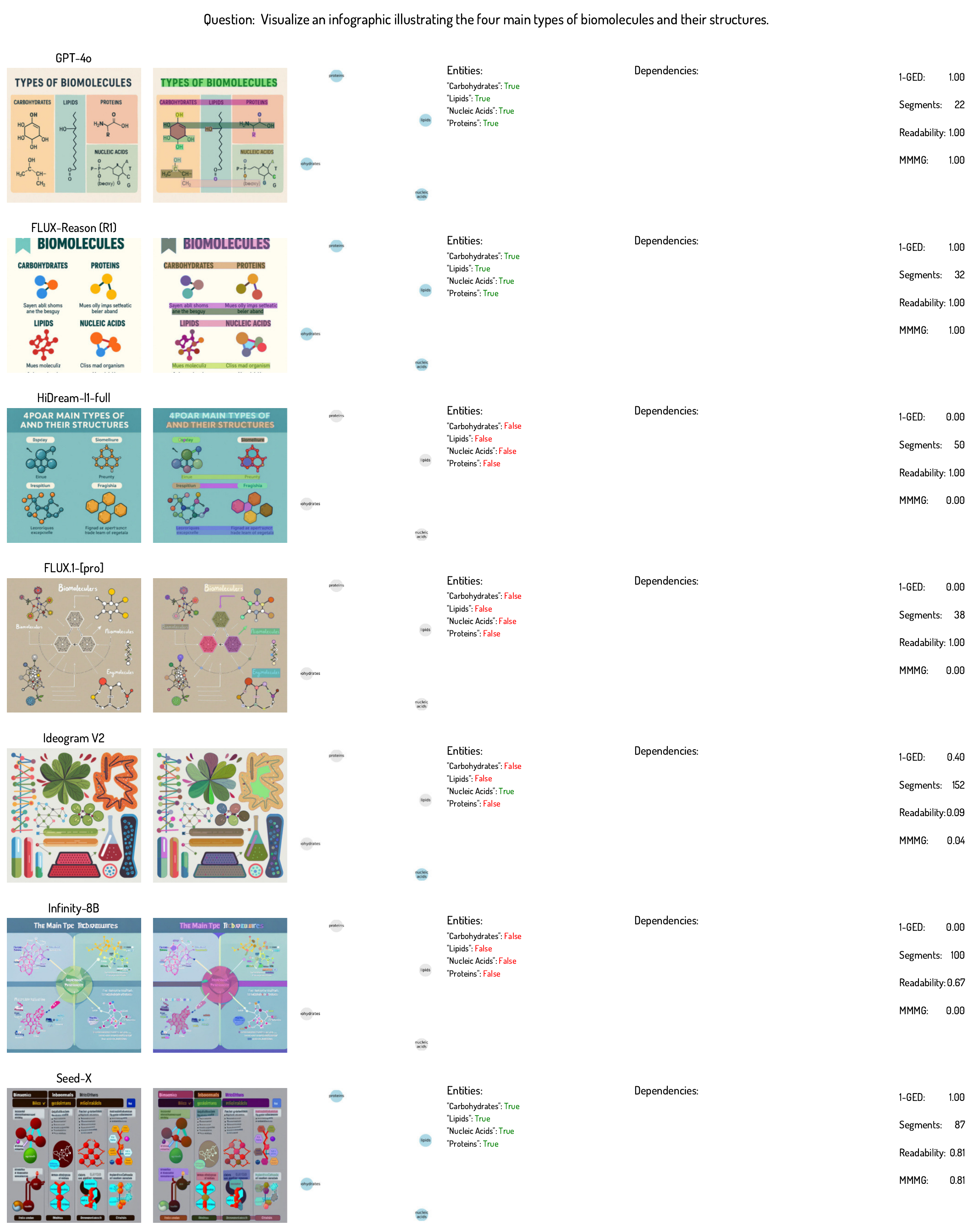}
    \caption{\texttt{MMMG} Benchmark visualization for seven representative models on a Secondaryschool‐Biology example. Each row corresponds to one model and, from left to right, displays the generated image, its segmentation map, the reconstructed knowledge graph, the extracted entity and dependency lists, and finally the overall \texttt{MMMG‐Score} along with its component sub‐scores.}
    \label{fig:enter-label}
\end{figure}
\newpage
\subsubsection{Chemistry}
\begin{figure}[h]
    \centering
    \includegraphics[width=\linewidth]{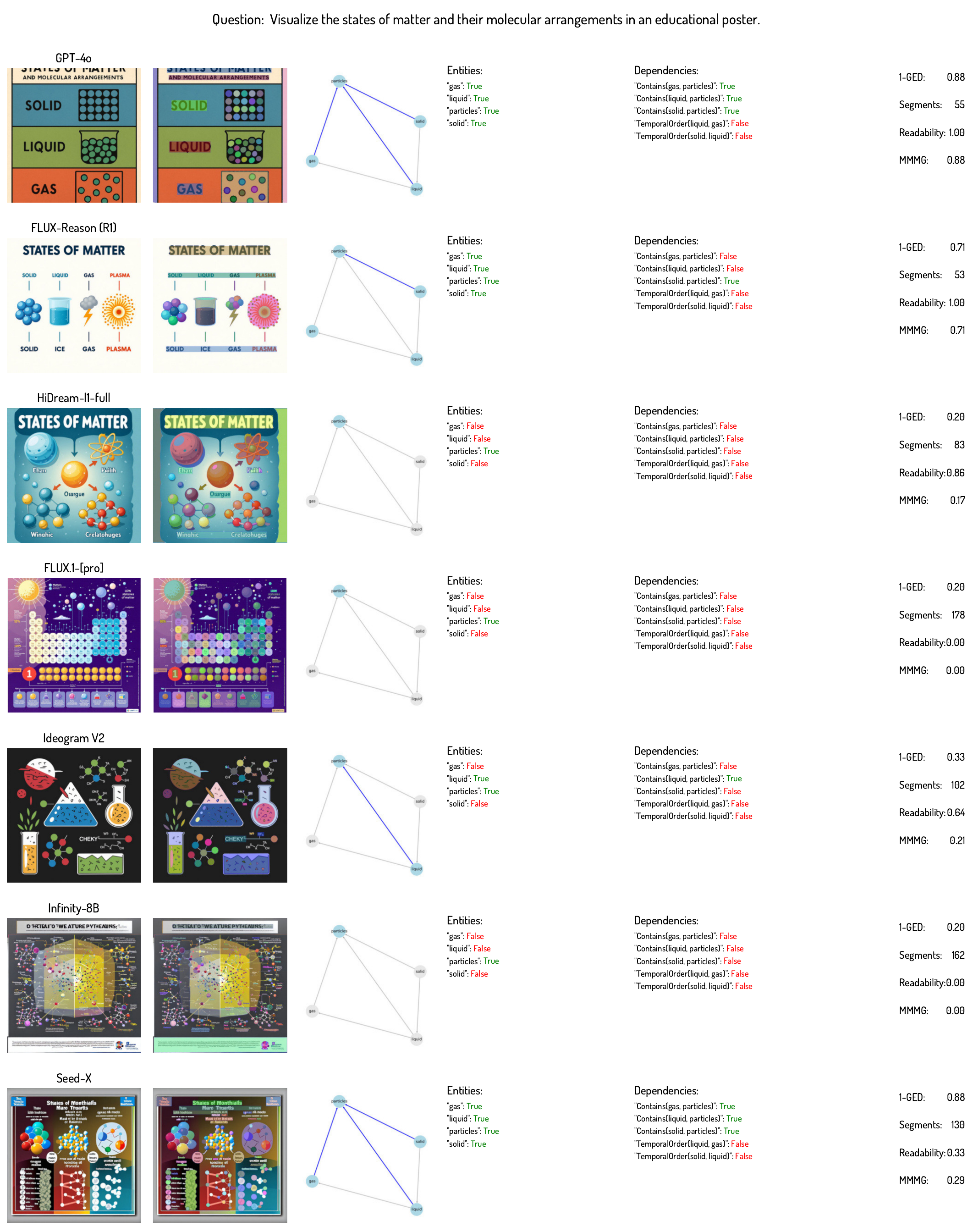}
    \caption{\texttt{MMMG} Benchmark visualization for seven representative models on a Secondaryschool‐Chemistry example. Each row corresponds to one model and, from left to right, displays the generated image, its segmentation map, the reconstructed knowledge graph, the extracted entity and dependency lists, and finally the overall \texttt{MMMG‐Score} along with its component sub‐scores.}
    \label{fig:enter-label}
\end{figure}
\newpage
\subsubsection{Mathematics}
\begin{figure}[h]
    \centering
    \includegraphics[width=\linewidth]{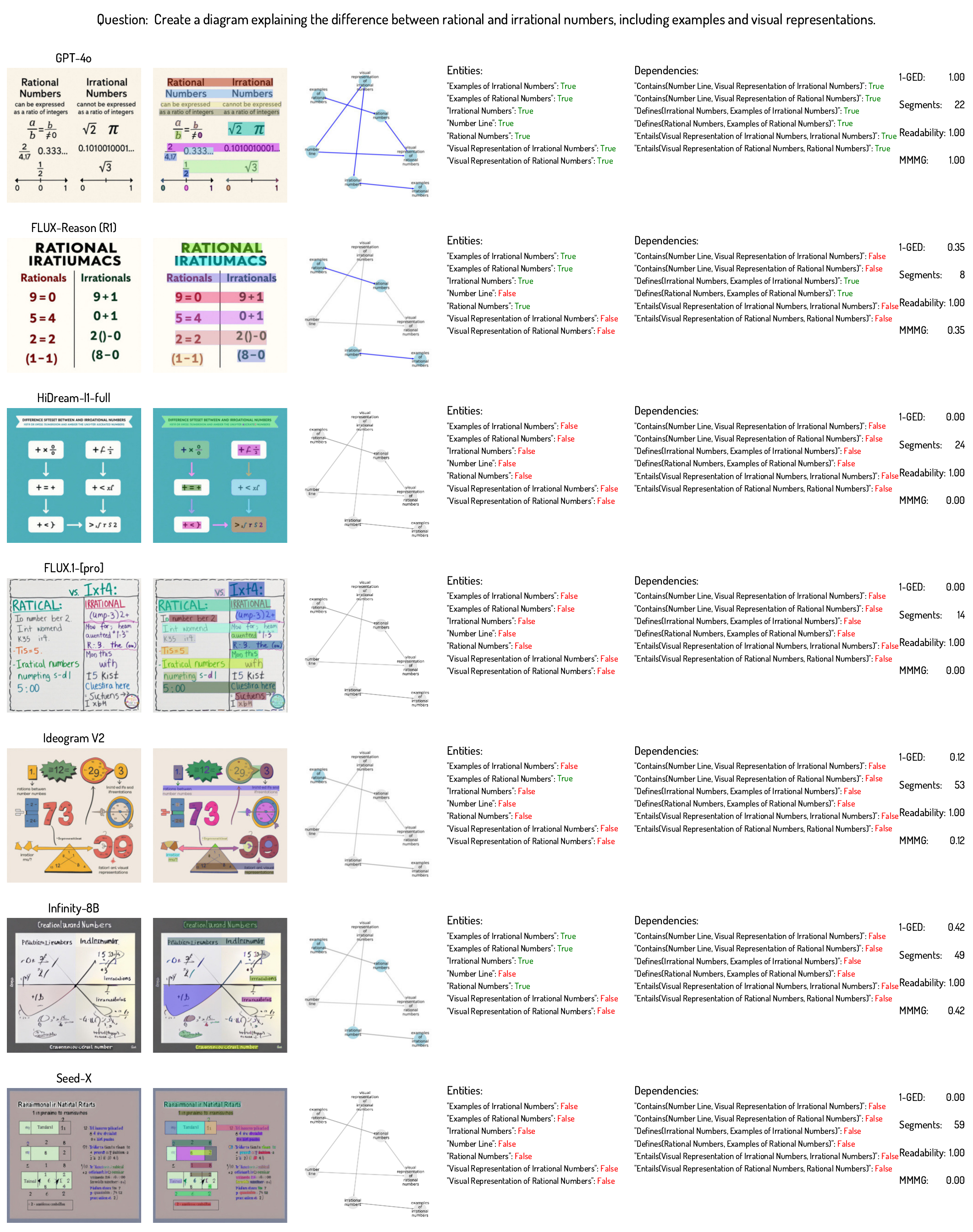}
    \caption{\texttt{MMMG} Benchmark visualization for seven representative models on a Secondaryschool‐Mathematics example. Each row corresponds to one model and, from left to right, displays the generated image, its segmentation map, the reconstructed knowledge graph, the extracted entity and dependency lists, and finally the overall \texttt{MMMG‐Score} along with its component sub‐scores.}
    \label{fig:enter-label}
\end{figure}
\newpage
\subsubsection{Engineering}
\begin{figure}[h]
    \centering
    \includegraphics[width=\linewidth]{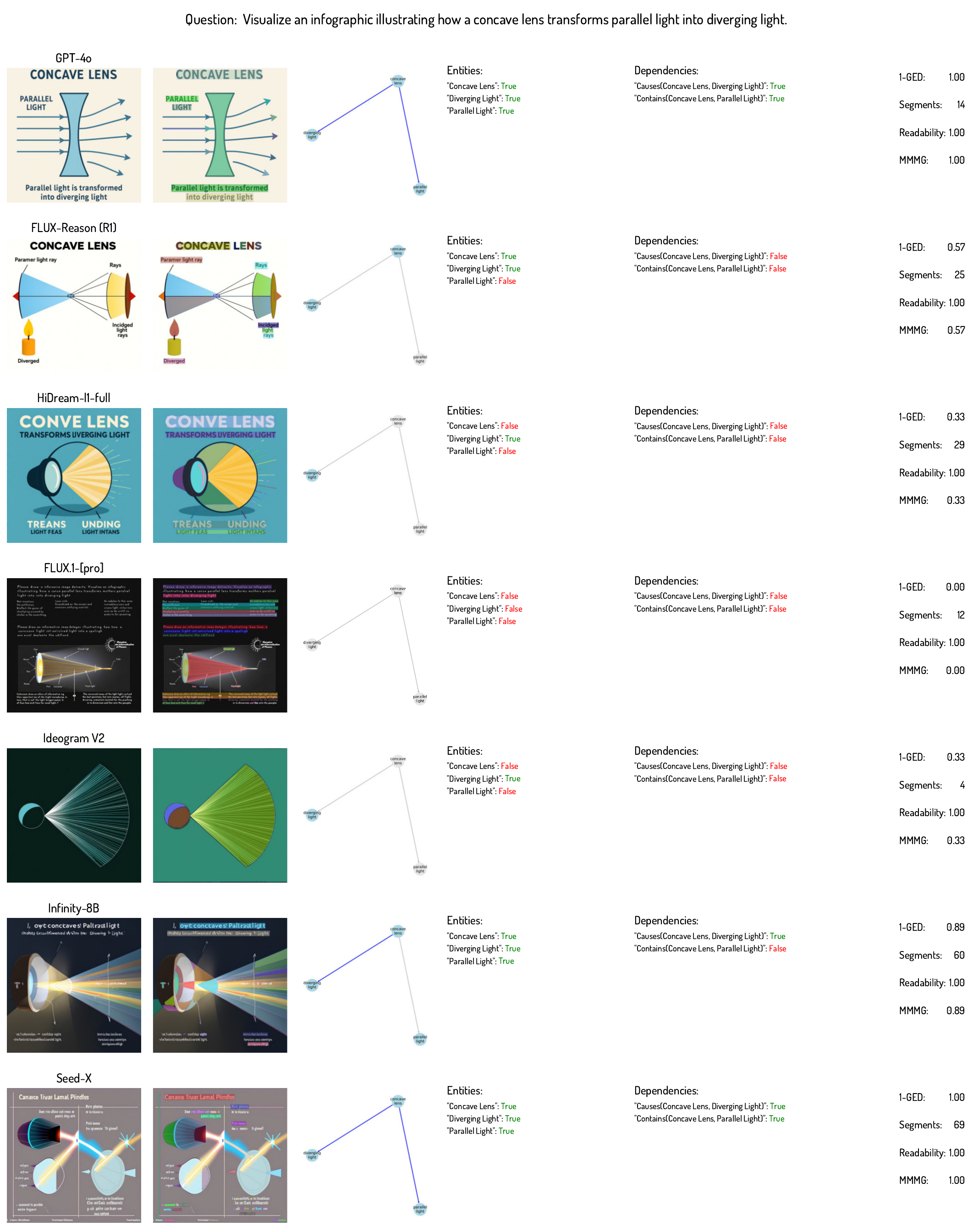}
    \caption{\texttt{MMMG} Benchmark visualization for seven representative models on a Secondaryschool‐Engineering example. Each row corresponds to one model and, from left to right, displays the generated image, its segmentation map, the reconstructed knowledge graph, the extracted entity and dependency lists, and finally the overall \texttt{MMMG‐Score} along with its component sub‐scores.}
    \label{fig:enter-label}
\end{figure}
\newpage
\subsubsection{Geography}
\begin{figure}[h]
    \centering
    \includegraphics[width=\linewidth]{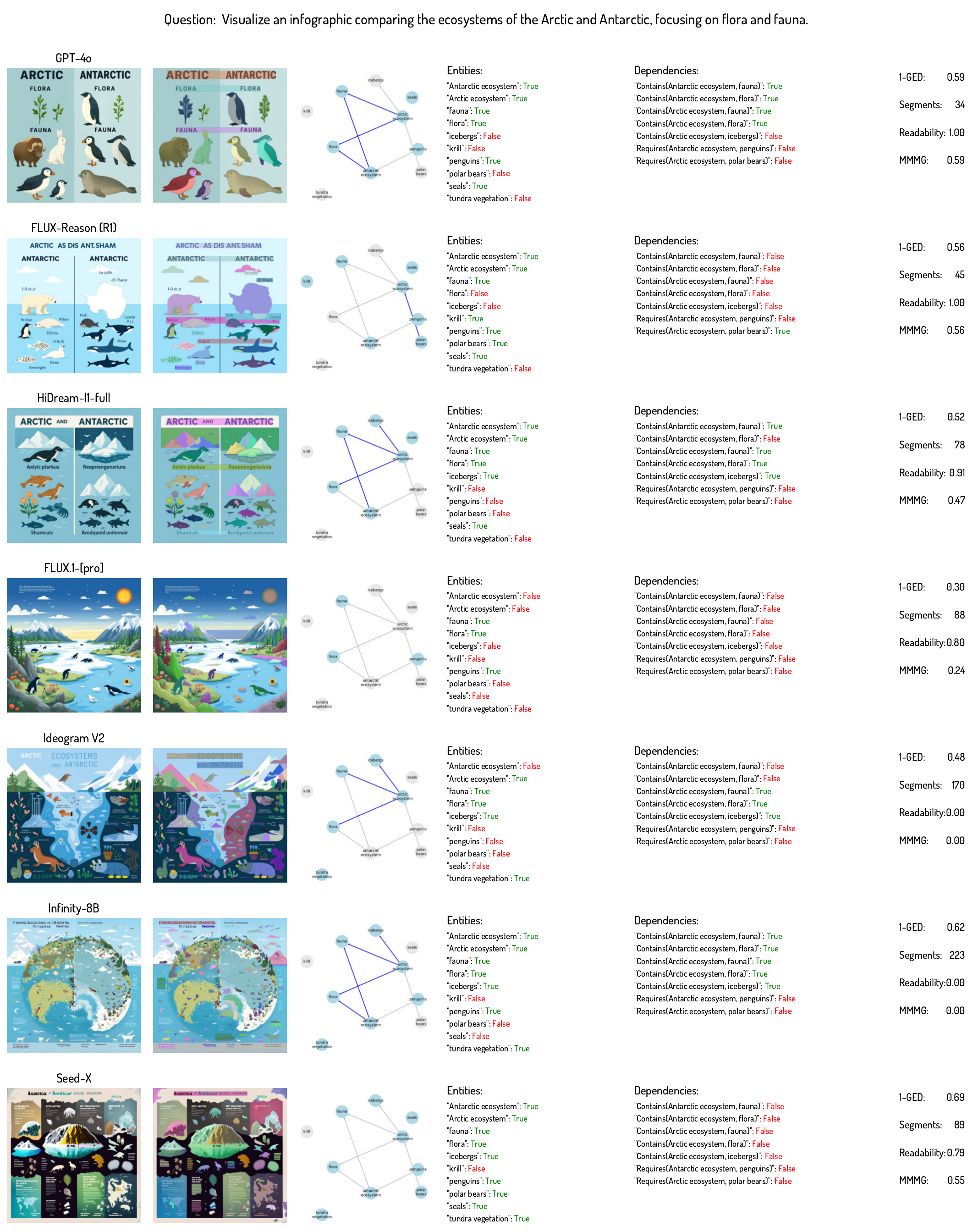}
    \caption{\texttt{MMMG} Benchmark visualization for seven representative models on a Secondaryschool‐Geography example. Each row corresponds to one model and, from left to right, displays the generated image, its segmentation map, the reconstructed knowledge graph, the extracted entity and dependency lists, and finally the overall \texttt{MMMG‐Score} along with its component sub‐scores.}
    \label{fig:enter-label}
\end{figure}
\newpage
\subsubsection{Economics}
\begin{figure}[h]
    \centering
    \includegraphics[width=\linewidth]{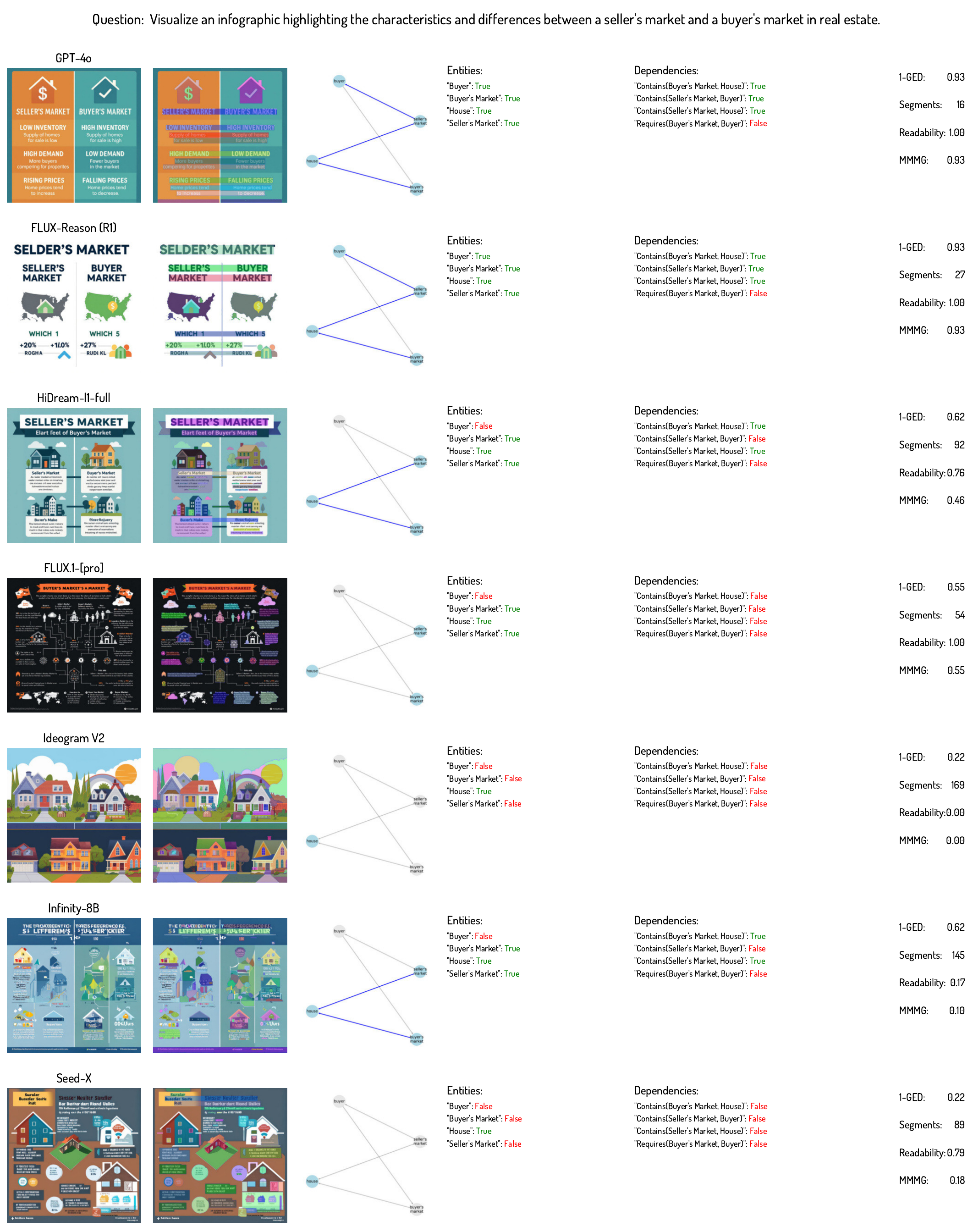}
    \caption{\texttt{MMMG} Benchmark visualization for seven representative models on a Secondaryschool‐Economics example. Each row corresponds to one model and, from left to right, displays the generated image, its segmentation map, the reconstructed knowledge graph, the extracted entity and dependency lists, and finally the overall \texttt{MMMG‐Score} along with its component sub‐scores.}
    \label{fig:enter-label}
\end{figure}
\newpage
\subsubsection{Sociology}
\begin{figure}[h]
    \centering
    \includegraphics[width=\linewidth]{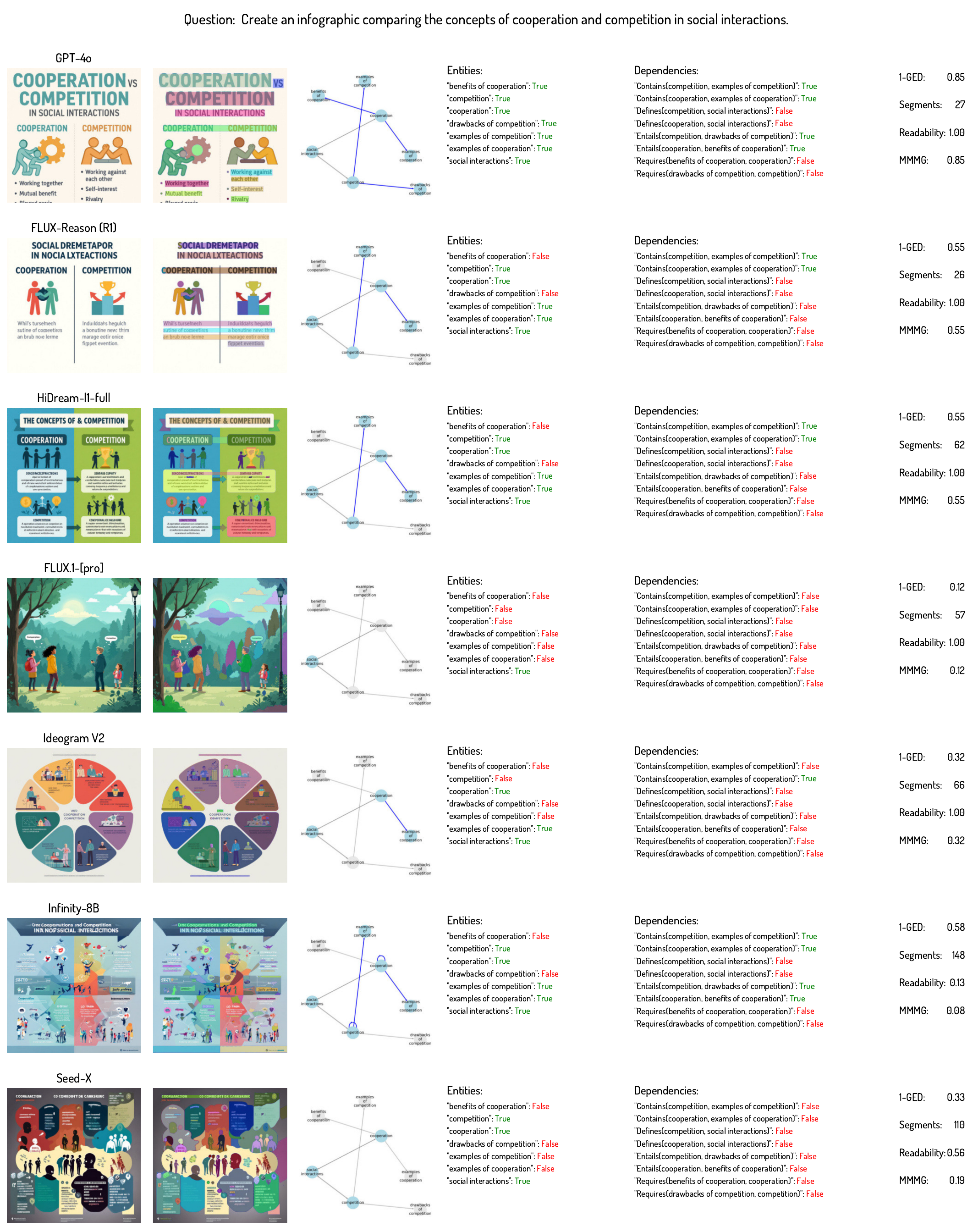}
    \caption{\texttt{MMMG} Benchmark visualization for seven representative models on a Secondaryschool‐Sociology example. Each row corresponds to one model and, from left to right, displays the generated image, its segmentation map, the reconstructed knowledge graph, the extracted entity and dependency lists, and finally the overall \texttt{MMMG‐Score} along with its component sub‐scores.}
    \label{fig:enter-label}
\end{figure}
\newpage
\subsubsection{History}
\begin{figure}[h]
    \centering
    \includegraphics[width=\linewidth]{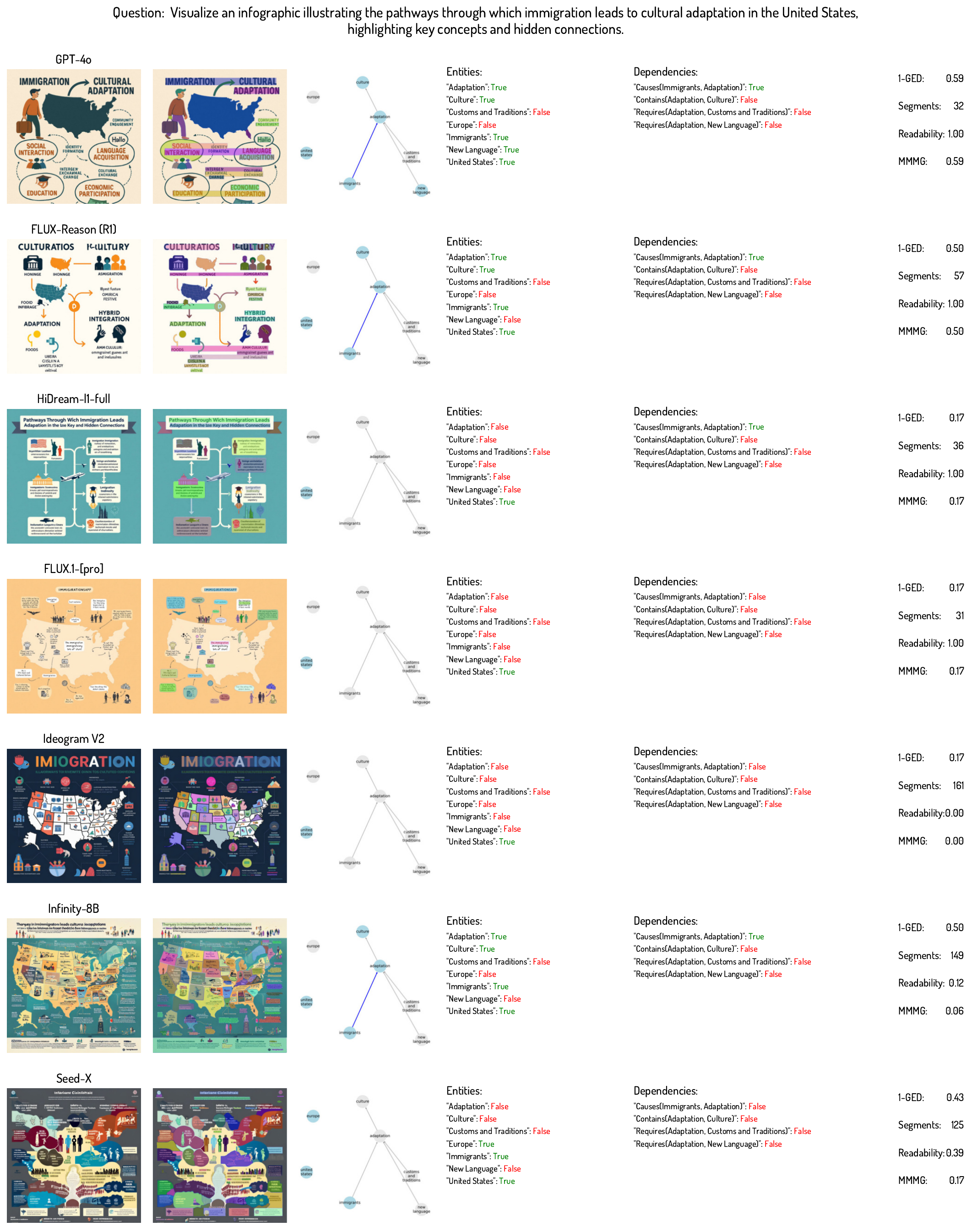}
    \caption{\texttt{MMMG} Benchmark visualization for seven representative models on a Secondaryschool‐History example. Each row corresponds to one model and, from left to right, displays the generated image, its segmentation map, the reconstructed knowledge graph, the extracted entity and dependency lists, and finally the overall \texttt{MMMG‐Score} along with its component sub‐scores.}
    \label{fig:enter-label}
\end{figure}
\newpage
\subsubsection{Philosophy}
\begin{figure}[h]
    \centering
    \includegraphics[width=\linewidth]{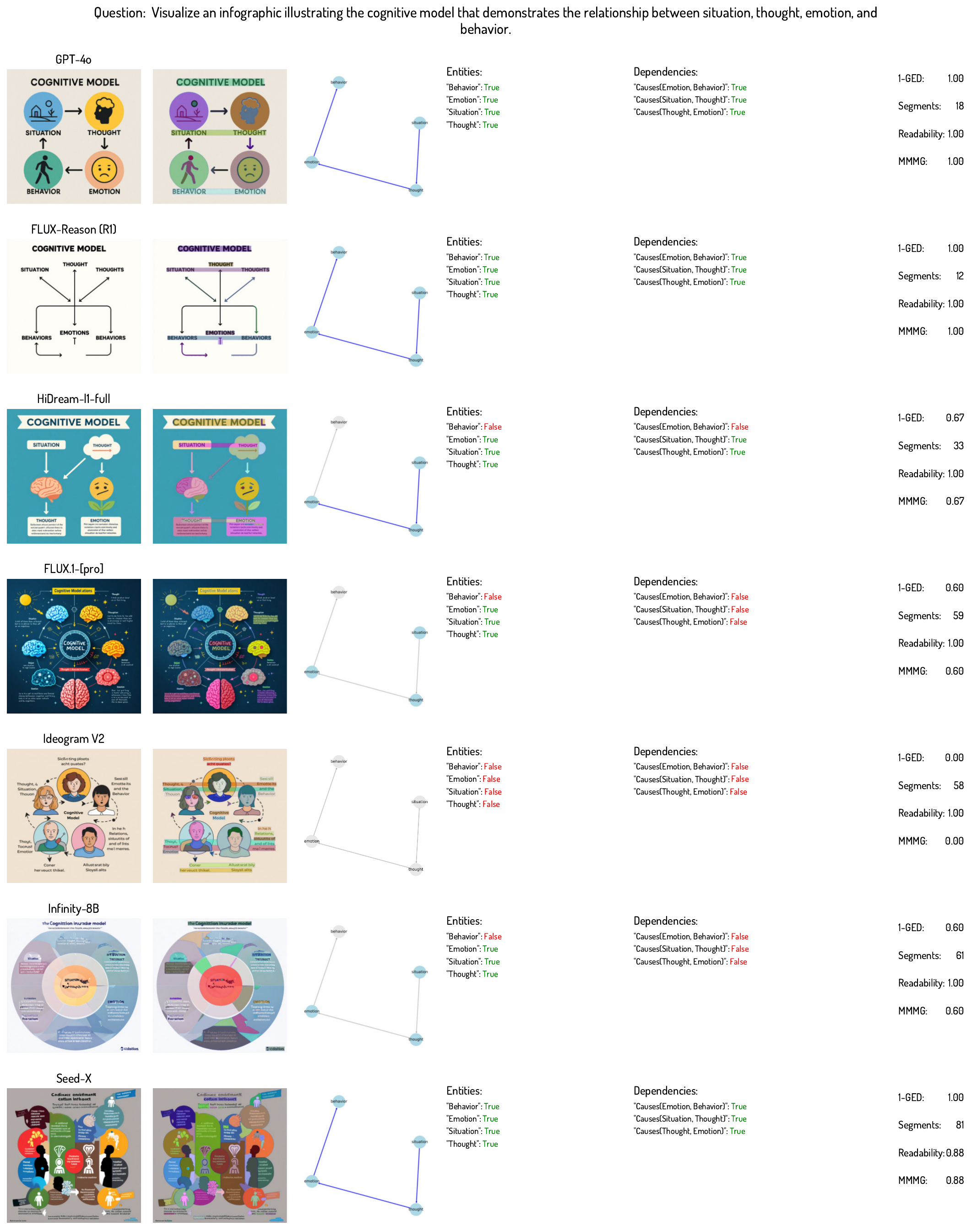}
    \caption{\texttt{MMMG} Benchmark visualization for seven representative models on a Secondaryschool‐Philosophy example. Each row corresponds to one model and, from left to right, displays the generated image, its segmentation map, the reconstructed knowledge graph, the extracted entity and dependency lists, and finally the overall \texttt{MMMG‐Score} along with its component sub‐scores.}
    \label{fig:enter-label}
\end{figure}
\newpage
\subsubsection{Literature}
\begin{figure}[h]
    \centering
    \includegraphics[width=\linewidth]{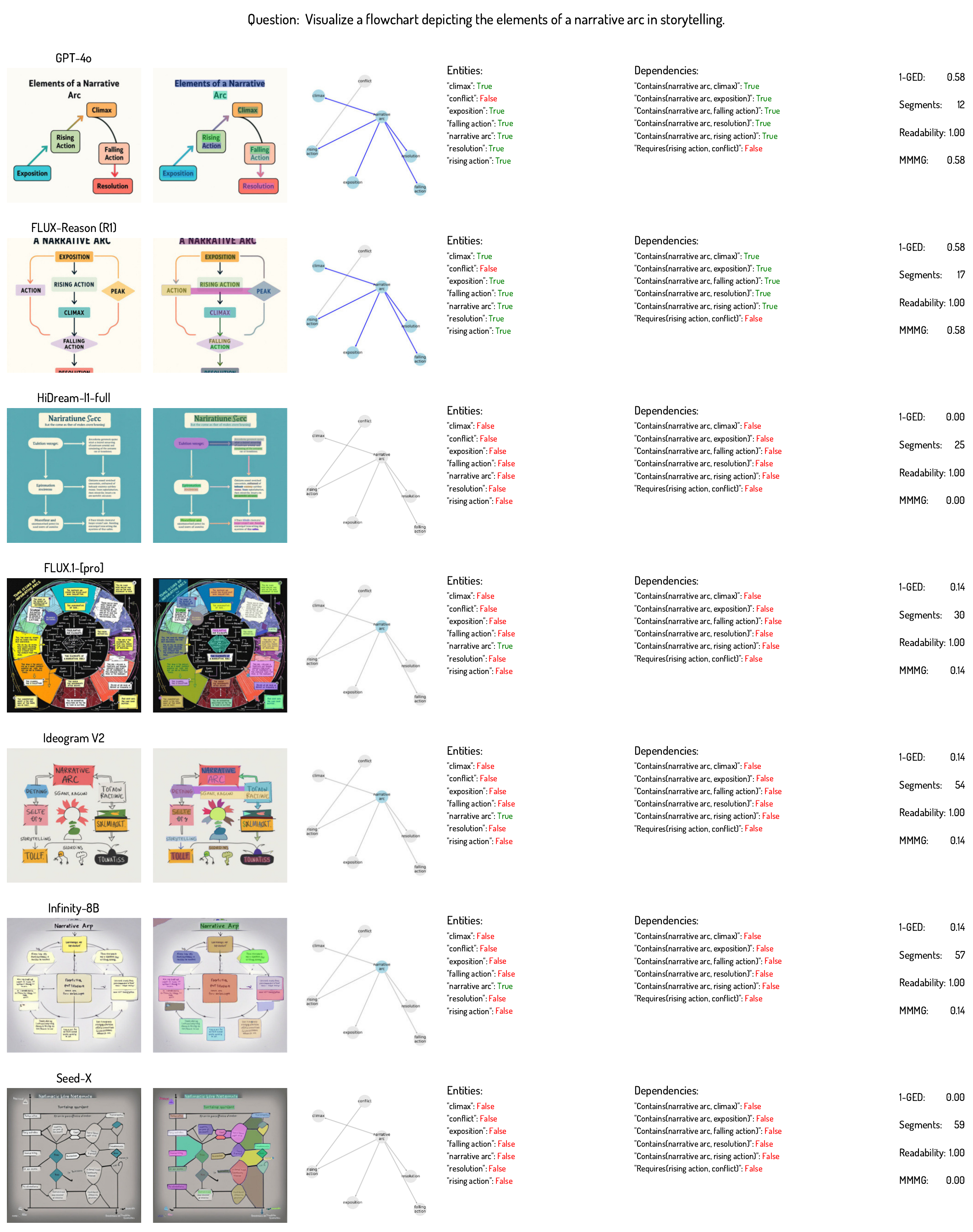}
    \caption{\texttt{MMMG} Benchmark visualization for seven representative models on a Secondaryschool‐Literature example. Each row corresponds to one model and, from left to right, displays the generated image, its segmentation map, the reconstructed knowledge graph, the extracted entity and dependency lists, and finally the overall \texttt{MMMG‐Score} along with its component sub‐scores.}
    \label{fig:enter-label}
\end{figure}

\newpage
\subsection{High School}

\subsubsection{Biology}
\begin{figure}[h]
    \centering
    \includegraphics[width=\linewidth]{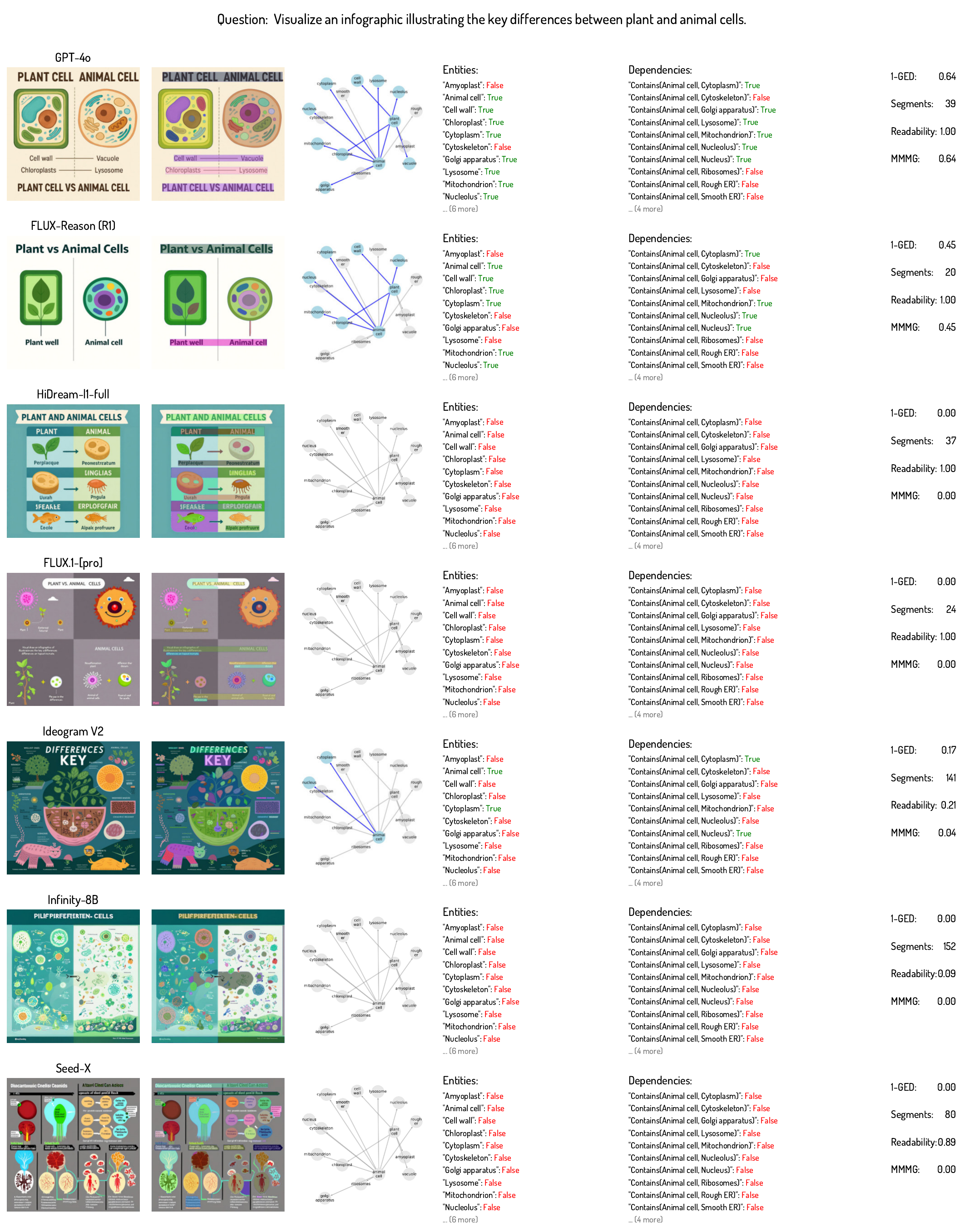}
    \caption{\texttt{MMMG} Benchmark visualization for seven representative models on a Highschool‐Biology example. Each row corresponds to one model and, from left to right, displays the generated image, its segmentation map, the reconstructed knowledge graph, the extracted entity and dependency lists, and finally the overall \texttt{MMMG‐Score} along with its component sub‐scores.}
    \label{fig:enter-label}
\end{figure}
\newpage
\subsubsection{Chemistry}
\begin{figure}[h]
    \centering
    \includegraphics[width=\linewidth]{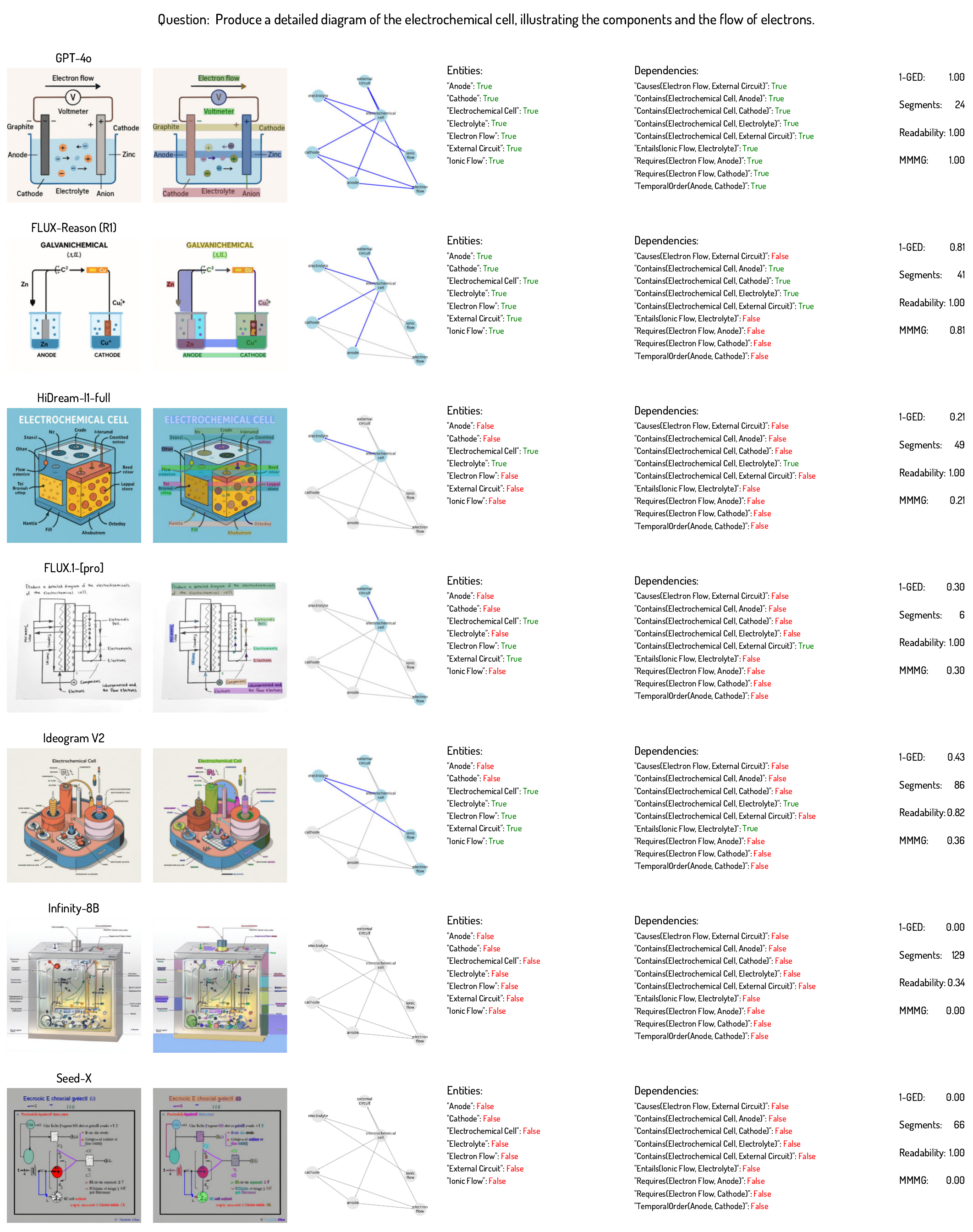}
    \caption{\texttt{MMMG} Benchmark visualization for seven representative models on a Highschool‐Chemistry example. Each row corresponds to one model and, from left to right, displays the generated image, its segmentation map, the reconstructed knowledge graph, the extracted entity and dependency lists, and finally the overall \texttt{MMMG‐Score} along with its component sub‐scores.}
    \label{fig:enter-label}
\end{figure}
\newpage
\subsubsection{Mathematics}
\begin{figure}[h]
    \centering
    \includegraphics[width=\linewidth]{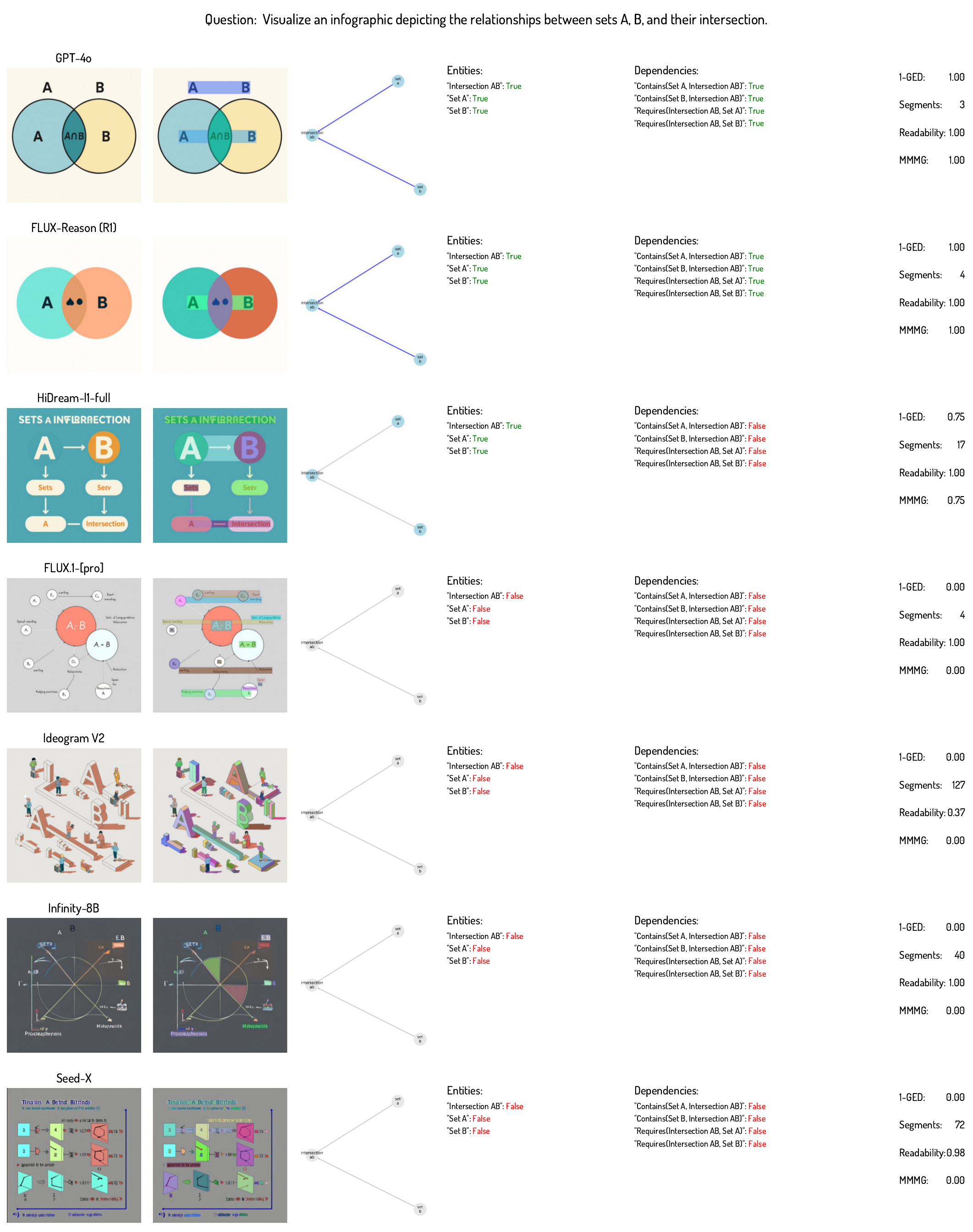}
    \caption{\texttt{MMMG} Benchmark visualization for seven representative models on a Highschool‐Mathematics example. Each row corresponds to one model and, from left to right, displays the generated image, its segmentation map, the reconstructed knowledge graph, the extracted entity and dependency lists, and finally the overall \texttt{MMMG‐Score} along with its component sub‐scores.}
    \label{fig:enter-label}
\end{figure}
\newpage
\subsubsection{Engineering}
\begin{figure}[h]
    \centering
    \includegraphics[width=\linewidth]{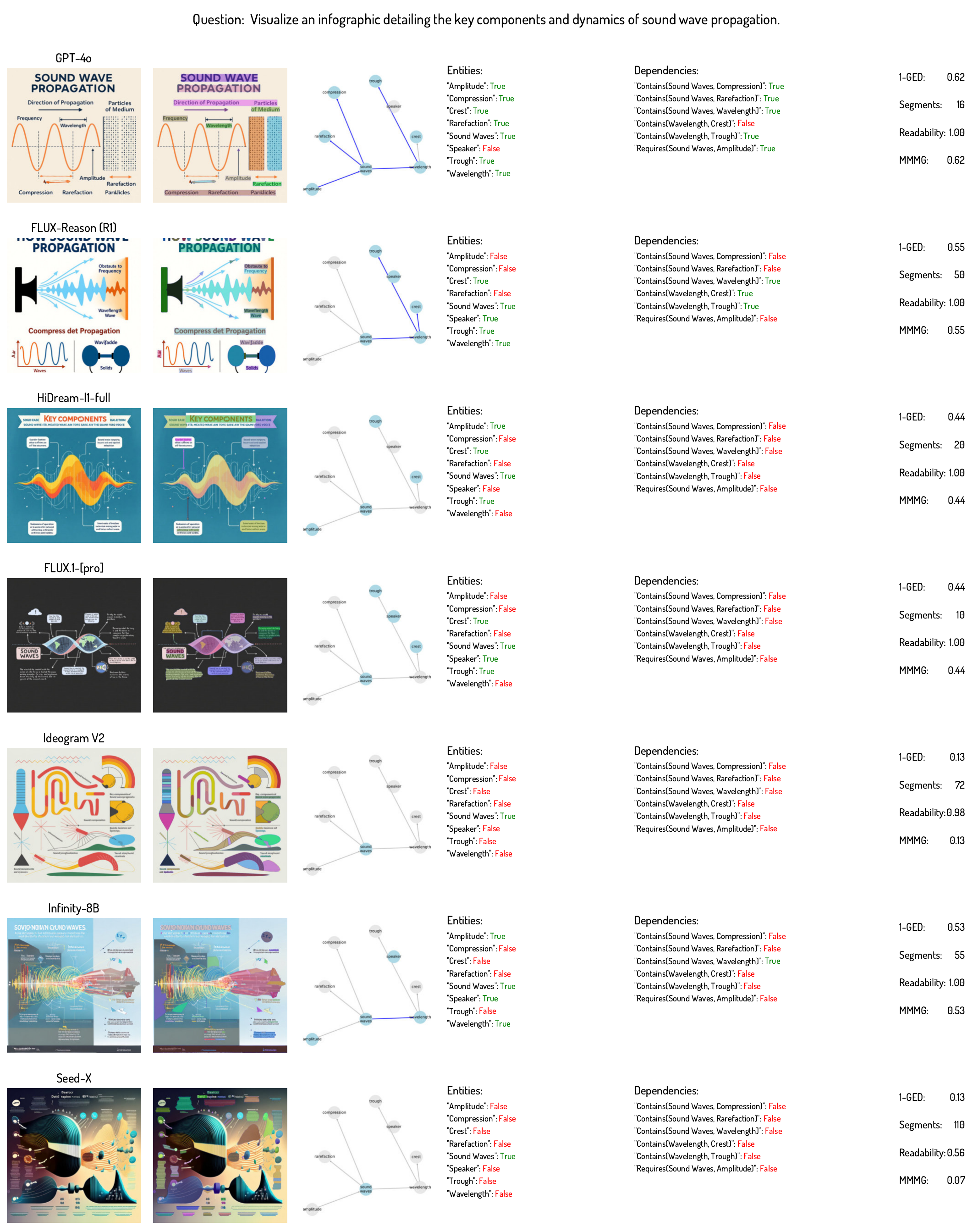}
    \caption{\texttt{MMMG} Benchmark visualization for seven representative models on a Highschool‐Engineering example. Each row corresponds to one model and, from left to right, displays the generated image, its segmentation map, the reconstructed knowledge graph, the extracted entity and dependency lists, and finally the overall \texttt{MMMG‐Score} along with its component sub‐scores.}
    \label{fig:enter-label}
\end{figure}
\newpage
\subsubsection{Geography}
\begin{figure}[h]
    \centering
    \includegraphics[width=\linewidth]{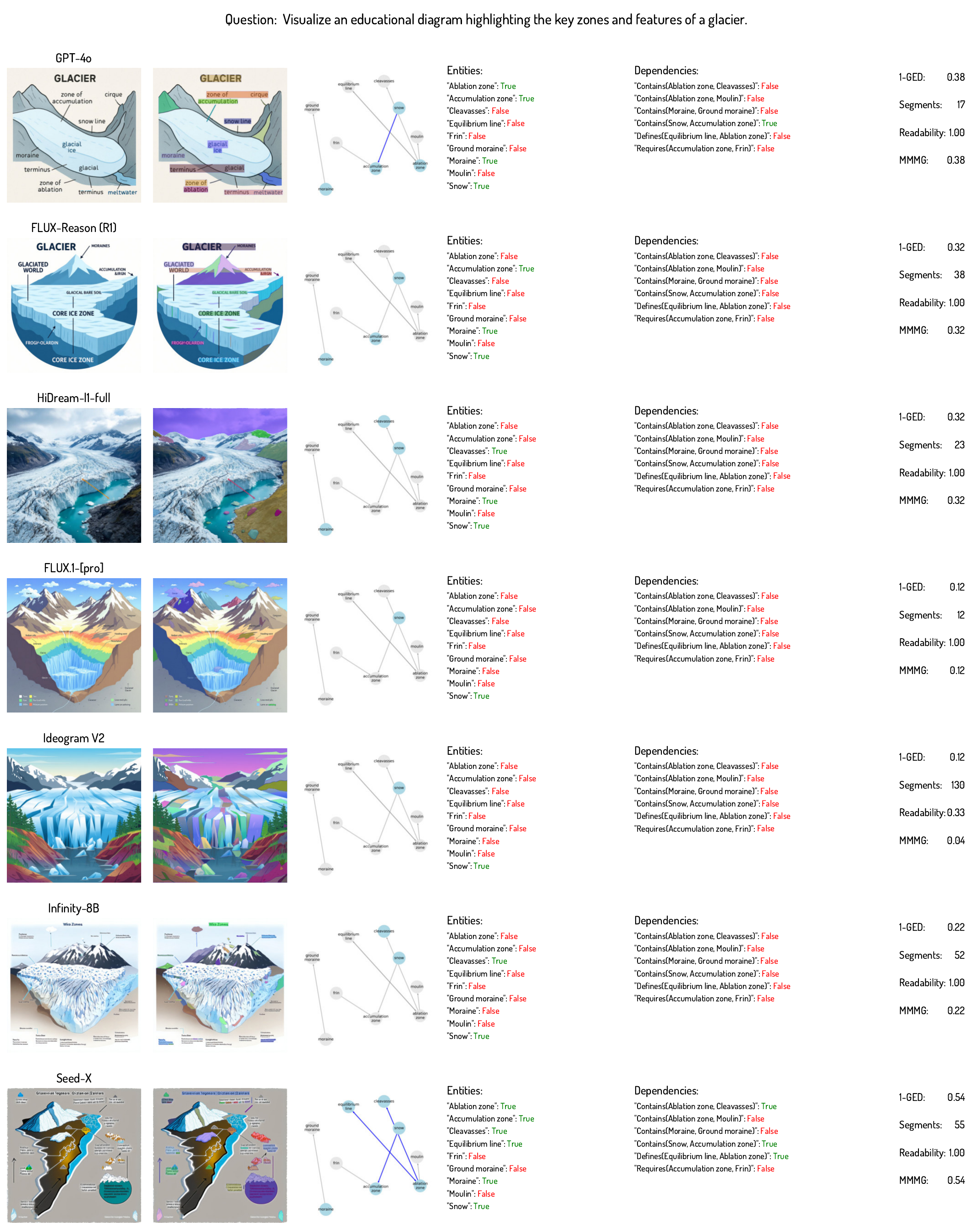}
    \caption{\texttt{MMMG} Benchmark visualization for seven representative models on a Highschool‐Geography example. Each row corresponds to one model and, from left to right, displays the generated image, its segmentation map, the reconstructed knowledge graph, the extracted entity and dependency lists, and finally the overall \texttt{MMMG‐Score} along with its component sub‐scores.}
    \label{fig:enter-label}
\end{figure}
\newpage
\subsubsection{Economics}
\begin{figure}[h]
    \centering
    \includegraphics[width=\linewidth]{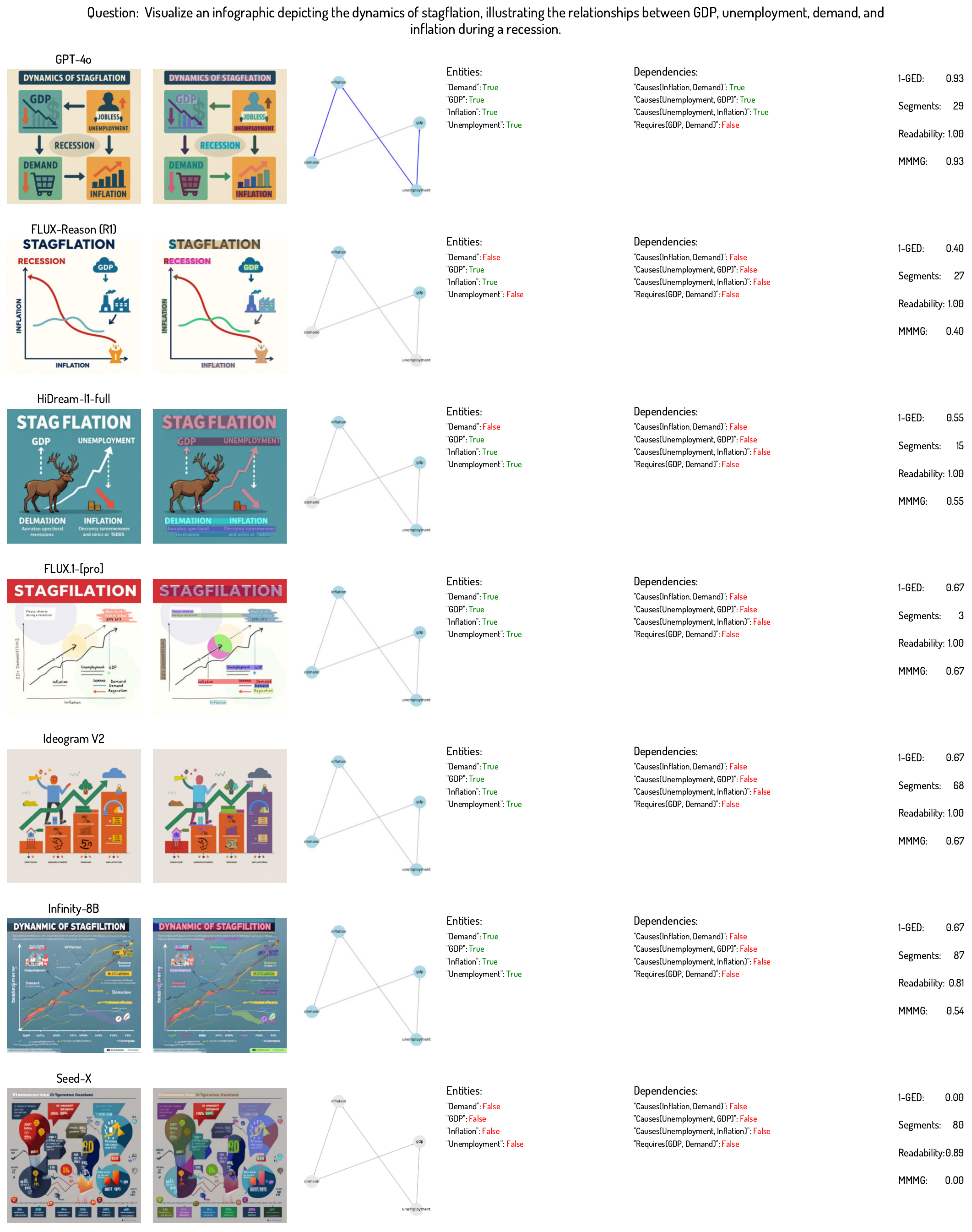}
    \caption{\texttt{MMMG} Benchmark visualization for seven representative models on a Highschool‐Economics example. Each row corresponds to one model and, from left to right, displays the generated image, its segmentation map, the reconstructed knowledge graph, the extracted entity and dependency lists, and finally the overall \texttt{MMMG‐Score} along with its component sub‐scores.}
    \label{fig:enter-label}
\end{figure}
\newpage
\subsubsection{Sociology}
\begin{figure}[h]
    \centering
    \includegraphics[width=\linewidth]{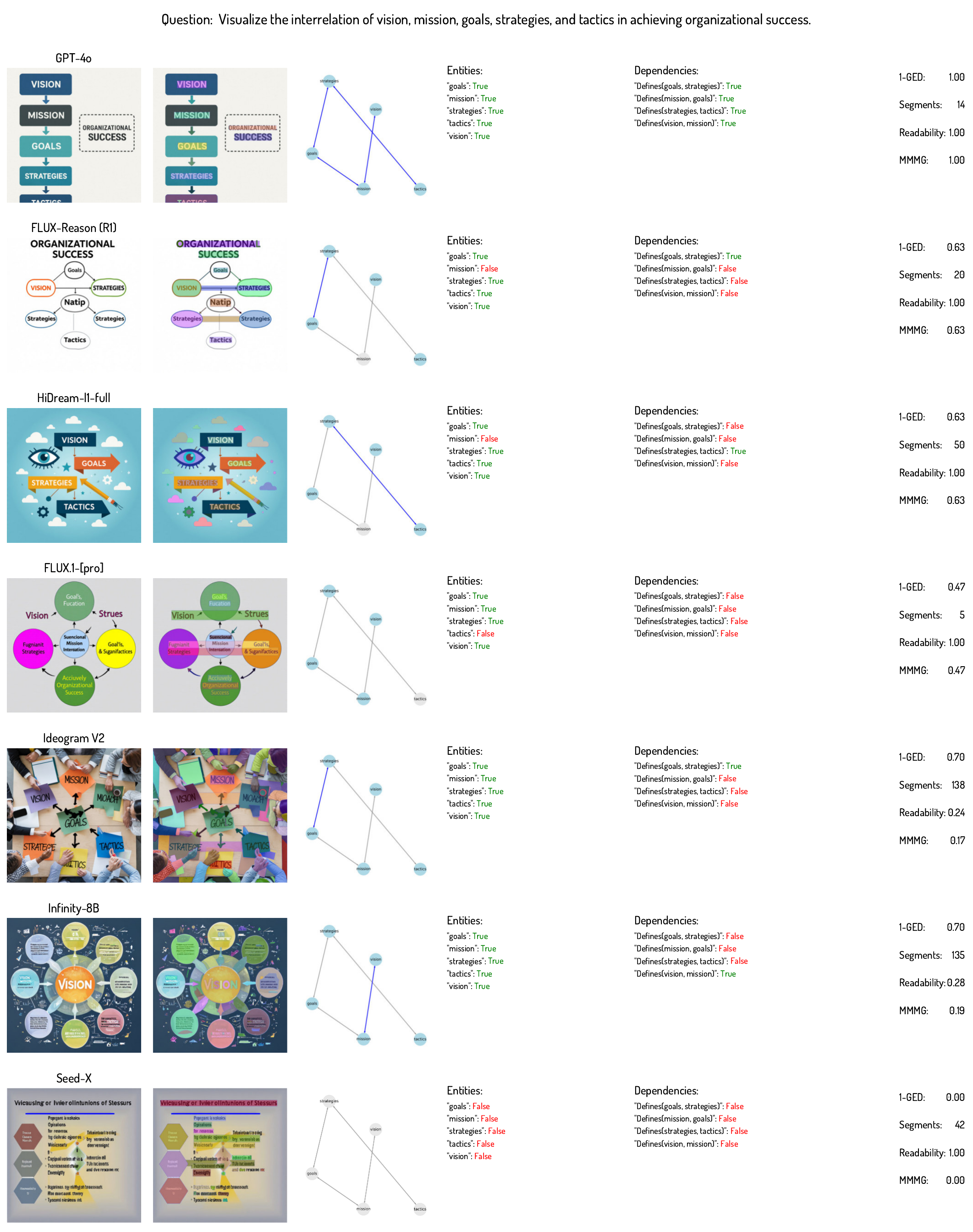}
    \caption{\texttt{MMMG} Benchmark visualization for seven representative models on a Highschool‐Sociology example. Each row corresponds to one model and, from left to right, displays the generated image, its segmentation map, the reconstructed knowledge graph, the extracted entity and dependency lists, and finally the overall \texttt{MMMG‐Score} along with its component sub‐scores.}
    \label{fig:enter-label}
\end{figure}
\newpage
\subsubsection{History}
\begin{figure}[h]
    \centering
    \includegraphics[width=\linewidth]{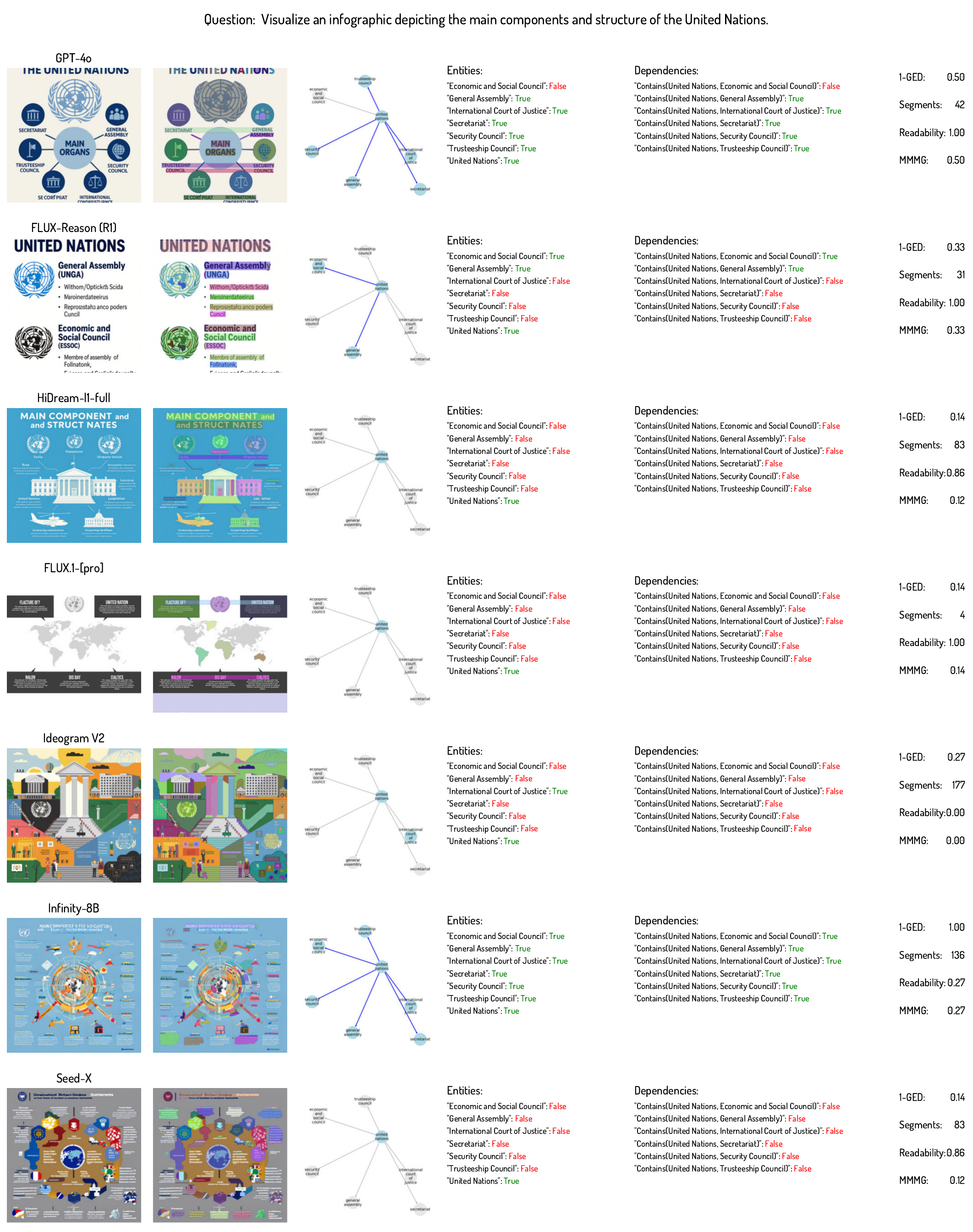}
    \caption{\texttt{MMMG} Benchmark visualization for seven representative models on a Highschool‐History example. Each row corresponds to one model and, from left to right, displays the generated image, its segmentation map, the reconstructed knowledge graph, the extracted entity and dependency lists, and finally the overall \texttt{MMMG‐Score} along with its component sub‐scores.}
    \label{fig:enter-label}
\end{figure}
\newpage
\subsubsection{Philosophy}
\begin{figure}[h]
    \centering
    \includegraphics[width=\linewidth]{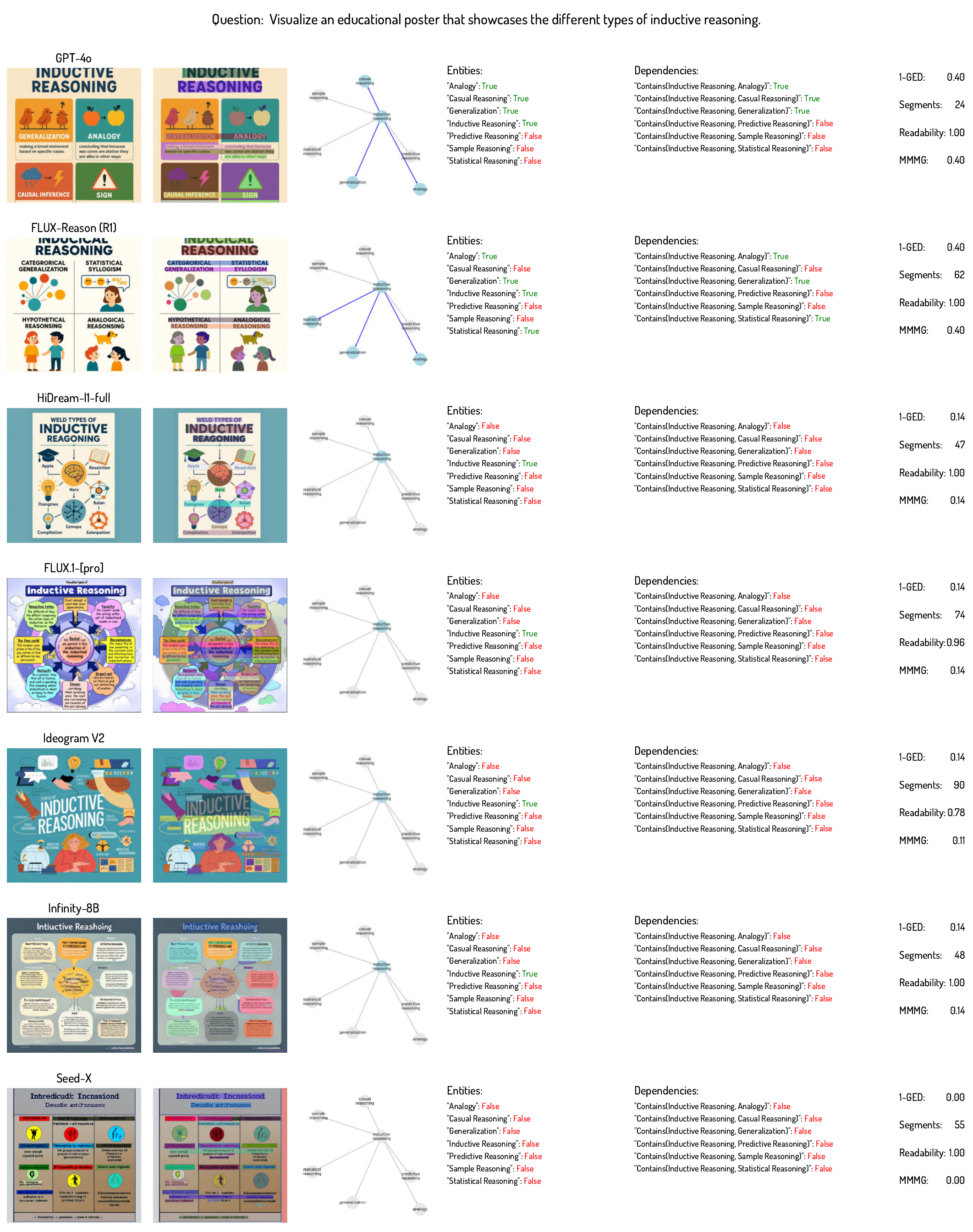}
    \caption{\texttt{MMMG} Benchmark visualization for seven representative models on a Highschool‐Philosophy example. Each row corresponds to one model and, from left to right, displays the generated image, its segmentation map, the reconstructed knowledge graph, the extracted entity and dependency lists, and finally the overall \texttt{MMMG‐Score} along with its component sub‐scores.}
    \label{fig:enter-label}
\end{figure}
\newpage
\subsubsection{Literature}
\begin{figure}[h]
    \centering
    \includegraphics[width=\linewidth]{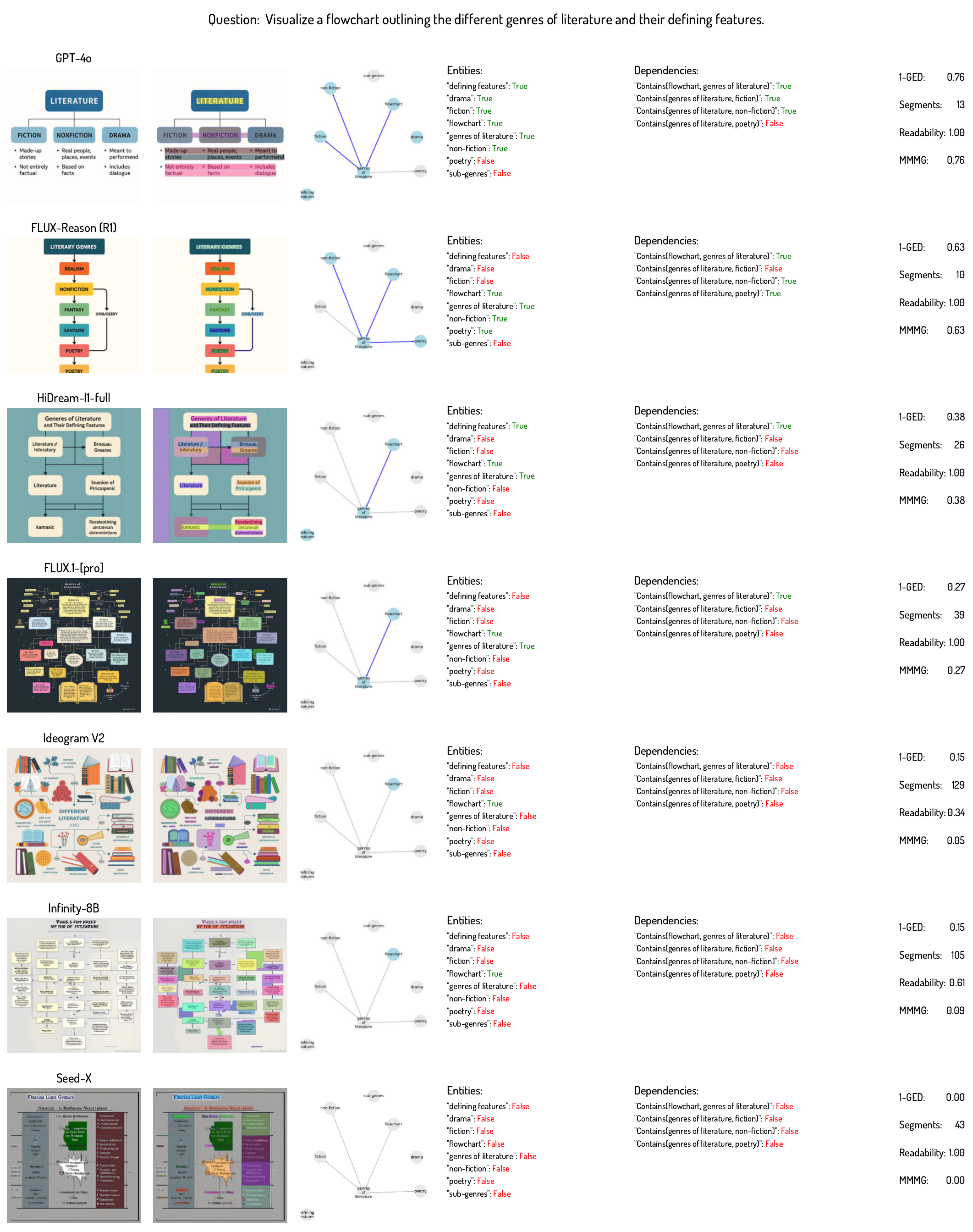}
    \caption{\texttt{MMMG} Benchmark visualization for seven representative models on a Highschool‐Literature example. Each row corresponds to one model and, from left to right, displays the generated image, its segmentation map, the reconstructed knowledge graph, the extracted entity and dependency lists, and finally the overall \texttt{MMMG‐Score} along with its component sub‐scores.}
    \label{fig:enter-label}
\end{figure}

\newpage
\subsection{Undergraduate}

\subsubsection{Biology}
\begin{figure}[h]
    \centering
    \includegraphics[width=\linewidth]{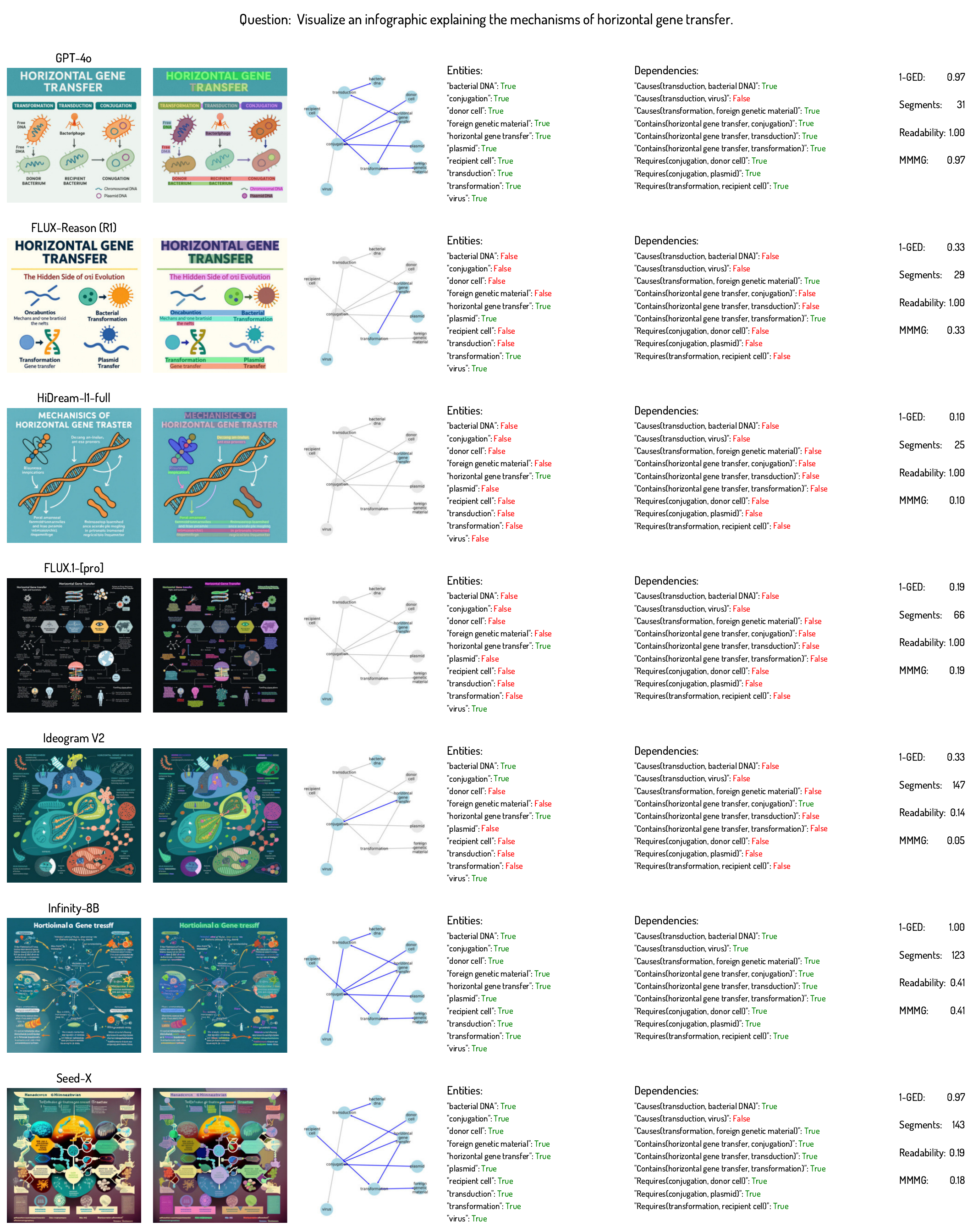}
    \caption{\texttt{MMMG} Benchmark visualization for seven representative models on a Undergraduate‐Biology example. Each row corresponds to one model and, from left to right, displays the generated image, its segmentation map, the reconstructed knowledge graph, the extracted entity and dependency lists, and finally the overall \texttt{MMMG‐Score} along with its component sub‐scores.}
    \label{fig:enter-label}
\end{figure}
\newpage
\subsubsection{Chemistry}
\begin{figure}[h]
    \centering
    \includegraphics[width=\linewidth]{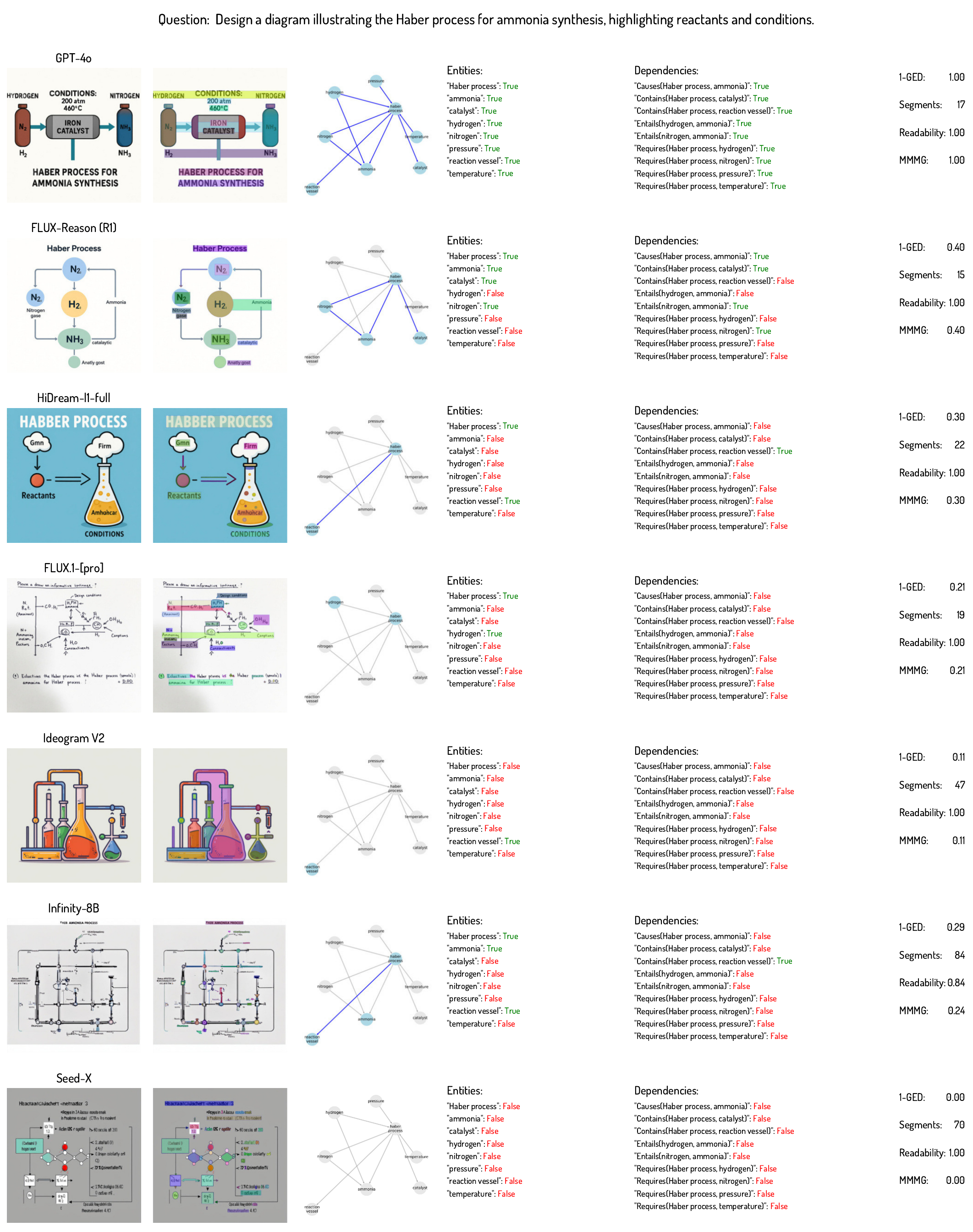}
    \caption{\texttt{MMMG} Benchmark visualization for seven representative models on a Undergraduate‐Chemistry example. Each row corresponds to one model and, from left to right, displays the generated image, its segmentation map, the reconstructed knowledge graph, the extracted entity and dependency lists, and finally the overall \texttt{MMMG‐Score} along with its component sub‐scores.}
    \label{fig:enter-label}
\end{figure}
\newpage
\subsubsection{Mathematics}
\begin{figure}[h]
    \centering
    \includegraphics[width=\linewidth]{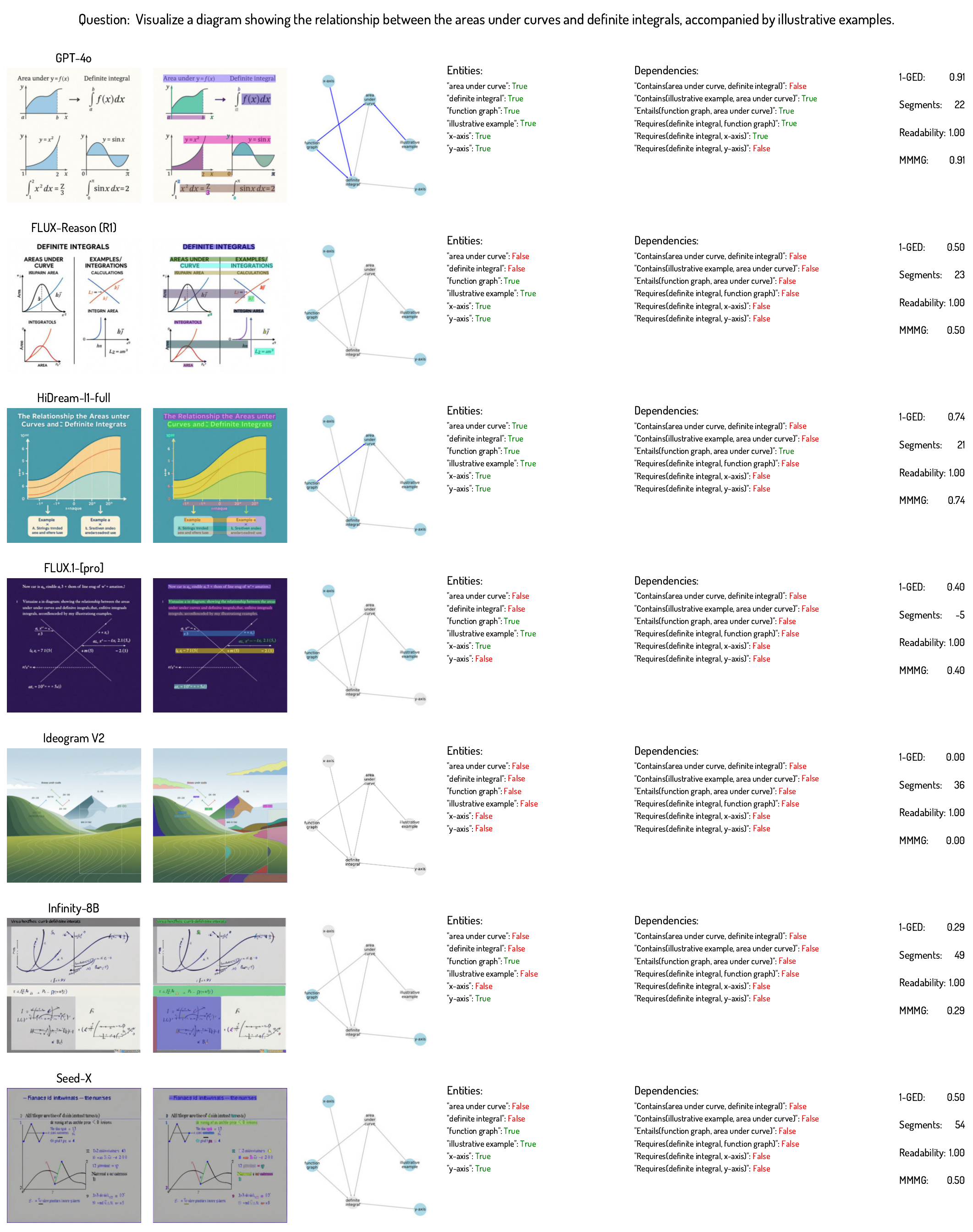}
    \caption{\texttt{MMMG} Benchmark visualization for seven representative models on a Undergraduate‐Mathematics example. Each row corresponds to one model and, from left to right, displays the generated image, its segmentation map, the reconstructed knowledge graph, the extracted entity and dependency lists, and finally the overall \texttt{MMMG‐Score} along with its component sub‐scores.}
    \label{fig:enter-label}
\end{figure}
\newpage
\subsubsection{Engineering}
\begin{figure}[h]
    \centering
    \includegraphics[width=\linewidth]{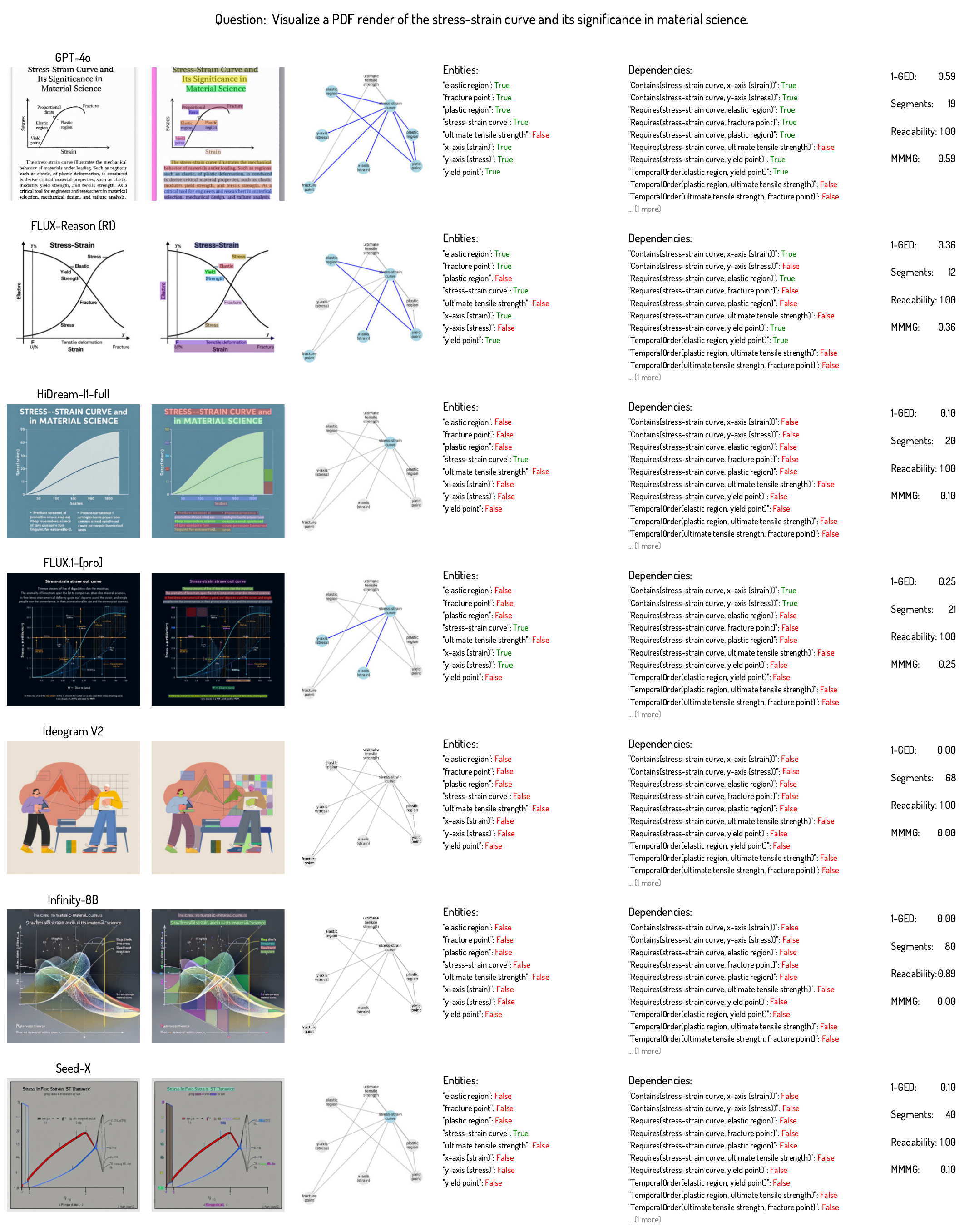}
    \caption{\texttt{MMMG} Benchmark visualization for seven representative models on a Undergraduate‐Engineering example. Each row corresponds to one model and, from left to right, displays the generated image, its segmentation map, the reconstructed knowledge graph, the extracted entity and dependency lists, and finally the overall \texttt{MMMG‐Score} along with its component sub‐scores.}
    \label{fig:enter-label}
\end{figure}
\newpage
\subsubsection{Geography}
\begin{figure}[h]
    \centering
    \includegraphics[width=\linewidth]{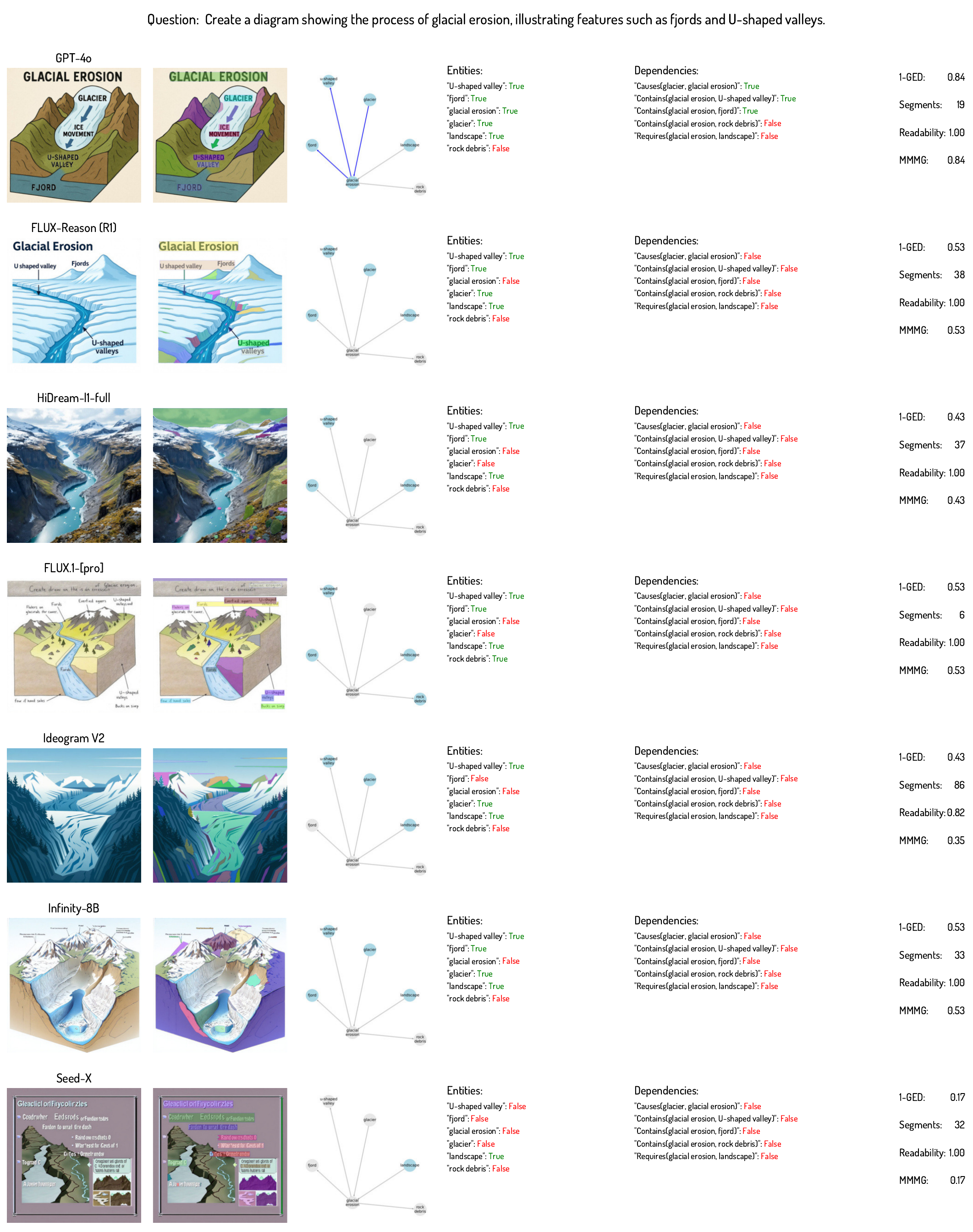}
    \caption{\texttt{MMMG} Benchmark visualization for seven representative models on a Undergraduate‐Geography example. Each row corresponds to one model and, from left to right, displays the generated image, its segmentation map, the reconstructed knowledge graph, the extracted entity and dependency lists, and finally the overall \texttt{MMMG‐Score} along with its component sub‐scores.}
    \label{fig:enter-label}
\end{figure}
\newpage
\subsubsection{Economics}
\begin{figure}[h]
    \centering
    \includegraphics[width=\linewidth]{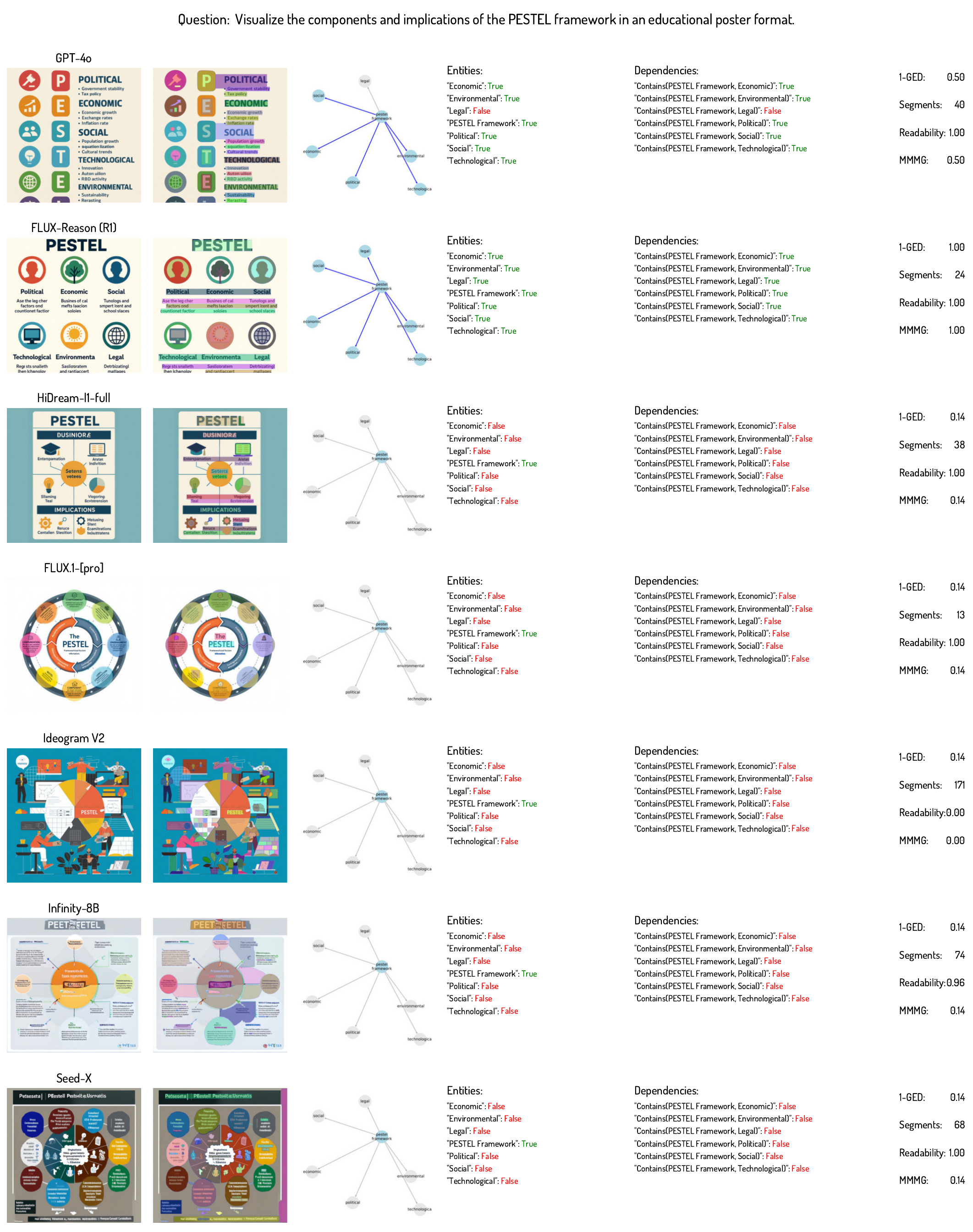}
    \caption{\texttt{MMMG} Benchmark visualization for seven representative models on a Undergraduate‐Economics example. Each row corresponds to one model and, from left to right, displays the generated image, its segmentation map, the reconstructed knowledge graph, the extracted entity and dependency lists, and finally the overall \texttt{MMMG‐Score} along with its component sub‐scores.}
    \label{fig:enter-label}
\end{figure}
\newpage
\subsubsection{Sociology}
\begin{figure}[h]
    \centering
    \includegraphics[width=\linewidth]{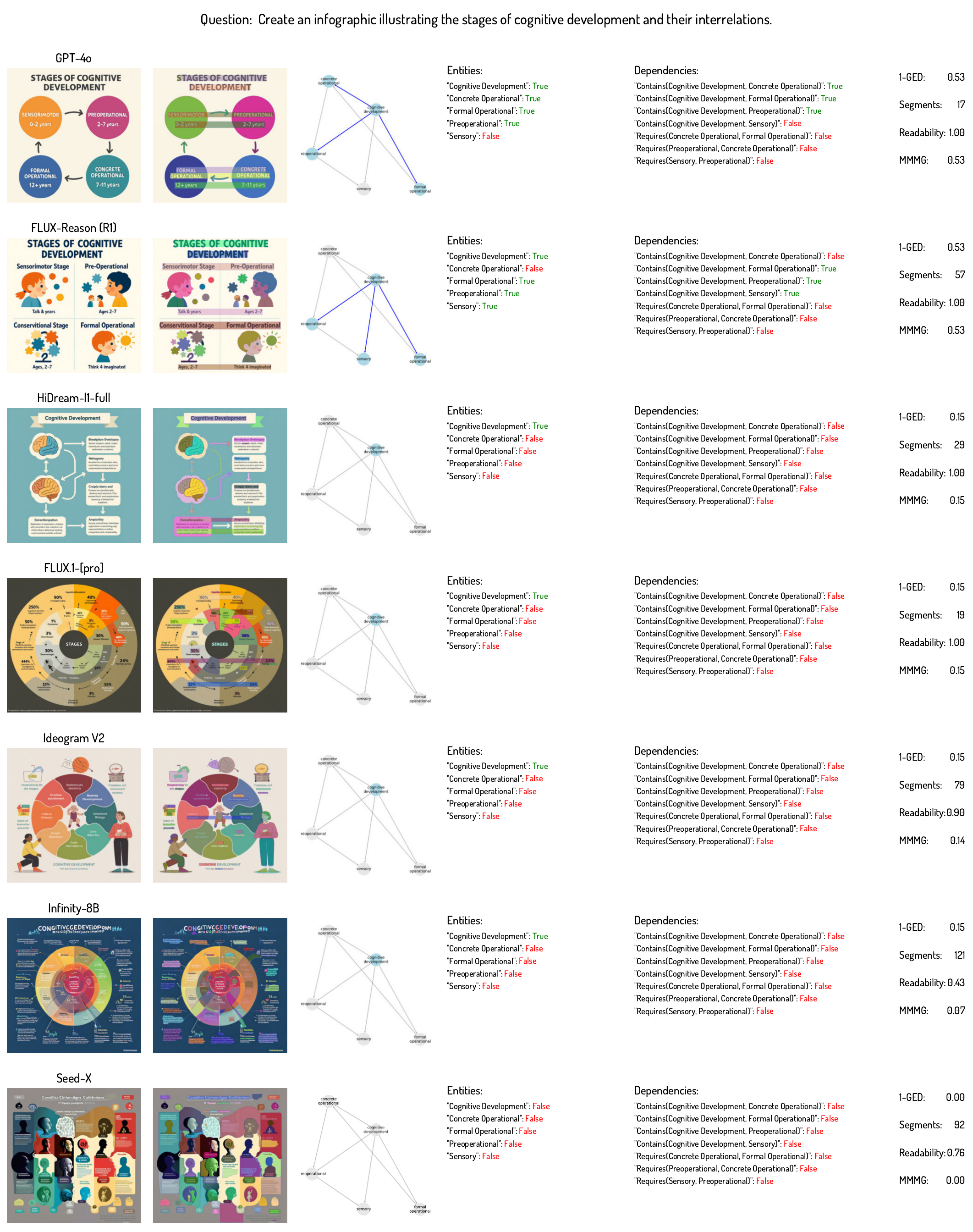}
    \caption{\texttt{MMMG} Benchmark visualization for seven representative models on a Undergraduate‐Sociology example. Each row corresponds to one model and, from left to right, displays the generated image, its segmentation map, the reconstructed knowledge graph, the extracted entity and dependency lists, and finally the overall \texttt{MMMG‐Score} along with its component sub‐scores.}
    \label{fig:enter-label}
\end{figure}
\newpage
\subsubsection{History}
\begin{figure}[h]
    \centering
    \includegraphics[width=\linewidth]{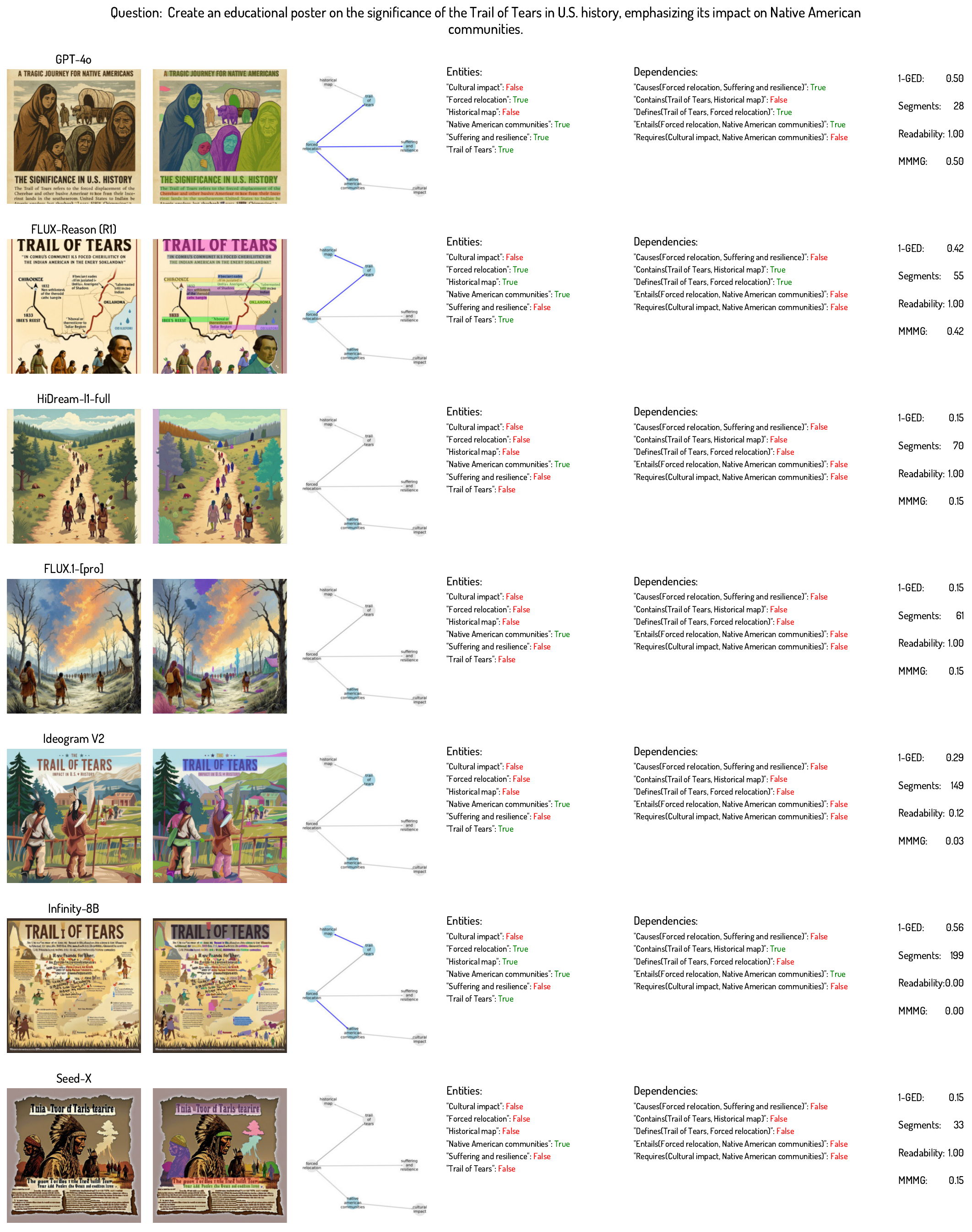}
    \caption{\texttt{MMMG} Benchmark visualization for seven representative models on a Undergraduate‐History example. Each row corresponds to one model and, from left to right, displays the generated image, its segmentation map, the reconstructed knowledge graph, the extracted entity and dependency lists, and finally the overall \texttt{MMMG‐Score} along with its component sub‐scores.}
    \label{fig:enter-label}
\end{figure}
\newpage
\subsubsection{Philosophy}
\begin{figure}[h]
    \centering
    \includegraphics[width=\linewidth]{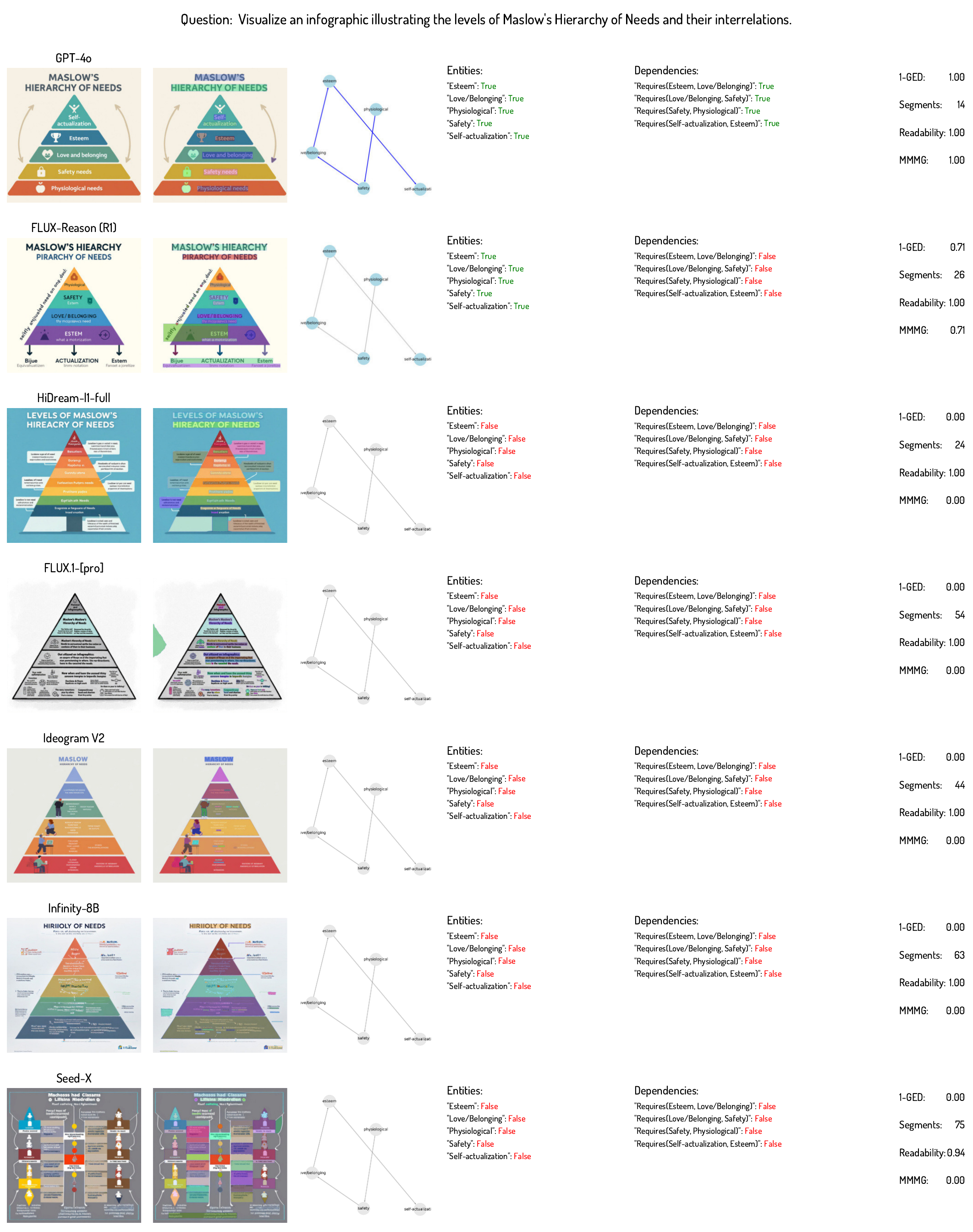}
    \caption{\texttt{MMMG} Benchmark visualization for seven representative models on a Undergraduate‐Philosophy example. Each row corresponds to one model and, from left to right, displays the generated image, its segmentation map, the reconstructed knowledge graph, the extracted entity and dependency lists, and finally the overall \texttt{MMMG‐Score} along with its component sub‐scores.}
    \label{fig:enter-label}
\end{figure}
\newpage
\subsubsection{Literature}
\begin{figure}[h]
    \centering
    \includegraphics[width=\linewidth]{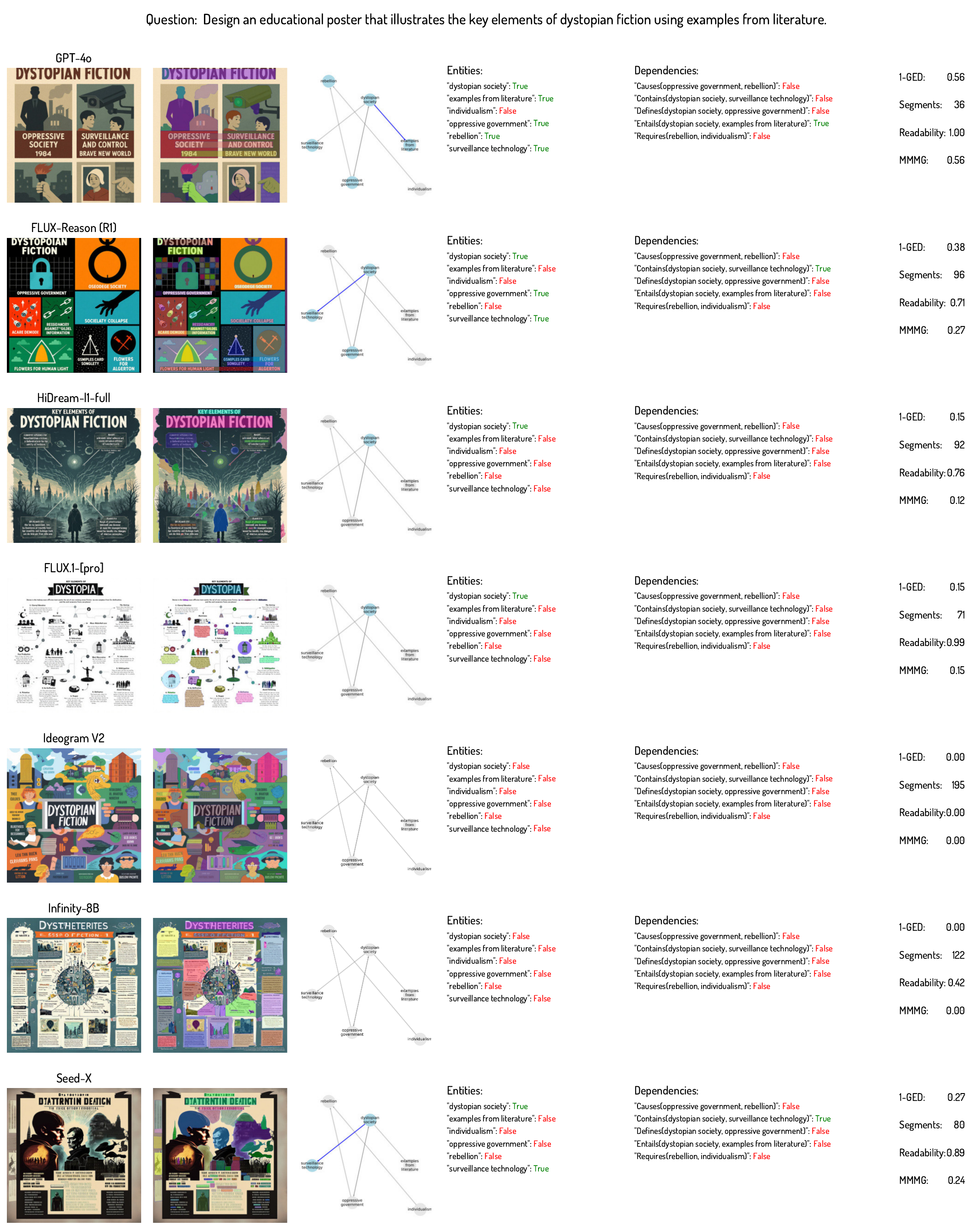}
    \caption{\texttt{MMMG} Benchmark visualization for seven representative models on a Undergraduate‐Literature example. Each row corresponds to one model and, from left to right, displays the generated image, its segmentation map, the reconstructed knowledge graph, the extracted entity and dependency lists, and finally the overall \texttt{MMMG‐Score} along with its component sub‐scores.}
    \label{fig:enter-label}
\end{figure}

\newpage
\subsection{PhD}

\subsubsection{Biology}
\begin{figure}[h]
    \centering
    \includegraphics[width=\linewidth]{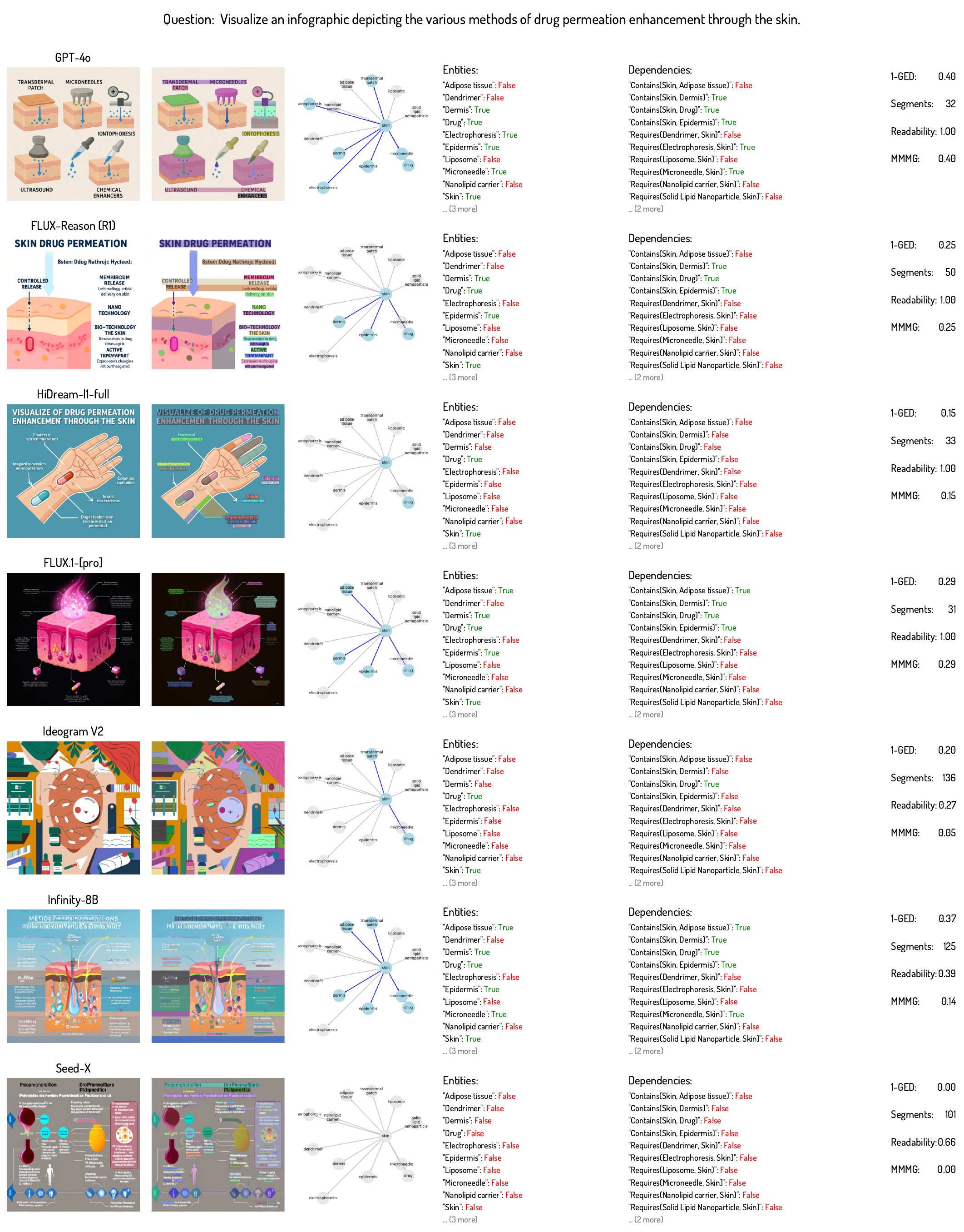}
    \caption{\texttt{MMMG} Benchmark visualization for seven representative models on a PhD‐Biology example. Each row corresponds to one model and, from left to right, displays the generated image, its segmentation map, the reconstructed knowledge graph, the extracted entity and dependency lists, and finally the overall \texttt{MMMG‐Score} along with its component sub‐scores.}
    \label{fig:enter-label}
\end{figure}
\newpage
\subsubsection{Chemistry}
\begin{figure}[h]
    \centering
    \includegraphics[width=\linewidth]{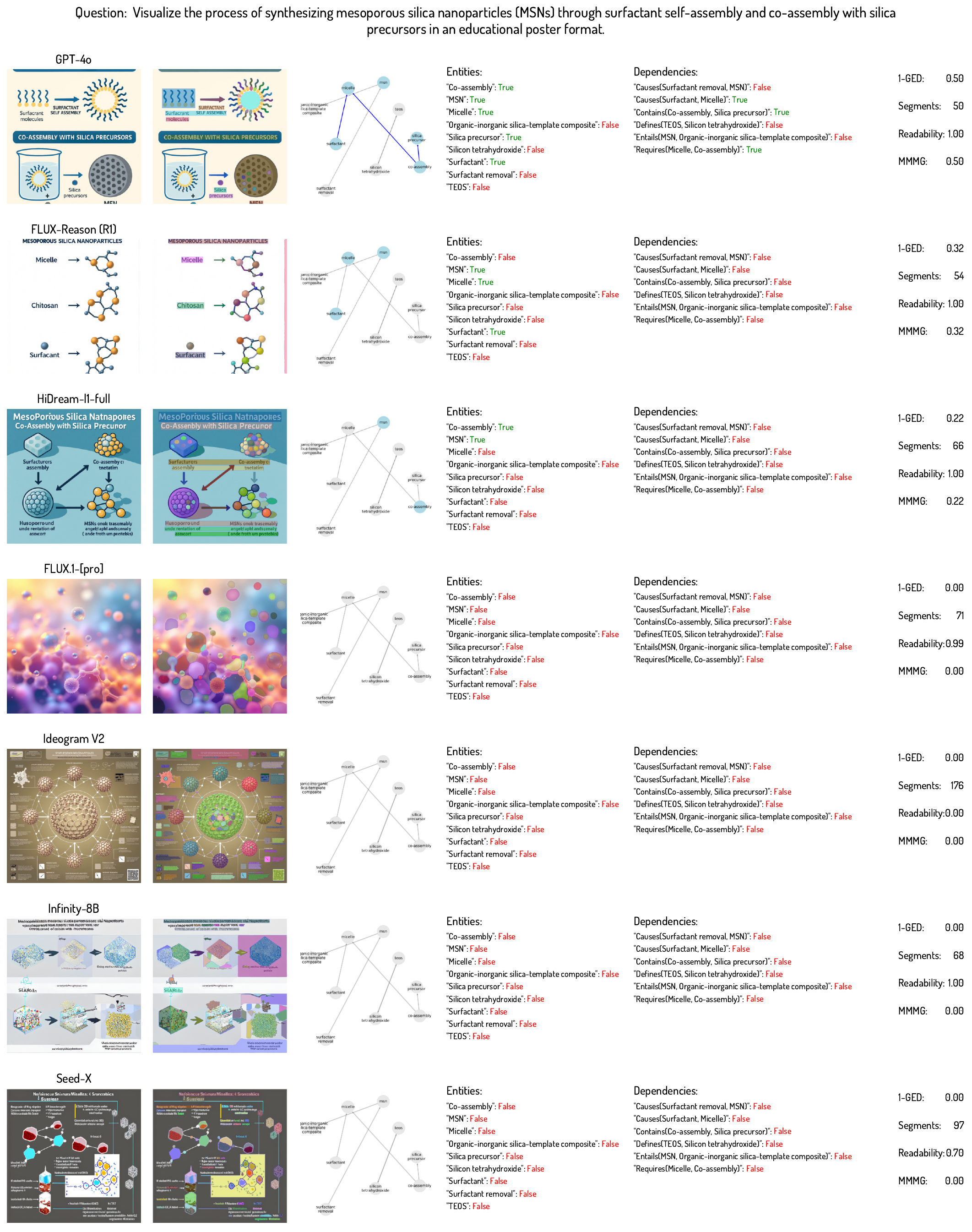}
    \caption{\texttt{MMMG} Benchmark visualization for seven representative models on a PhD‐Chemistry example. Each row corresponds to one model and, from left to right, displays the generated image, its segmentation map, the reconstructed knowledge graph, the extracted entity and dependency lists, and finally the overall \texttt{MMMG‐Score} along with its component sub‐scores.}
    \label{fig:enter-label}
\end{figure}
\newpage
\subsubsection{Mathematics}
\begin{figure}[h]
    \centering
    \includegraphics[width=\linewidth]{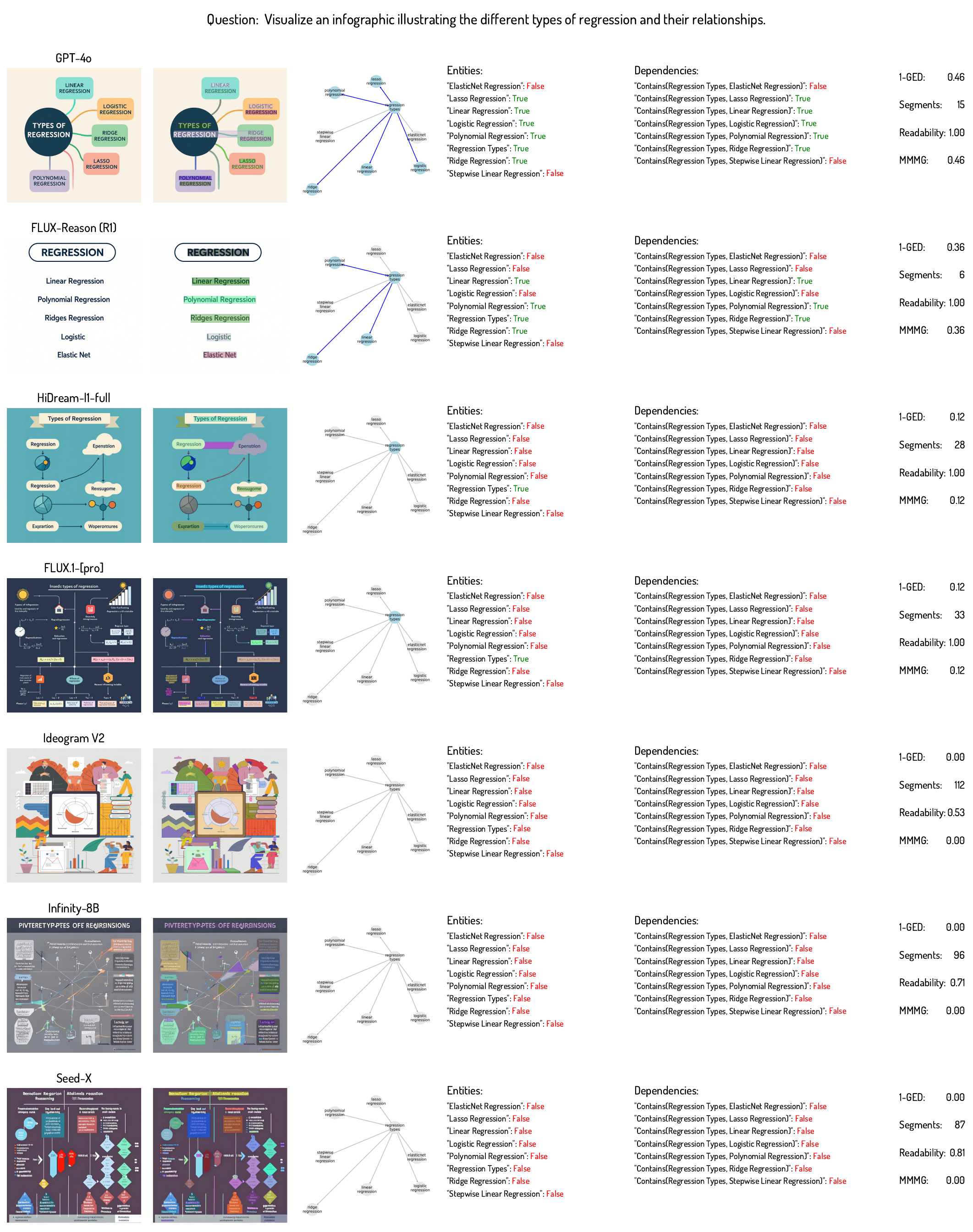}
    \caption{\texttt{MMMG} Benchmark visualization for seven representative models on a PhD‐Mathematics example. Each row corresponds to one model and, from left to right, displays the generated image, its segmentation map, the reconstructed knowledge graph, the extracted entity and dependency lists, and finally the overall \texttt{MMMG‐Score} along with its component sub‐scores.}
    \label{fig:enter-label}
\end{figure}
\newpage
\subsubsection{Engineering}
\begin{figure}[h]
    \centering
    \includegraphics[width=\linewidth]{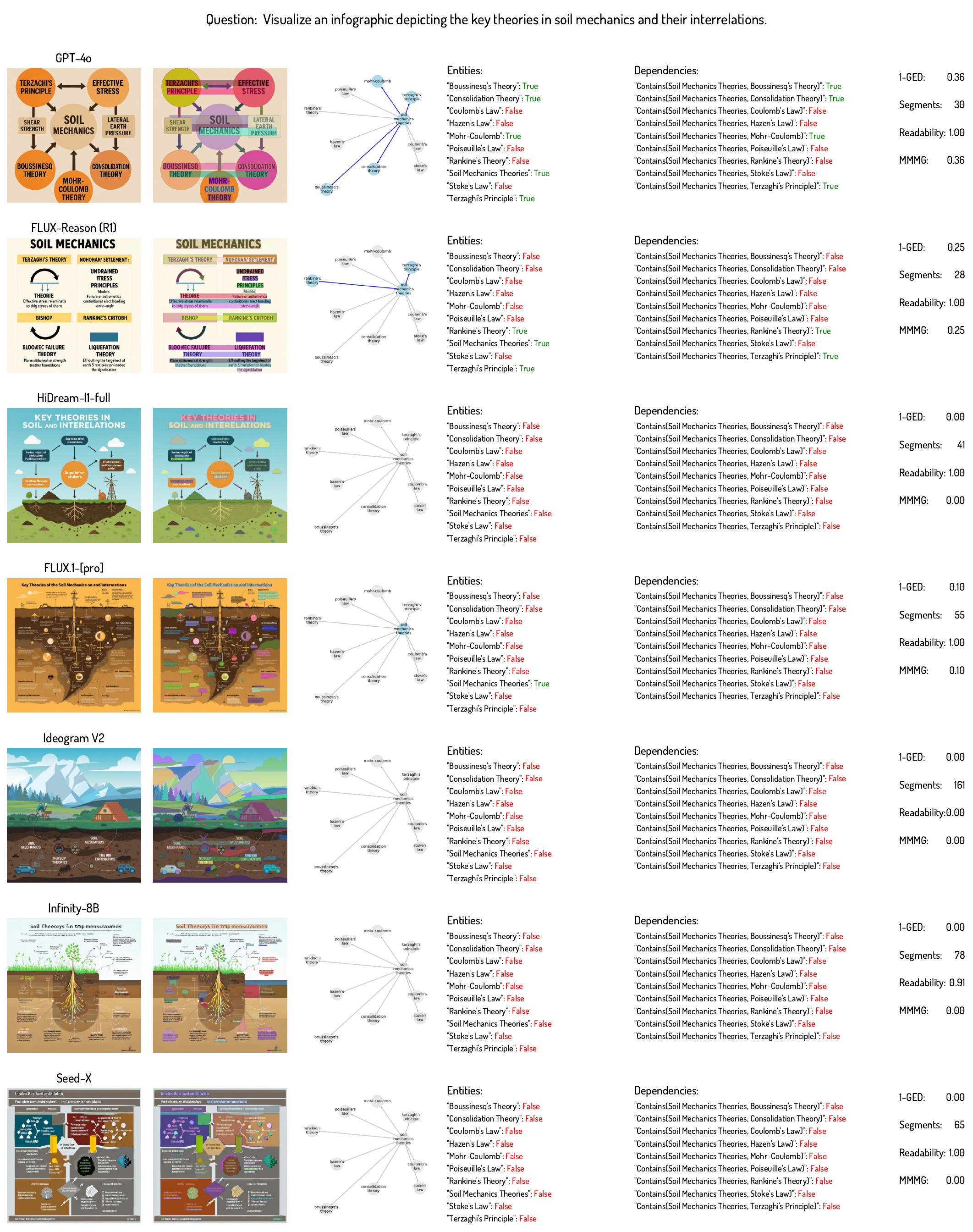}
    \caption{\texttt{MMMG} Benchmark visualization for seven representative models on a PhD‐Engineering example. Each row corresponds to one model and, from left to right, displays the generated image, its segmentation map, the reconstructed knowledge graph, the extracted entity and dependency lists, and finally the overall \texttt{MMMG‐Score} along with its component sub‐scores.}
    \label{fig:enter-label}
\end{figure}
\newpage
\subsubsection{Geography}
\begin{figure}[h]
    \centering
    \includegraphics[width=\linewidth]{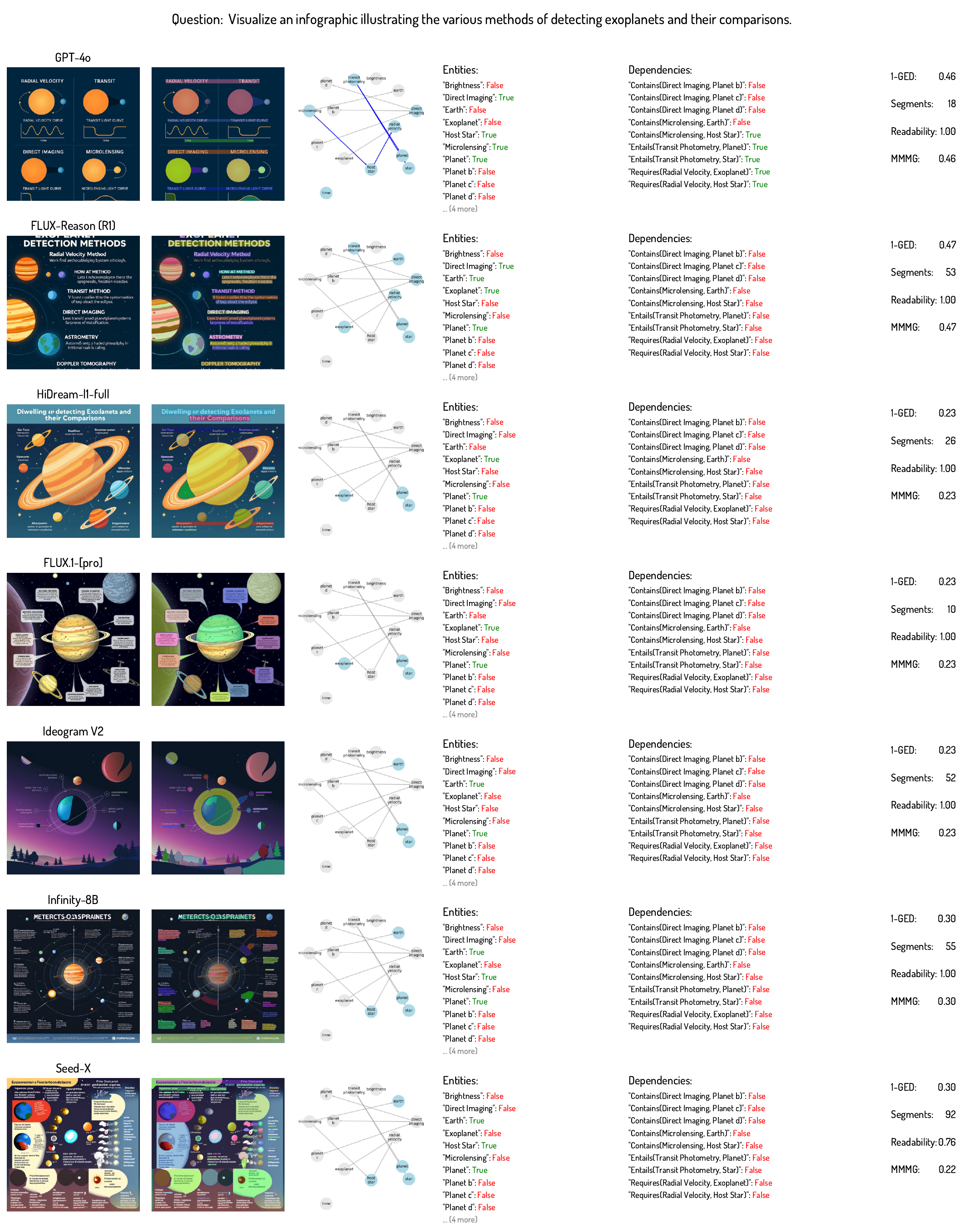}
    \caption{\texttt{MMMG} Benchmark visualization for seven representative models on a PhD‐Geography example. Each row corresponds to one model and, from left to right, displays the generated image, its segmentation map, the reconstructed knowledge graph, the extracted entity and dependency lists, and finally the overall \texttt{MMMG‐Score} along with its component sub‐scores.}
    \label{fig:enter-label}
\end{figure}
\newpage
\subsubsection{Economics}
\begin{figure}[h]
    \centering
    \includegraphics[width=\linewidth]{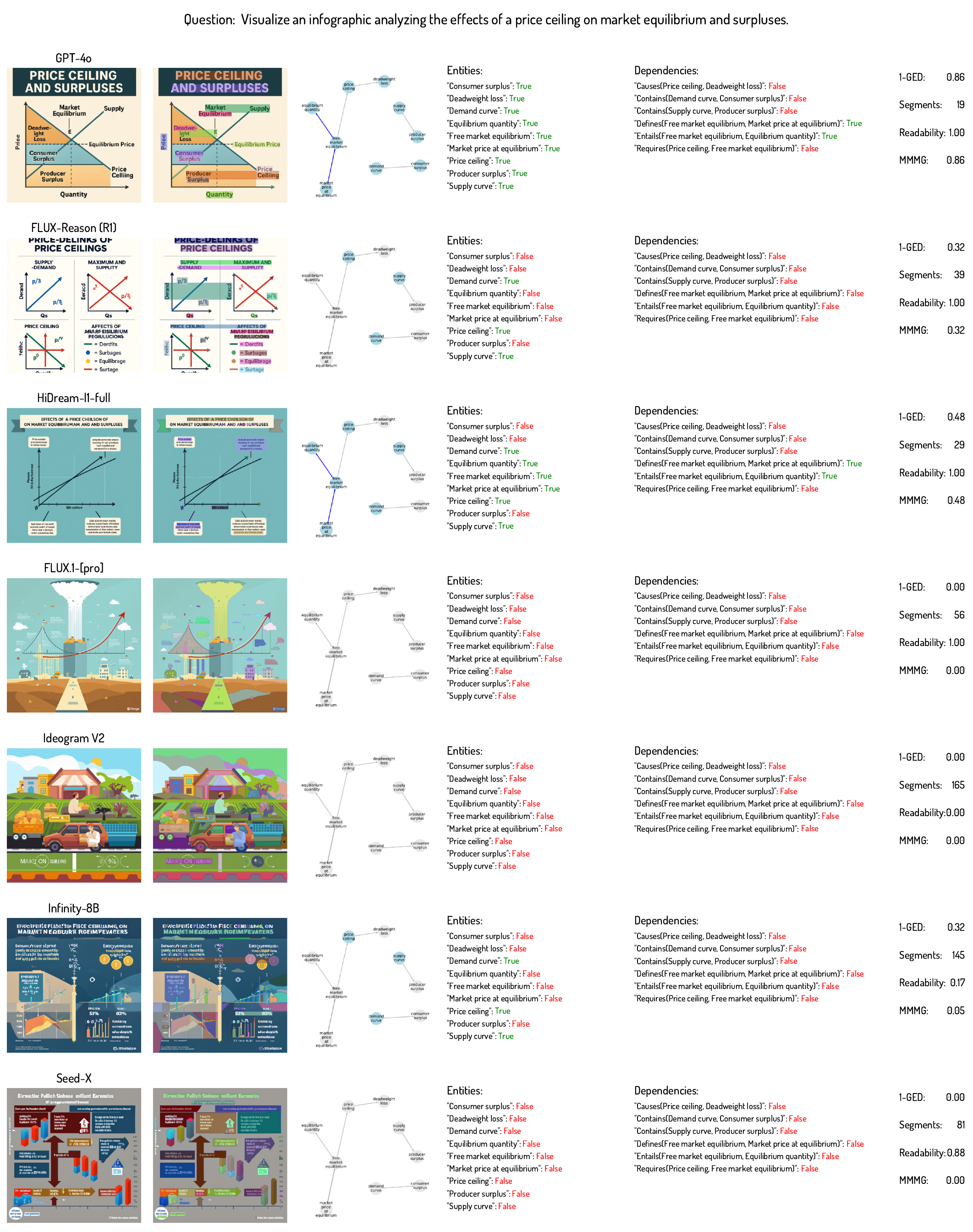}
    \caption{\texttt{MMMG} Benchmark visualization for seven representative models on a PhD‐Economics example. Each row corresponds to one model and, from left to right, displays the generated image, its segmentation map, the reconstructed knowledge graph, the extracted entity and dependency lists, and finally the overall \texttt{MMMG‐Score} along with its component sub‐scores.}
    \label{fig:enter-label}
\end{figure}
\newpage
\subsubsection{Sociology}
\begin{figure}[h]
    \centering
    \includegraphics[width=\linewidth]{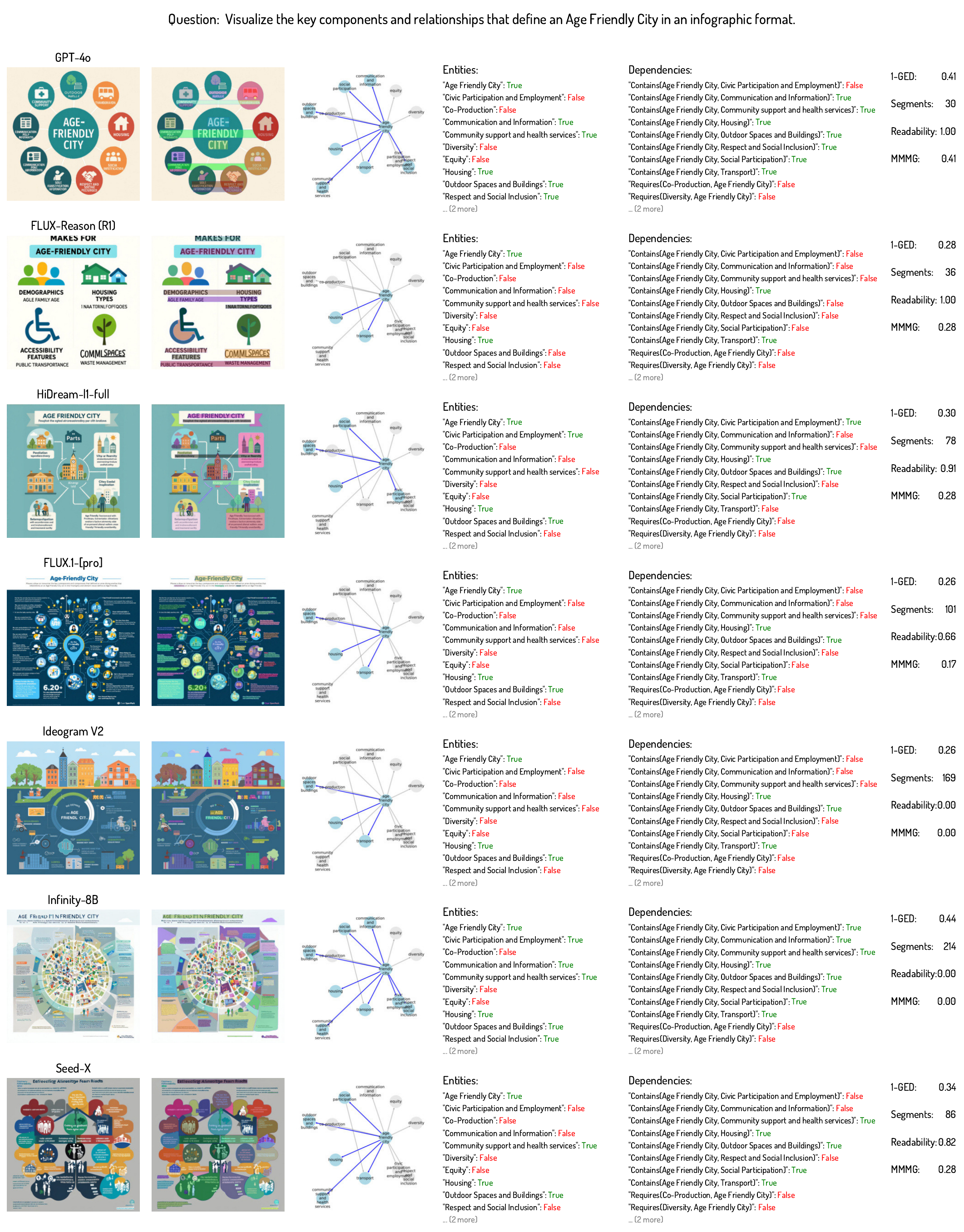}
    \caption{\texttt{MMMG} Benchmark visualization for seven representative models on a PhD‐Sociology example. Each row corresponds to one model and, from left to right, displays the generated image, its segmentation map, the reconstructed knowledge graph, the extracted entity and dependency lists, and finally the overall \texttt{MMMG‐Score} along with its component sub‐scores.}
    \label{fig:enter-label}
\end{figure}
\newpage
\subsubsection{History}
\begin{figure}[h]
    \centering
    \includegraphics[width=\linewidth]{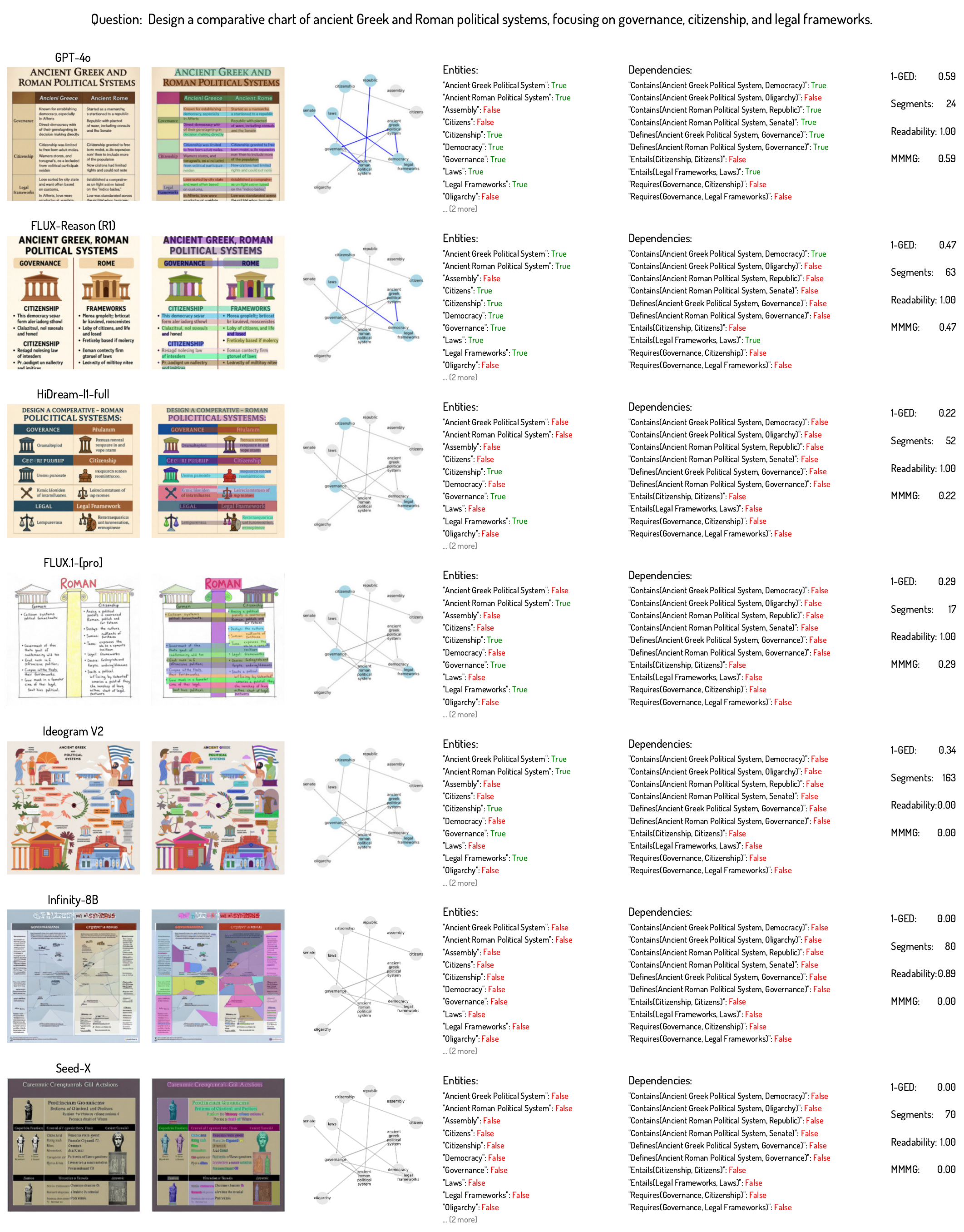}
    \caption{\texttt{MMMG} Benchmark visualization for seven representative models on a PhD‐History example. Each row corresponds to one model and, from left to right, displays the generated image, its segmentation map, the reconstructed knowledge graph, the extracted entity and dependency lists, and finally the overall \texttt{MMMG‐Score} along with its component sub‐scores.}
    \label{fig:enter-label}
\end{figure}
\newpage
\subsubsection{Philosophy}
\begin{figure}[h]
    \centering
    \includegraphics[width=\linewidth]{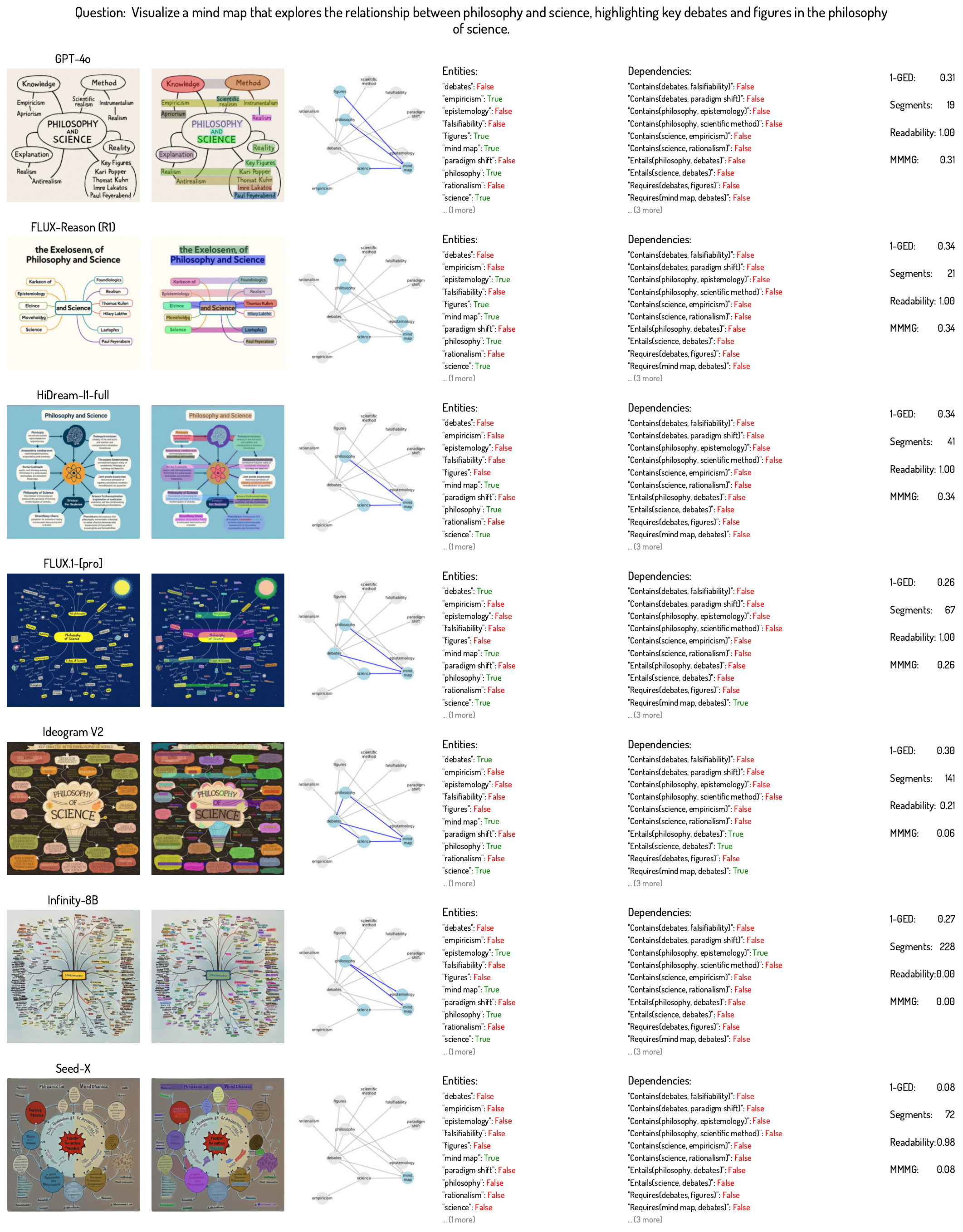}
    \caption{\texttt{MMMG} Benchmark visualization for seven representative models on a PhD‐Philosophy example. Each row corresponds to one model and, from left to right, displays the generated image, its segmentation map, the reconstructed knowledge graph, the extracted entity and dependency lists, and finally the overall \texttt{MMMG‐Score} along with its component sub‐scores.}
    \label{fig:enter-label}
\end{figure}
\newpage
\subsubsection{Literature}
\begin{figure}[h]
    \centering
    \includegraphics[width=\linewidth]{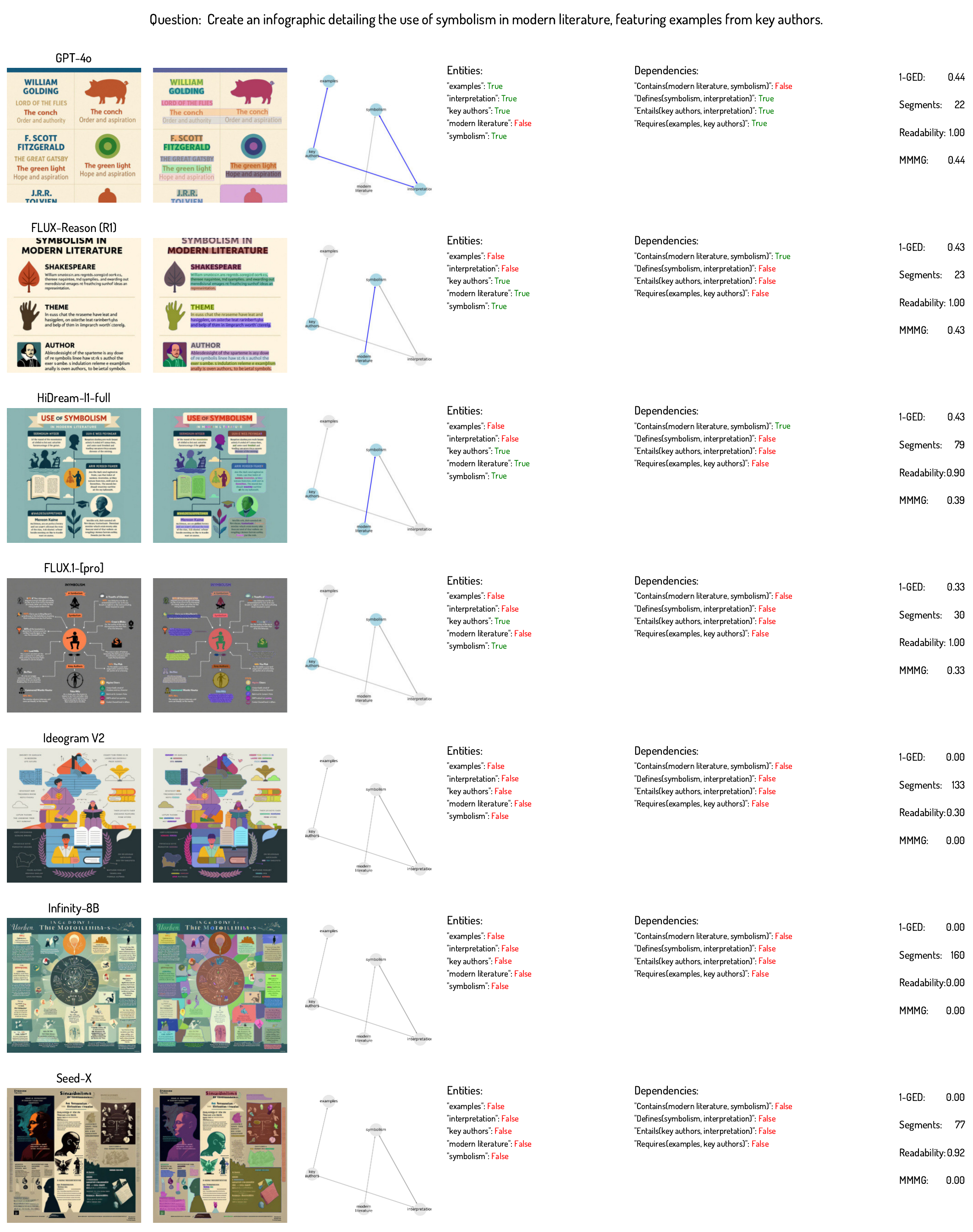}
    \caption{\texttt{MMMG} Benchmark visualization for seven representative models on a PhD‐Literature example. Each row corresponds to one model and, from left to right, displays the generated image, its segmentation map, the reconstructed knowledge graph, the extracted entity and dependency lists, and finally the overall \texttt{MMMG‐Score} along with its component sub‐scores.}
    \label{fig:enter-label}
\end{figure}

\bibliographystyle{abbrv}
\bibliography{ref}

\end{document}